\documentclass[iicol,pdflatex,sn-basic]{sn-jnl}

\usepackage{graphicx}%
\usepackage{multirow}%
\usepackage{amsmath,amssymb,amsfonts}%
\usepackage{amsthm}%
\usepackage{mathrsfs}%
\usepackage[title]{appendix}%
\usepackage{xcolor}%
\usepackage{textcomp}%
\usepackage{manyfoot}%
\usepackage{algorithm}%
\usepackage{algorithmicx}%
\usepackage{algpseudocode}%
\usepackage{listings}%

\usepackage{subcaption}
\usepackage{stfloats}
\usepackage{cleveref}
\usepackage[table]{xcolor} 
\usepackage{booktabs}      
\usepackage{tabularx}      

\usepackage{pifont}
\newcommand{\cmark}{\textcolor{green!60!black}{\ding{51}}}
\newcommand{\xmark}{\textcolor{red}{\ding{55}}}

\definecolor{verylightgray}{gray}{0.9}
\definecolor{gaincolor}{rgb}{0.0, 0.4, 0.0} 
\definecolor{losscolor}{rgb}{0.7, 0.0, 0.0} 

\newcommand{\pos}[1]{{\color{gaincolor}#1}}
\newcommand{\nega}[1]{{\color{losscolor}#1}}

\theoremstyle{thmstyleone}%
\theoremstyle{thmstyletwo}%

\theoremstyle{thmstylethree}%

\begin{document}


\title[Article Title]{PerCoV2: Ultra-Low Bit-Rate Perceptual Image Compression via Query-Based 1D Multimodal Image Tokens}


\author*[1,3]{\fnm{Nikolai} \sur{Körber}}\email{nikolai.koerber@tum.de}

\author[2]{\fnm{Eduard} \sur{Kromer}}\email{eduard.kromer@haw-landshut.de}

\author[2]{\fnm{Andreas} \sur{Siebert}}\email{andreas.siebert@haw-landshut.de}

\author[4]{\fnm{Sascha} \sur{Hauke}}\email{sascha.hauke@uni-giessen.de}

\author[5]{\fnm{Daniel} \sur{Mueller-Gritschneder}}\email{daniel.mueller-gritschneder@tuwien.ac.at}

\author[1,3]{\fnm{Björn} \sur{Schuller}}\email{schuller@tum.de}

\affil[1]{\orgname{Technical University of Munich},
\orgaddress{\country{Germany}}}

\affil[2]{\orgname{University of Applied Sciences Landshut},
\orgaddress{\country{Germany}}}

\affil[3]{\orgname{Munich Center for Machine Learning (MCML)},
\orgaddress{\country{Germany}}}

\affil[4]{\orgname{University of Giessen},
\orgaddress{\country{Germany}}}

\affil[5]{\orgname{TU Wien},
\orgaddress{\country{Austria}}}


\abstract{

Despite recent progress in learned image compression, current image codecs still struggle to maintain realistic reconstructions at low bit-rates, often producing structured artifacts such as grid patterns or repetitive textures, even when trained with perceptual or adversarial losses. In this work, we introduce PerCoV2, an ultra-low bit-rate perceptual image compression system that unifies semantic tokenization, flow-based generation, and learned entropy modeling within a single framework. Building on the fully open flow-based SANA architecture, PerCoV2 introduces a novel resolution-adaptive 1D query-based tokenizer whose compact image tokens serve a dual role in flow matching: they provide a data-dependent reconstruction prior for initialization while simultaneously acting as a conditioning signal for flow-based refinement. By explicitly decoupling semantic representation from perceptual generation, our dual representation simplifies the flow-based learning objective, leading to more stable optimization and improved perceptual compression performance. PerCoV2 further introduces a dedicated 1D masked entropy model to improve rate efficiency and optional decoder-side multimodal enhancement via a vision–language model (Molmo) without increasing the transmitted bit budget. On MSCOCO-30k, PerCoV2 achieves state-of-the-art statistical fidelity, measured by FID and KID, across ultra-low and extreme bit-rates (0.0015–0.025 bpp). When trained solely on the general-purpose SA-1B dataset, PerCoV2 further demonstrates strong zero-shot generalization to widely adopted high-resolution benchmarks, including DIV2K and CLIC 2020, achieving competitive statistical fidelity with the current leading method, AEIC-ME. Finally, we introduce PerCoV2-distilled, a practical single-step variant derived from multi-step flow matching that accelerates decoding by $5.37\times$ over PerCoV1, while preserving perceptual compression performance. Code and pre-trained models will be released upon publication at \url{https://github.com/nikolai10/PerCoV2}.}

\keywords{Perceptual Image Compression, Learned Image Compression, Flow-based Generative Models, Entropy Modeling, Vision-Language Models}



\maketitle


\section{Introduction}\label{sec1}

Perceptual compression, also known as generative~\citep{Agustsson_2019_ICCV, mentzer2020high} or distribution-preserving compression~\citep{NEURIPS2018_801fd8c2}, represents a class of neural image compression techniques that integrate generative models---such as generative adversarial networks (GANs)~\citep{NIPS2014_5ca3e9b1} and diffusion models~\citep{pmlr-v37-sohl-dickstein15, NEURIPS2020_4c5bcfec}---into their optimization objectives. Unlike traditional codecs like JPEG, which primarily focus on minimizing pixel-wise distortion, these methods further constrain reconstructions to align with the underlying data distribution~\citep{ICML-2019-BlauM}. By leveraging powerful generative priors, they can synthesize realistic details and textures, producing reconstructions that better match the natural image distribution while often improving perceived visual quality at considerably lower bit-rates. These advantages make perceptual compression particularly compelling for storage- and bandwidth-constrained applications.

Recently, foundation models~\citep{Bommasani2021FoundationModels_short}---large-scale machine learning models trained on broad data at scale---have shown great potential in their adaptation to a wide variety of downstream tasks, including ultra-low bit-rate perceptual image compression~\citep{pan_2022, lei2023text+sketch, careil2024towards, li2024misc, li2024towards, xia2025diffpc, Ke2025resulic}. In particular, PerCo~\citep{careil2024towards}, which we also refer to as PerCoV1 throughout this paper, is the first method to explore bit-rates as low as $0.003$~bpp. This is achieved by extending the conditioning mechanism of a pre-trained text-conditional latent diffusion model (LDM) with vector-quantized hyper-latent image features. Consequently, only a short text description and a compact compressed image representation are required during decoding. Despite its potential and fascinating results, PerCo remains unavailable to the public, mainly due to its reliance on a proprietary LDM built upon the GLIDE~\citep{pmlr-v162-nichol22a} architecture.

\begin{figure}[tb]
  \centering
  \includegraphics[width=1\linewidth]{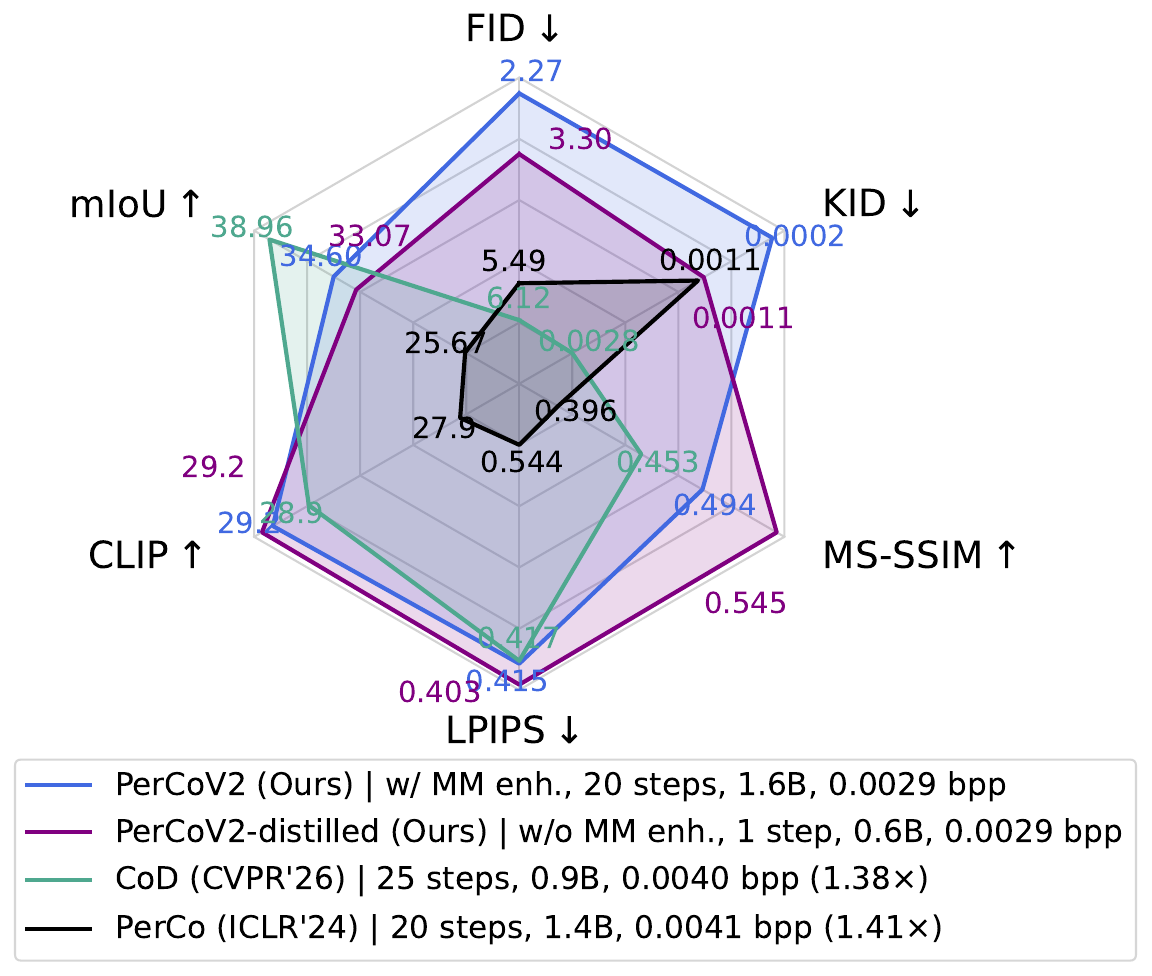}
  \caption{Perceptual compression performance of PerCoV2 on MSCOCO-30k. A larger area indicates better performance.}
  \label{fig:teaser}
\end{figure}

\begin{figure*}[tb]
    \setlength{\tabcolsep}{1.0pt}  
    \renewcommand{\arraystretch}{1.0}  
    \centering
    \scriptsize
    \begin{tabular}{ccccc|c}
        \toprule
        Original & MS-ILLM & DiffC & PerCoV1 & CoD latent & PerCoV2 \\
                 & (ICML'23) & (ICLR'25) & (ICLR'24) & (CVPR'26) & (Ours) \\
        \midrule
        \begin{subfigure}{0.160\textwidth}
            \centering
            \includegraphics[width=\linewidth]{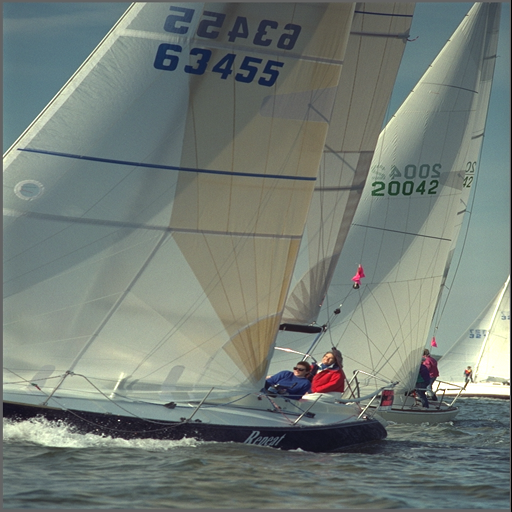}
        \end{subfigure}
        &
        \begin{subfigure}{0.160\textwidth}
            \centering
            \includegraphics[width=\linewidth]{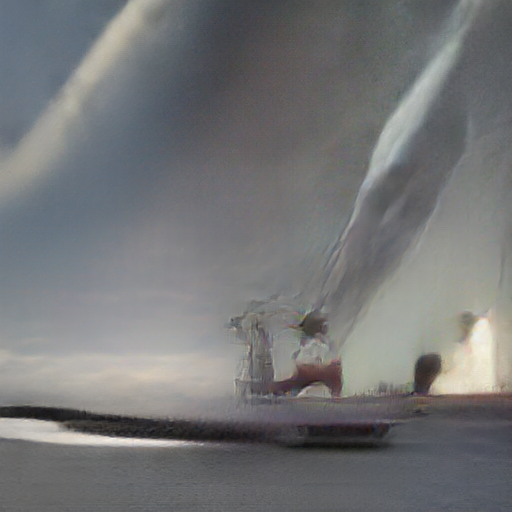}
        \end{subfigure}
        &
        \begin{subfigure}{0.160\textwidth}
            \centering
            \includegraphics[width=\linewidth]{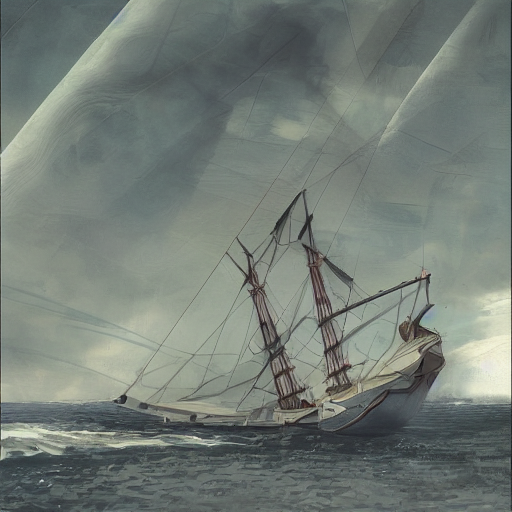}
        \end{subfigure}
        &
        \begin{subfigure}{0.160\textwidth}
            \centering
            \includegraphics[width=\linewidth]{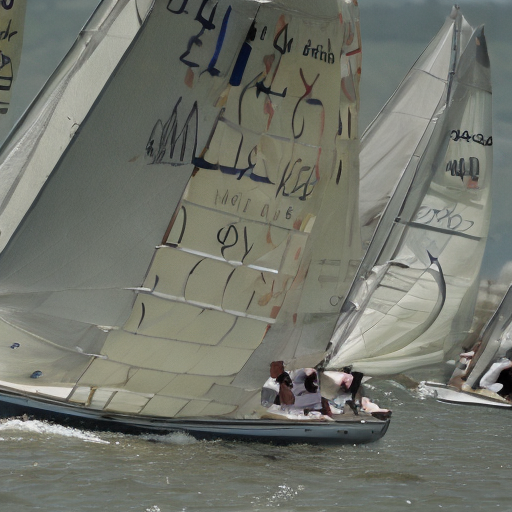}
        \end{subfigure}
        &
        \begin{subfigure}{0.160\textwidth}
            \centering
            \includegraphics[width=\linewidth]{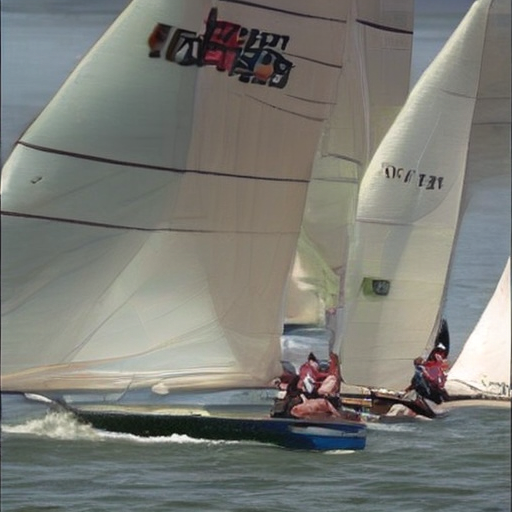}
        \end{subfigure}
        &
        \begin{subfigure}{0.160\textwidth}
            \centering
            \includegraphics[width=\linewidth]{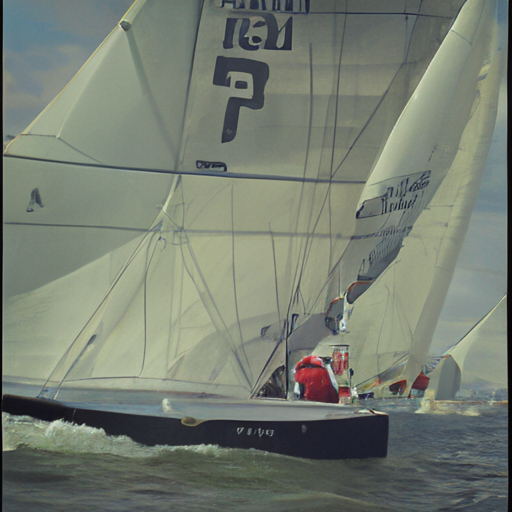}
        \end{subfigure} \\
        
        \texttt{kodim10} & $0.02248$~bpp & $0.00415$~bpp & $0.00339$~bpp & $0.00403$~bpp & $0.00293$~bpp \\
        & ($2.49\times$) & ($1.42\times$) & ($1.16\times$) & ($1.37\times$)&  \\
        \begin{subfigure}{0.160\textwidth}
            \centering
            \includegraphics[width=\linewidth]{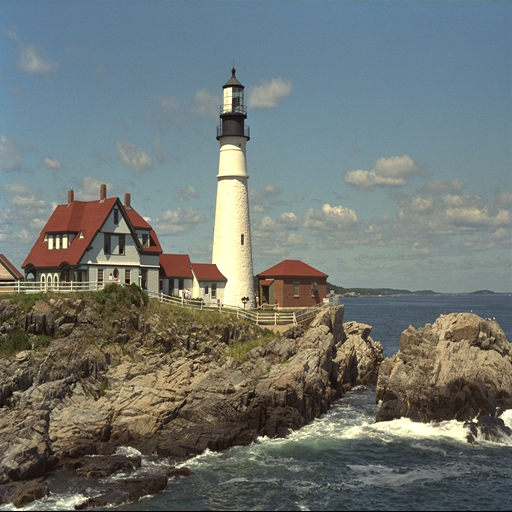}
        \end{subfigure}
        &
        \begin{subfigure}{0.160\textwidth}
            \centering
            \includegraphics[width=\linewidth]{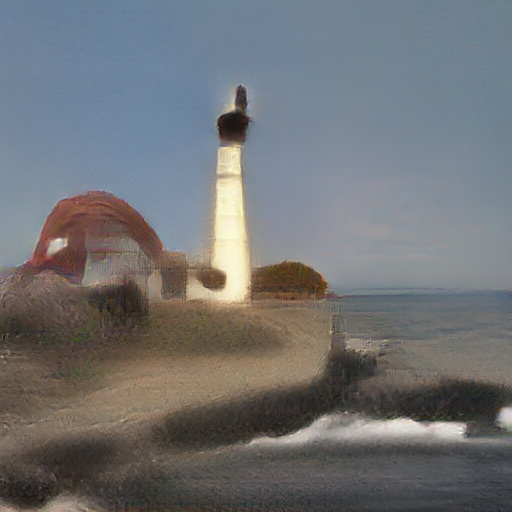}
        \end{subfigure}
        &
        \begin{subfigure}{0.160\textwidth}
            \centering
            \includegraphics[width=\linewidth]{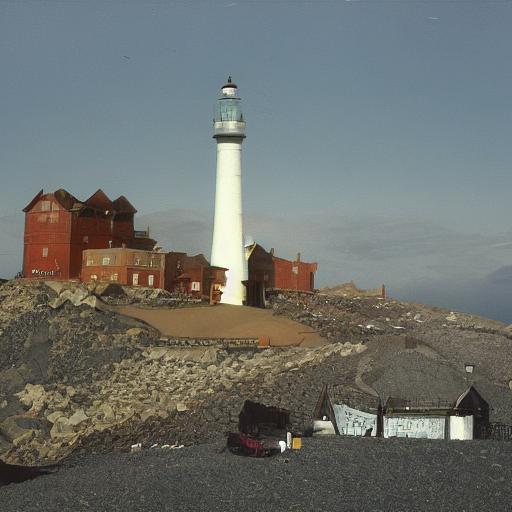}
        \end{subfigure}
        &
        \begin{subfigure}{0.160\textwidth}
            \centering
            \includegraphics[width=\linewidth]{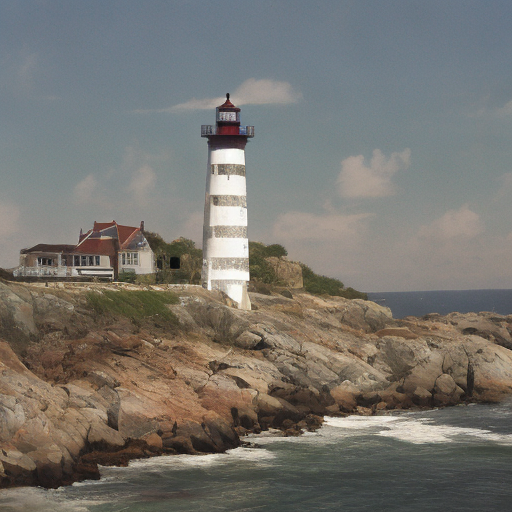}
        \end{subfigure}
        &
        \begin{subfigure}{0.160\textwidth}
            \centering
            \includegraphics[width=\linewidth]{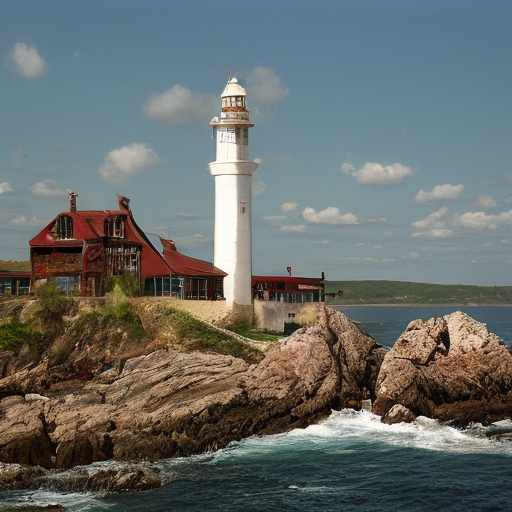}
        \end{subfigure}
        &        
        \begin{subfigure}{0.160\textwidth}
            \centering
            \includegraphics[width=\linewidth]{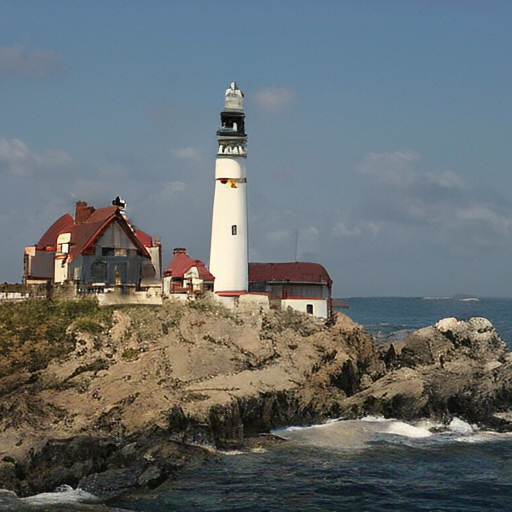}
        \end{subfigure} \\        
        \texttt{kodim21} & $0.02038$~bpp & $0.00421$~bpp & $0.00388$~bpp & $0.00403$~bpp & $0.00293$~bpp \\
         & ($2.67\times$) & ($1.44\times$) & ($1.32\times$) & ($1.37\times$)  & \\
        \begin{subfigure}{0.160\textwidth}
            \centering
            \includegraphics[width=\linewidth]{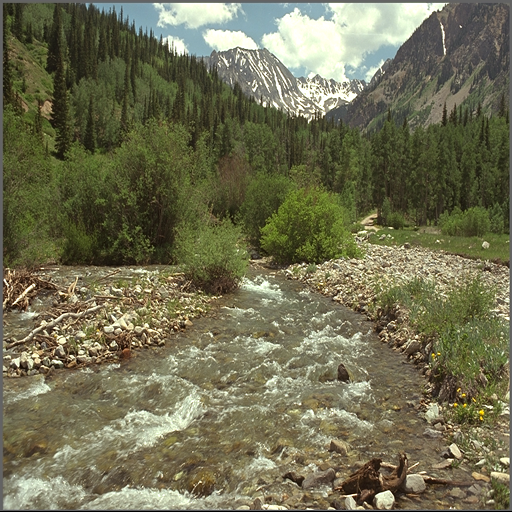}
        \end{subfigure}
        &
        \begin{subfigure}{0.160\textwidth}
            \centering
            \includegraphics[width=\linewidth]{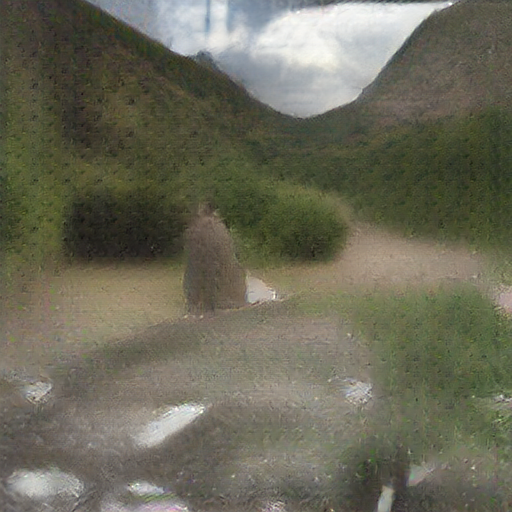}
        \end{subfigure}
        &
        \begin{subfigure}{0.160\textwidth}
            \centering
            \includegraphics[width=\linewidth]{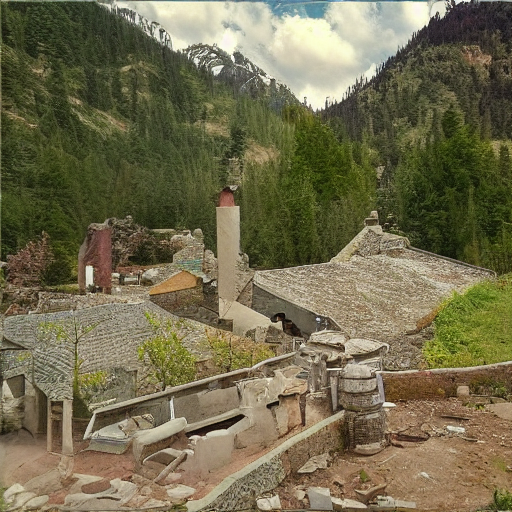}
        \end{subfigure}
        &
        \begin{subfigure}{0.160\textwidth}
            \centering
            \includegraphics[width=\linewidth]{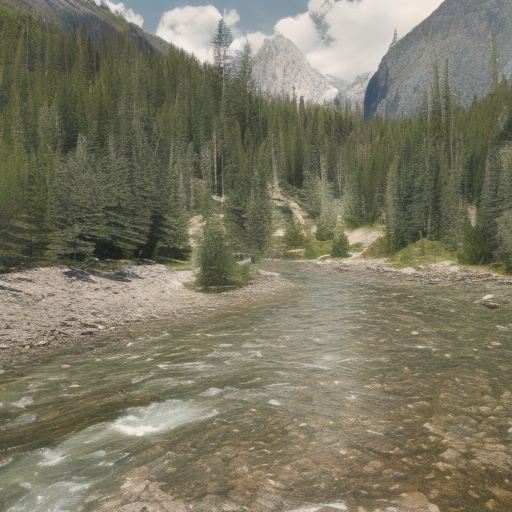}
        \end{subfigure}       
        &
        \begin{subfigure}{0.160\textwidth}
            \centering
            \includegraphics[width=\linewidth]{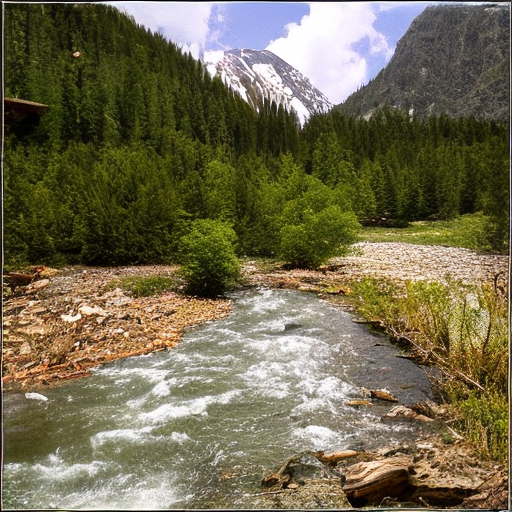}
        \end{subfigure}
        & 
        \begin{subfigure}{0.160\textwidth}
            \centering
            \includegraphics[width=\linewidth]{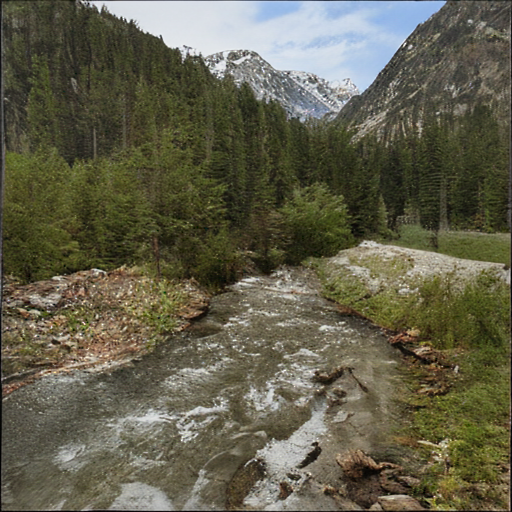}
        \end{subfigure} \\
        \texttt{kodim13} & $0.00757$~bpp & $0.00464$~bpp & $0.00406$~bpp & $0.00403$~bpp & $0.00293$~bpp \\
        & ($2.58\times$) & ($1.58\times$) & ($1.39\times$) & ($1.37\times$) & \\
        \begin{subfigure}{0.160\textwidth}
            \centering
            \includegraphics[width=\linewidth]{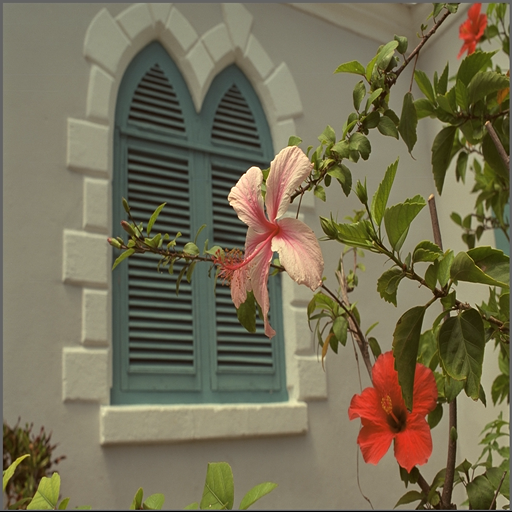}
        \end{subfigure}
        &
        \begin{subfigure}{0.160\textwidth}
            \centering
            \includegraphics[width=\linewidth]{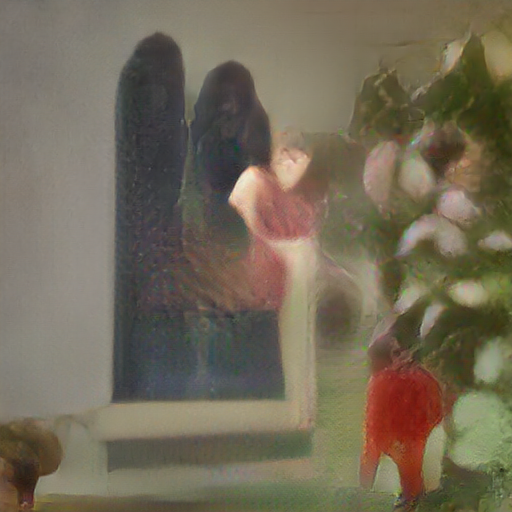}
        \end{subfigure}
        &
        \begin{subfigure}{0.160\textwidth}
            \centering
            \includegraphics[width=\linewidth]{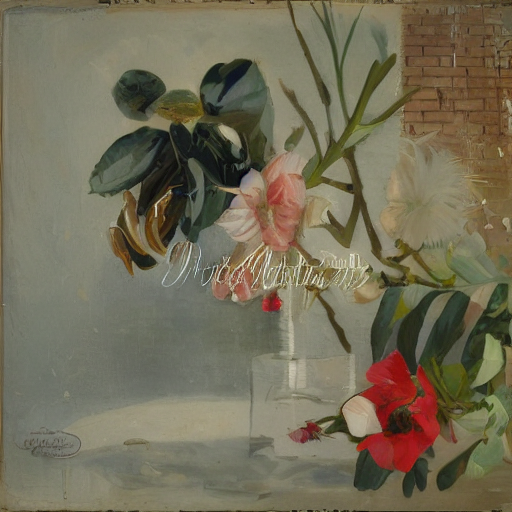}
        \end{subfigure}
        &
        \begin{subfigure}{0.160\textwidth}
            \centering
            \includegraphics[width=\linewidth]{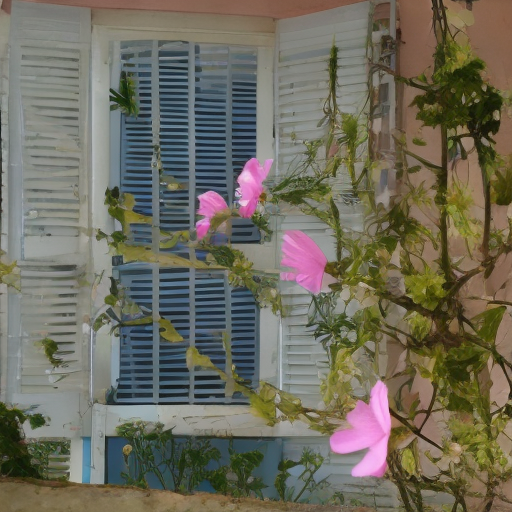}
        \end{subfigure}
        &
        \begin{subfigure}{0.160\textwidth}
            \centering
            \includegraphics[width=\linewidth]{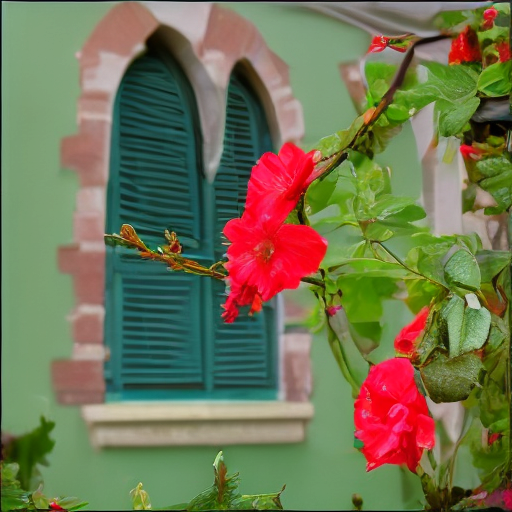}
        \end{subfigure}
        &
        \begin{subfigure}{0.160\textwidth}
            \centering
            \includegraphics[width=\linewidth]{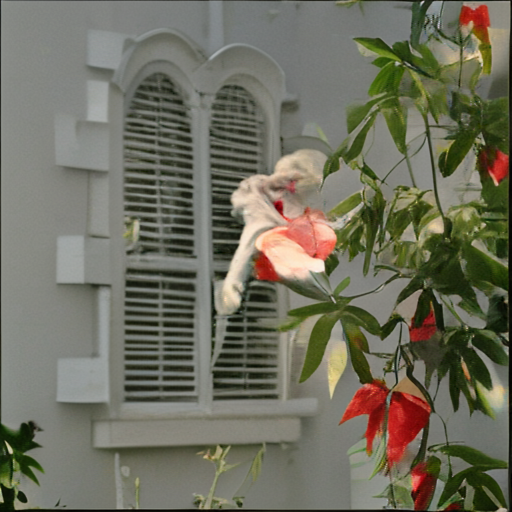}
        \end{subfigure} \\
        \texttt{kodim07} & $0.00793$~bpp & $0.00491$~bpp & $0.00372$~bpp & $0.00403$~bpp & $0.00293$~bpp \\
        & ($2.71\times$) & ($1.68\times$) & ($1.27\times$) & ($1.37\times$) & \\
        \bottomrule

    \end{tabular}

    \caption{Visual comparison of PerCoV2 on the Kodak dataset. Bit-rate increases relative to our method are indicated by $(\times)$. \textbf{Best viewed electronically.}}

    \label{fig:vis_impressions}
\end{figure*}

Although community efforts have attempted to bring PerCo into the public domain, e.g., PerCo (SD)~\citep{koerber2024perco}, we find that such reconstructions typically deviate considerably from the original results. This observation suggests that the design of the latent space and the LDM capacity play a critical role in overall perceptual compression performance. Currently, it remains unclear how the proprietary LDM performs in comparison to existing off-the-shelf models, particularly when analyzed across the consistency-diversity-realism trade-offs~\citep{Astolfi2024}. 

In this work, we introduce PerCoV2, a novel ultra-low bit-rate perceptual image compression system that unifies semantic tokenization, flow-based generation, and learned entropy modeling within a single framework. PerCoV2 retains the core design principles of PerCo~\citep{careil2024towards}, while introducing several key advancements.

\textbf{Open flow-based LDM.} PerCoV2 features a modern, open flow-based alternative based on the SANA~\citep{xie2025sana} architecture. In our preliminary studies, we extensively evaluated the impact of various LDMs on the perceptual compression performance under the original PerCoV1 formulation. While we generally found Stable Diffusion 3~\citep{esser2024scaling} to be superior, we chose SANA specifically for its highly efficient design---featuring a linear diffusion transformer and a highly compressed latent space---as well as its modern decoder-only LLM support (Gemma 2~\citep{gemma}). This integration removes the need for three separate, and partially redundant, text models (CLIP-G/14~\citep{pmlr-v139-radford21a}, CLIP-L/14~\citep{pmlr-v139-radford21a}, and T5 XXL~\citep{JMLR:v21:20-074}). Furthermore, the SANA/Gemma 2 framework is not constrained by the fixed 77-token CLIP limit, thereby offering superior potential for multimodal integration.
 
\textbf{Improved representation learning via dual-purpose 1D query-based tokenizer.} 


As its core contribution, PerCoV2 introduces a novel resolution-adaptive 1D query-based image tokenizer for conditioning the generative model. Unlike conventional tokenizers based on spatial feature grids (e.g., CNNs), our query-based design eliminates spatial redundancy, producing compact image tokens that improve compression efficiency. These 1D image tokens constitute the sole transmitted representation and are used directly to extract conditioning features for the LDM. In addition, they enable a coarse semantic reconstruction that serves as a data-dependent initialization for generation. From a flow-matching perspective, we replace the standard Gaussian initialization with a noised version of this reconstruction and learn the transition from this semantically grounded prior to the target image distribution~\citep{10.5555/3692070.3692108,10.1007/978-3-031-73030-6_19}. By initializing generation from this data-dependent prior, the resulting flow-matching objective is simplified, leading to more efficient learning in the ultra-low bit-rate regime.

\textbf{Explicit entropy modeling via 1D masked image model (1D~MIM).} PerCoV2 introduces a novel 1D~MIM for efficient entropy coding of discrete VQ image latents. While its design is inspired by recent masked modeling approaches in 2D latent spaces~\citep{chang2022maskgit, el-nouby2023image, Mentzer_2023_ICCV}, PerCoV2 directly operates on a 1D tokenized representation induced by our query-based tokenizer. This design yields an entropy model that is fully compatible with both compression and generative sampling within a unified latent space.

\textbf{Decoder-side VLM enhancement.} PerCoV2 optionally enables multimodal enhancement via a Vision-Language Model (Molmo~\citep{DBLP:journals/corr/abs-2409-17146}). In contrast to prior approaches that integrate language-based refinement into joint rate-distortion optimization or rely on prompt-level optimization strategies~\citep{wen2023hard, lei2023text+sketch, gal2023an, pan_2022, Ke2025resulic}, PerCoV2 instead adopts a post-hoc enhancement strategy. Specifically, the VLM operates on a downscaled version of the initial base reconstruction produced by our tokenizer-based decoder. This representation is, due to our dual-purpose tokenizer design, sufficiently semantically faithful and robust to compression artifacts, making downstream VLM processing effective even in ultra-low bit-rate regimes.

\textbf{High-resolution support, modular design, and one-step diffusion decoding.} PerCoV2 introduces native support for high-resolution image compression, extending beyond PerCoV1 by handling images up to 2K resolution with flexible aspect ratios (multiples of $256$). While PerCoV2 is primarily evaluated under the official PerCoV1 $512 \times 512$ protocol, we also demonstrate consistent performance at higher resolutions with strong statistical fidelity.

In addition, PerCoV2 features a modular design that enables different complexity configurations across the tokenizer, entropy model, SANA flow backbone, and optional VLM enhancement. PerCoV2 further features a practical single-step variant derived from multi-step flow matching that accelerates decoding by $5.37\times$ and $5.16\times$ over PerCoV1 and PerCoV2, respectively, while preserving perceptual compression performance.

\textbf{Scope and limitations.} PerCoV2 is designed specifically for ultra-low bit-rate perceptual image compression, where preserving semantic and statistical fidelity is prioritized over distortion-optimal reconstruction (following \citep{careil2024towards}). As such, it is not intended as a distortion-centric codec or as a method optimized for higher bit-rate regimes, where traditional approaches remain more suitable.

Compared to previous work, PerCoV2 particularly excels at ultra-low to extreme bit-rates ($0.0015 - 0.025$~bpp), delivering consistent improvements in statistical fidelity over strong baselines (see~\cref{fig:teaser,fig:vis_impressions,fig:vis_impressions_2,fig:vis_impressions_2_ext1,fig:vis_impressions_2_ext2}).

In summary, our contributions are as follows:

\begin{itemize}

\item We present PerCoV2, an ultra-low bit-rate perceptual image compression framework that builds upon the fully open SANA foundation model and integrates semantic tokenization, flow-based generation, and learned entropy coding within a unified architecture.

\item We propose a resolution-adaptive 1D query-based tokenizer that produces compact semantic image tokens with a dual role in flow matching: providing a data-dependent reconstruction prior for initialization and a conditioning signal for generative refinement. This decoupling of semantic representation and perceptual generation simplifies the flow-based learning objective, resulting in more stable optimization and improved perceptual compression performance.

\item We introduce a dedicated 1D masked entropy model to improve rate efficiency and an optional decoder-side multimodal enhancement stage based on Molmo to improve perceptual quality without additional transmitted bits.

\item We demonstrate state-of-the-art statistical fidelity, measured by FID and KID, on the official PerCoV1 evaluation setting (MSCOCO-30k at 512$\times$512), while achieving strong zero-shot generalization to native high-resolution benchmarks (DIV2K and CLIC 2020) when trained solely on SA-1B.

\item We introduce PerCoV2-distilled, a practical single-step variant that retains the perceptual compression performance of the multi-step model while accelerating decoding by $5.37\times$ and $5.16\times$ compared with PerCoV1 and PerCoV2, respectively.

\end{itemize}


\section{Related Work}\label{sec:related_work}

\textbf{Generative and Foundation Models.} 
Generative models serve as powerful priors for reconstructing visually plausible images from highly underspecified signals, forming the foundation of modern text-to-image systems such as Stable Diffusion~\citep{Rombach_2022_CVPR, esser2024scaling}, SANA~\citep{xie2025sana}, and FLUX.2\footnote{https://huggingface.co/black-forest-labs/FLUX.2-dev}. Early approaches were dominated by adversarial generative models, e.g., GigaGAN~\citep{NIPS2014_5ca3e9b1,kang2023gigagan}, while recent advances have shifted toward iterative refinement paradigms, most notably diffusion models~\citep{pmlr-v37-sohl-dickstein15, NEURIPS2020_4c5bcfec} and flow-based generative models~\citep{lipman2023flow, liu2023flow, albergo2023building}. Latent diffusion models~\citep{Rombach_2022_CVPR} further enable scalable training by operating in compressed latent spaces.

In parallel, autoregressive approaches have demonstrated competitive performance and are emerging as viable alternatives~\citep{tian2024visual, Infinity}. Building on these developments, foundation models~\citep{Bommasani2021FoundationModels_short} trained on large-scale multimodal datasets exhibit strong generalization across generative tasks~\citep{esser2024scaling, pmlr-v202-li23q, pmlr-v139-radford21a}, making them particularly attractive as priors for perceptual image compression.

\textbf{Perceptual Image Compression.}
Early work in learned image compression demonstrated that neural codecs can outperform traditional approaches~\citep{theis2017lossy, balle2017endtoend}, motivating extensive research into improved entropy models~\citep{ballé2018variational, 9190935, He_2021_CVPR, He_2022_CVPR} and network architectures~\citep{zhu2022transformerbased, He_2022_CVPR, 10222381}. The rate--distortion--perception (RDP) trade-off formalizes the fundamental relationship between reconstruction distortion and perception, where perception is defined as the statistical discrepancy between the distribution of reconstructed images and the natural image distribution~\citep{ICML-2019-BlauM}. In this work, we refer to this notion as statistical fidelity to distinguish it from subjective perceptual quality. Motivated by this trade-off, subsequent methods adopted GANs~\citep{Agustsson_2019_ICCV, NEURIPS2018_801fd8c2, mentzer2020high, pmlr-v202-muckley23a, 10.1007/978-3-031-72761-0_12} and diffusion models~\citep{NEURIPS2023_ccf6d8b4, Hoogeboom2023a, ghouse2023neural}, enabling more realistic reconstructions, particularly at low bit-rates.

VQ-VAE~\citep{NIPS2017_7a98af17} inspired approaches have also shown strong performance~\citep{el-nouby2023image, Mentzer_2023_ICCV, Jia_2024_CVPR, mao2024extreme, xue2024unifying}. Of particular relevance is TiTok~\citep{yu2024an}, which introduces 1D image tokenizers that reduce spatial redundancy through query-based tokenization, leading to improved compression efficiency~\citep{xue2025dlf, liu_2025, Sargent_2025_ICCV}. Nevertheless, many existing approaches rely on high-dimensional latent grids, which can be inefficient in representing compact visual information at ultra-low bit-rates.

\textbf{Foundation Models as Priors for Compression.}
Recent work has explored pretrained foundation models as strong generative priors for neural image compression~\citep{pan_2022, lei2023text+sketch, careil2024towards, li2024misc, li2024towards, xia2025diffpc, Ke2025resulic}, differing primarily in conditioning modalities. Proposed signals include prompt inversion and compressed sketches~\citep{wen2023hard, lei2023text+sketch}, text descriptions generated by LLMs combined with semantic features~\citep{careil2024towards, li2024misc, xia2025diffpc, Ke2025resulic}, CLIP-derived image embeddings~\citep{pmlr-v139-radford21a, bachard:hal-04478601}, as well as textual inversion paired with classifier-free guidance~\citep{gal2023an, pan_2022}.

Most prior approaches rely on SD-based backbones~\citep{Rombach_2022_CVPR}, often via extensions such as ControlNet~\citep{Zhang_2023_ICCV}, DiffBIR~\citep{lin2024diffbir}, or Stable unCLIP, while keeping pretrained weights fixed. In contrast, our approach builds directly on modern flow-based foundation models and employs highly compact \emph{multimodal image tokens} that jointly encode semantic, structural, and appearance information.

Recent studies further show that diffusion models can be distilled into few-step or single-step generators while maintaining competitive performance~\citep{Zhang_2025_ICCV, Zhang_2026_CVPR, li2025rdeic, xue2025one, guo2025oscar, jia2026codlite, Shi_2026_CVPR}, enabling more practical deployment. We refer readers to the respective literature for a detailed discussion of generative model distillation techniques, which are beyond the scope of this work. Nevertheless, for completeness, we provide a distilled PerCoV2 variant based on Distribution Matching Distillation (DMD)~\citep{Yin_2024_CVPR,3737916.3739421}, following recent compression-oriented distillation approaches~\citep{xue2025one, jia2025cod, jia2026codlite}.

Finally, a promising direction is diffusion-based reverse channel coding~\citep{NEURIPS2020_4c5bcfec, theis2023lossy}. Recent work such as DiffC~\citep{vonderfecht2025lossy} demonstrates an early practical prototype in this direction, while CoD~\citep{jia2025cod} introduces a compression-oriented diffusion foundation model trained specifically for downstream diffusion codecs. We compare directly against both approaches in our experiments.


\section{Background}\label{sec:background}

\textbf{Neural Image Compression.} Neural image compression uses deep learning techniques to learn compact image representations. This is typically achieved by an auto-encoder-like structure, consisting of an image encoder $y=E(x)$, a decoder $x'=D(y)$ that operates on the quantized latent representation $y$, and an entropy model $P(y)$. The learning objective is to minimize the rate-distortion trade-off~\citep{cover2012elements}, with $\lambda > 0$:

\begin{equation}\label{eq:rd_objective}
\mathcal{L}_{RD}=\mathop{\mathbb{E}_{x\sim p_X}}[\lambda r(y) + d(x, x')].
\end{equation}

In~\cref{eq:rd_objective}, $d(x, x')$ captures the distance of the reconstruction $x'$ to the original input image $x$, whereas the bit-rate is estimated using the cross entropy $r(y)=-\log{P(y)}$. In practice, an entropy coding method based on the probability model $P$ is used to obtain the final bit representation, see~\citep{CGV-107} for a more general overview.

\textbf{Flow Models.} 
Flow models~\citep{lipman2023flow, liu2023flow, albergo2023building} are a popular choice for generative modeling. A flow, $\phi : [0, 1] \times \mathbb{R}^d \to \mathbb{R}^d$ is a time-dependent function that characterizes the transition from a (simple) prior distribution $p_0$ to a target distribution $q$ via the ordinary differential equation (ODE):

\begin{equation}\label{eq:flow_ode}
\frac{d}{dt} \phi_t(x) = v_t(\phi_t(x)), \quad \phi_0(x) = x;
\end{equation}

where $v : [0, 1] \times \mathbb{R}^d \to \mathbb{R}^d$ is a time-dependent vector field. Assuming we have access to a target vector field $u_t$ and its corresponding probability density path $p: [0, 1] \times \mathbb{R}^d \to \mathbb{R}_{>0}$, the flow matching objective is given by:

\begin{equation}\label{eq:flow_matching}
\mathcal{L}_{\text{FM}}(\theta) = \mathbb{E}_{t, p_t(x)} \left\| v_t(x) - u_t(x) \right\|^2,
\end{equation}

which boils down to directly regressing the vector field $u_t$ via a neural network $v_t(x, \theta)$. During inference, an ODE solver can then be used to sample from the generative model. In practice, however, we do not have access to $u_t$ in closed form that generates $p_t$.

A tractable yet equivalent learning objective is the conditional flow matching objective ~\citep{lipman2023flow}:

\begin{equation}\label{eq:conditional_flow_matching}
\mathcal{L}_{\text{CFM}}(\theta) = \mathbb{E}_{t, q(x_1), p_t(x|x_1)} \left\| v_t(x) - u_t(x|x_1) \right\|^2,
\end{equation}

which defines a conditional probability path $p_t$ and a conditional vector field $u_t$ per sample $x_1 \sim q(x_1)$. 

For practical applications, $p_t$ typically takes the form:

\begin{equation}\label{eq:pt}
p_t(x|x_1) = \mathcal{N}(x \,|\, \mu_t(x_1), \sigma_t(x_1)^2 I),
\end{equation}

with $\mu_t(x) = t x_1$ and $\sigma_t(x) = 1 - (1 - \sigma_{\text{min}})t$, such that for $p_0$, we get a standard Gaussian distribution; for $p_1$, a Gaussian distribution concentrated around $x_1$, while for all other $p_t$ the mean and standard deviation simply change linearly in time. The corresponding conditional vector field $u_t$ and conditional flow $\phi_t(x)$ are given by:

\begin{equation}\label{eq:ut}
u_t(x|x_1) = \frac{x_1 - (1 - \sigma_{\text{min}}) x}{1 - (1 - \sigma_{\text{min}}) t},
\end{equation}

\begin{equation}\label{eq:psi_t}
\phi_t(x) = \big(1 - (1 - \sigma_{\text{min}}) t\big)x + t x_1.
\end{equation}

The final objective is obtained by reparameterizing $p_t(x|x_1)$ in terms of $x_0$:

\begin{align}\label{eq:conditional_flow_matching_extended}
&\mathcal{L}_{\text{CFM}}(\theta) = \mathbb{E}_{t, q(x_1), p(x_0)} \notag \\
&\quad \left\| v_t(\phi_t(x_0)) - \left( x_1 - (1 - \sigma_{\text{min}}) x_0 \right) \right\|^2.
\end{align}


\section{On Latent Representation and Backbone Scale in PerCoV1} \label{sec:percov1}

\begin{figure*}[tb]
  \centering
  \includegraphics[width=1\linewidth]{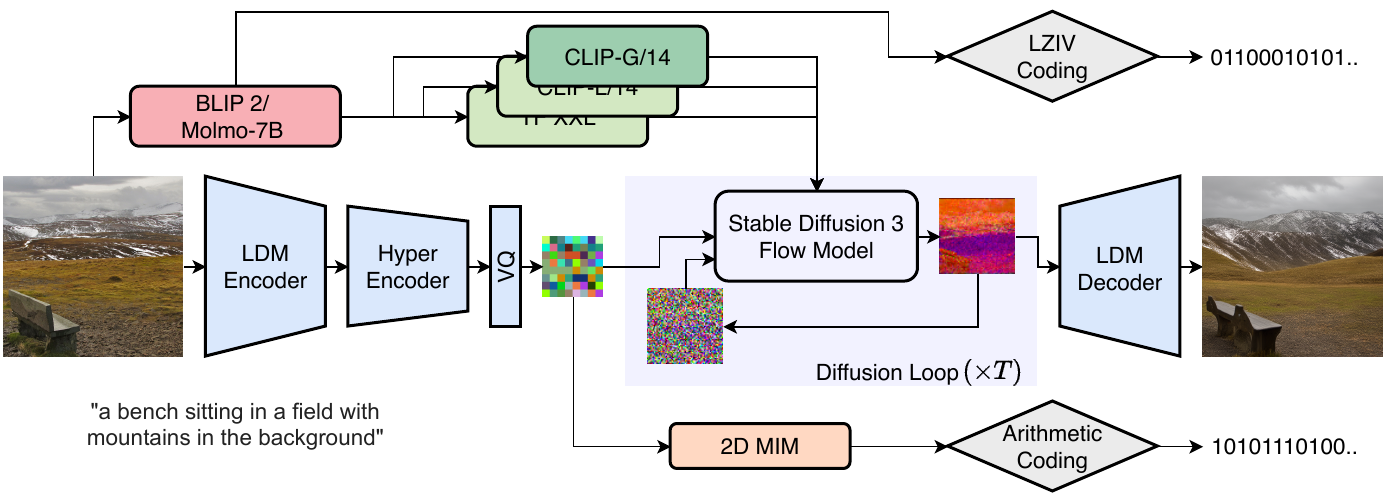}
  \caption{PerCoV1++ (SD v3) architecture overview
  }
  \label{fig:perco_overview}
\end{figure*}

\begin{table}[tb]
\centering
\caption{Comparison of generative backbones in terms of AE compression design and model size. F denotes the spatial downsampling factor, C the number of latent channels, and P2 indicates $2\times2$ latent patchification.}
\label{tab:backbone_compare}
\vspace{-1mm}
\setlength{\tabcolsep}{6pt}
\renewcommand{\arraystretch}{1.0}
\begin{tabular}{lcccc}
\toprule
Backbone & AE & Model Size & Open Source \\
\midrule
GLIDE        & F8C4 & $1.4$B  & \xmark \\
SD v2.1      & F8C4 & $866$M  & \cmark \\
SD v3.0      & F8C16-P2 & $2$B    & \cmark \\
FLUX.2      & F8C32-P2  & $4$B    & \cmark \\
SANA 1.5     & F32C32  & $1.6$B  & \cmark \\
\bottomrule
\end{tabular}
\vspace{-3mm}
\end{table}

\textbf{Background.} PerCoV1~\citep{careil2024towards} formulates perceptual compression as a conditional generative coding scheme, where an image $x$ is mapped by an LDM encoder/ hyper-encoder pipeline into a compact latent representation. This latent, together with an image caption, is transmitted and used to condition a pretrained generative backbone (GLIDE~\citep{pmlr-v162-nichol22a}) that performs the final reconstruction.

This formulation is inherently backbone-agnostic and can be combined with different conditional generative models, e.g., GLIDE~\citep{pmlr-v162-nichol22a} as done in PerCoV1, or any other diffusion- and flow-based architectures such as Stable Diffusion v2.1/ v3.0~\citep{Rombach_2022_CVPR, esser2024scaling}, SANA~\citep{xie2025sana}, and FLUX.2. It also naturally extends to alternative latent parameterizations beyond standard continuous feature maps, including discretized or structured representations. We illustrate this unified architecture in \cref{fig:perco_overview}, using Stable Diffusion 3 as a representative backbone and extending it with a masked image modeling component similar to PQ-MIM~\citep{el-nouby2023image}.

\begin{figure}[tb]
  \centering
  \includegraphics[width=1\linewidth]{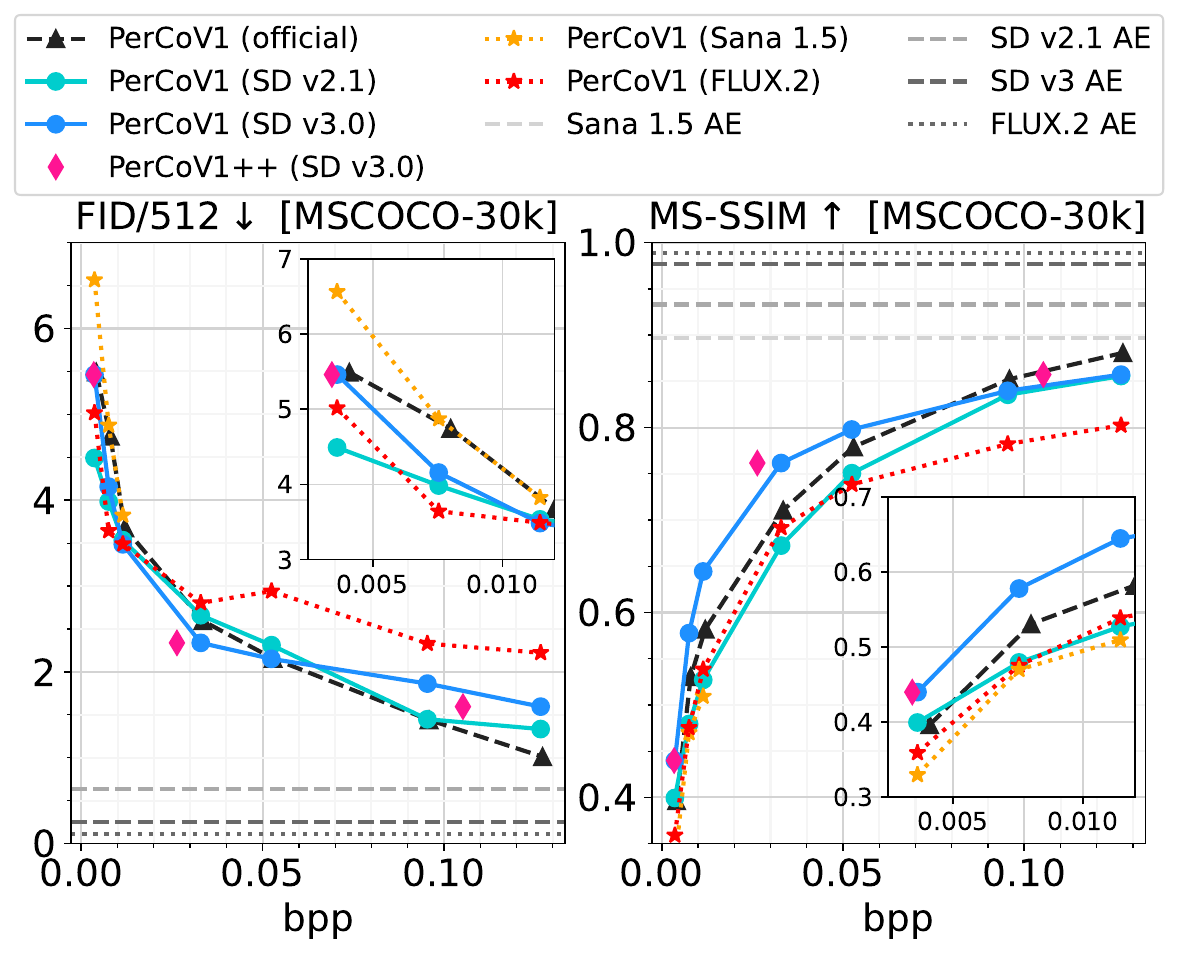}
  \caption{PerCoV1 preliminary investigation
  }
  \label{fig:preliminary}
\end{figure}

\textbf{Preliminary Study/ Insights.} During our preliminary investigations, we observed a non-linear relationship between generative backbone capacity and perceptual compression performance (\cref{tab:backbone_compare}, \cref{fig:preliminary}). At higher bit-rates, scaling to larger architectures such as Stable Diffusion 3 (2B) yields diminishing returns compared to more compact backbones such as Stable Diffusion v2.1 (866M), which are often simpler to optimize in practice\footnote{These preliminary comparisons use different optimization budgets due to computational and hardware constraints. SD-based models follow the PerCo setup and are trained for up to $150$k steps with a global batch size of $160$ (vs.\ $300$k steps in the original PerCo setting), whereas SANA and FLUX.2 use $50$k steps with a batch size of $40$ and $20$, respectively. Larger configurations typically lead to improved distortion metrics, while perceptual metrics remain stable.}. Notably, Stable Diffusion 3 employs a higher-capacity auto-encoder with $16$ latent channels, whereas Stable Diffusion v2.1 uses only $4$. This observation is consistent with recent findings that improved auto-encoder reconstruction fidelity does not necessarily translate into better generative compression performance~\citep{ram_2024, yao2025vavae}. Earlier auto-encoders were primarily optimized for low perceptual distortion, often preserving high-frequency details that improve reconstruction quality but increase the complexity of downstream generative modeling. In contrast, FLUX.2 incorporates additional semantic regularization of the latent space (e.g., REPA~\citep{yu2025repa}), encouraging representations that better capture high-level image semantics rather than low-level reconstruction details. While we did not observe improvements from these mechanisms in our preliminary experiments, this may reflect suboptimal adaptation to our application domain rather than an inherent limitation of the approach.

Recent architectures such as SANA~\citep{xie2025sana} instead adopt aggressive spatial downsampling up to $32\times$. Compared to conventional $8\times$ latent diffusion auto-encoders, this reduces the number of latent tokens by $16\times$, substantially improving efficiency and scalability. However, this stronger compression inevitably reduces reconstruction fidelity, as reflected in distortion-oriented metrics such as MS-SSIM. We refer to~\citep{bfl2025representation} for a broader discussion of the trade-off between learnability, reconstruction quality, and compression efficiency.

\begin{figure}[t]
\setlength{\tabcolsep}{1pt}
\renewcommand{\arraystretch}{1.0}
\centering

\begin{tabular}{c@{\hspace{1pt}}c@{\hspace{1pt}}c@{\hspace{1pt}}c@{\hspace{1pt}}c}
\raisebox{-0.5\height}{\includegraphics[width=0.29\linewidth]{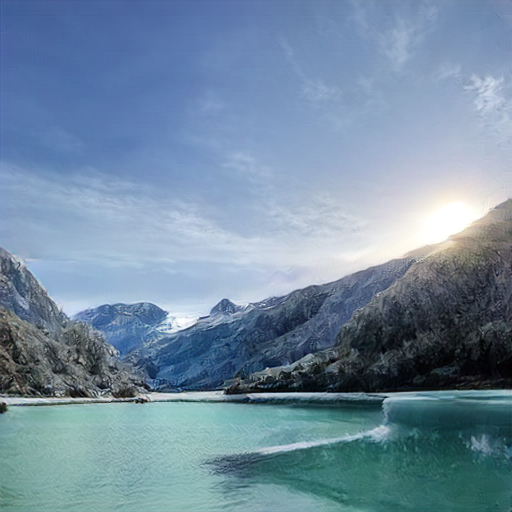}} &
\raisebox{-0.5\height}{\small $+$} &
\raisebox{-0.5\height}{\includegraphics[width=0.29\linewidth]{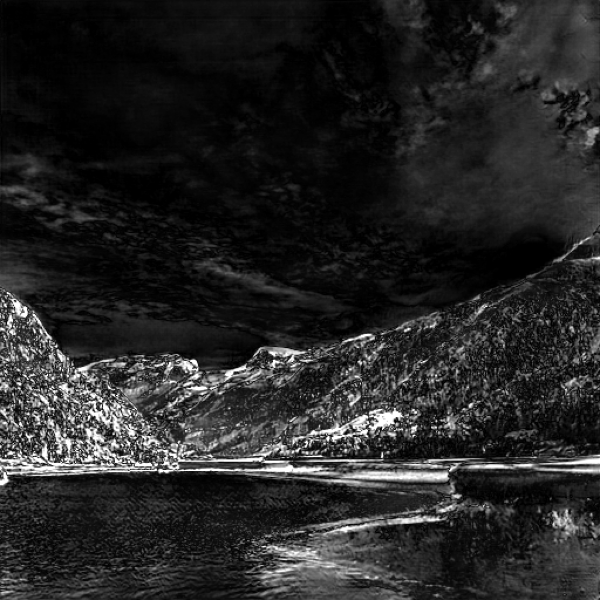}} &
\raisebox{-0.5\height}{\small $=$} &
\raisebox{-0.5\height}{\includegraphics[width=0.29\linewidth]{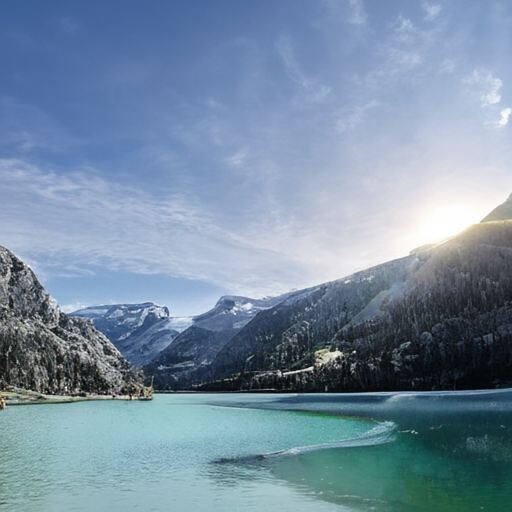}} \\
\end{tabular}

\caption{Tokenizer output (left), implicit residual refinement visualized as a normalized per-pixel difference map (middle), and the final reconstructed image (right). Bright regions indicate stronger refinement.}
\label{fig:impl_residual}
\end{figure}

\textbf{Rethinking Representation Learning for Perceptual Compression}. In PerCoV1, the hyper-latent serves as the primary conditioning signal for the generative decoder and jointly encodes high-level semantic information and denoising guidance. Consequently, reconstruction quality depends strongly on the properties of the underlying auto-encoder and the structure of its latent space. In PerCoV2, we instead simplify this objective through a dual-representation design that explicitly separates global semantic structure from high-frequency detail. Concretely, semantic information is conveyed via a tokenized representation, while the generative backbone focuses on local detail synthesis and refinement (\cref{fig:impl_residual}). This formulation effectively initializes the generative process closer to the semantic manifold of the target distribution, reducing the distributional gap during denoising and the complexity of the underlying flow-matching objective. Importantly, this decoupled formulation enables the use of SANA’s highly efficient architecture while mitigating the distortion performance degradation observed in the PerCoV1 formulation.


\begin{figure*}[tb]
  \centering
  \includegraphics[width=1.0\linewidth]{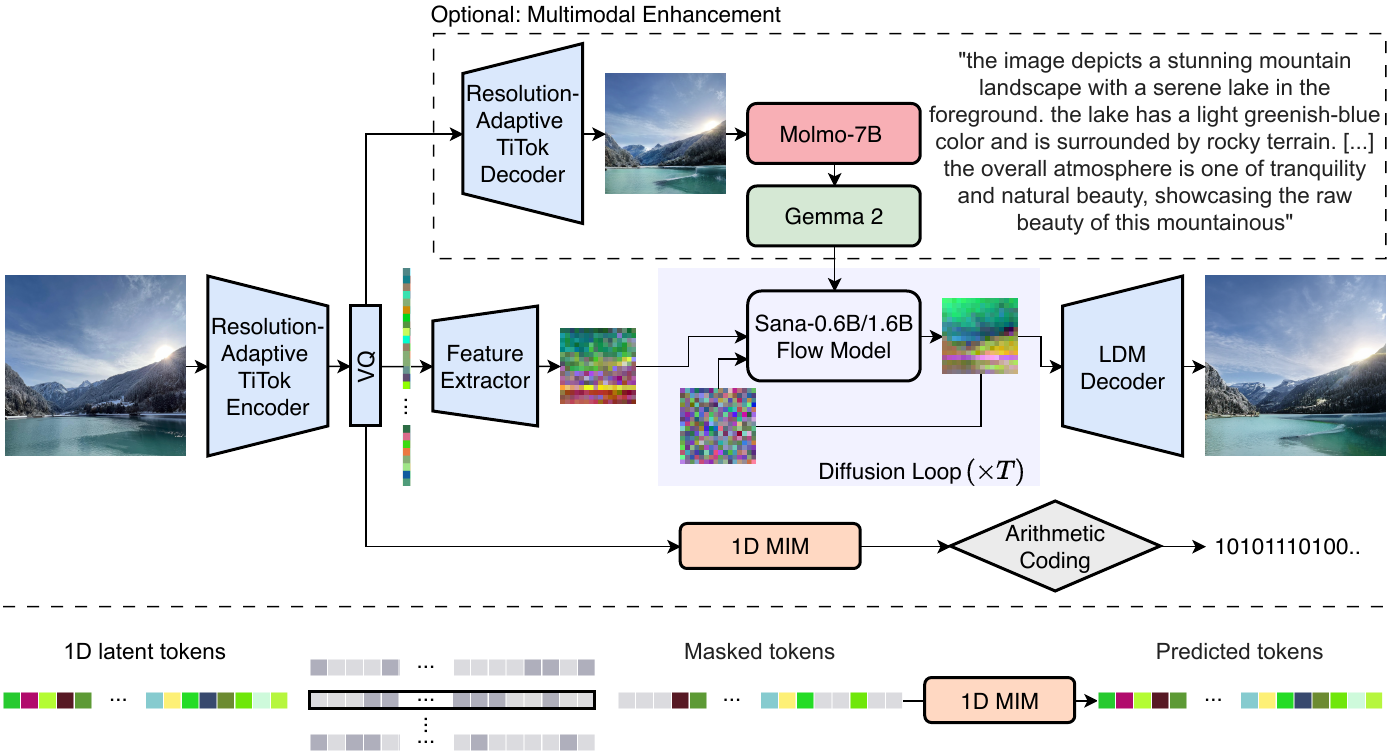}
  \caption{PerCoV2 architecture overview. For visualization clarity, the semantic reconstruction prior used for flow initialization is omitted and the flow process is illustrated from random noise.
  }
  \label{fig:percoV2_overview}
\end{figure*}

\section{Our Method}\label{sec:method}

We present an architectural overview of PerCoV2 in~\cref{fig:percoV2_overview}. PerCoV2 consists of four main components:
(i) a SANA-based generative backbone~\citep{xie2025sana} incorporating an LDM auto-encoder, a Gemma-2 LLM text encoder~\citep{gemma}, and a latent flow model~\citep{liu2023flow, esser2024scaling};
(ii) a resolution-adaptive 1D query-based tokenizer that implements our dual representation for flow matching, producing compact latent tokens that simultaneously serve as the transmitted representation and as a semantic reconstruction prior for initializing and conditioning the latent flow model;
(iii) a 1D masked image model (1D-MIM) that enables adaptive entropy modeling of the discrete latent representation; and
(iv) a Molmo VLM~\citep{DBLP:journals/corr/abs-2409-17146} that provides optional multimodal semantic enhancement during decoding.

\subsection{Open Perceptual Compression}\label{subsec:open_perco_v2}

\noindent\textbf{Encoding and Latent Representation.}
PerCoV2 encodes an input image $\mathbf{x} \in \mathbb{R}^{3 \times H \times W}$ into compact latent tokens $\mathbf{z}_l$ that serve as a semantic reconstruction prior for initializing and conditioning the latent flow model. Unlike conventional tokenizers based on spatial feature grids (e.g., CNNs), our query-based design eliminates spatial redundancy, producing compact image tokens that improve compression efficiency (\cref{subsec:resadatok}). These quantized tokens constitute the only transmitted representation, without requiring image captions or other auxiliary information. To further improve rate efficiency, we introduce adaptive arithmetic coding based on a 1D masked image model (\cref{subsec:1dmim}), replacing the uniform coding strategy of PerCoV1 with a learned entropy model that more accurately captures the latent token distribution.

\noindent\textbf{Decoding.}
During decoding, the latent representation $\mathbf{z}_l$ is first recovered through arithmetic decoding and then processed by a resolution-adaptive 1D tokenizer decoder, which reconstructs the semantic reconstruction prior used to initialize the generative process. The decoded semantic representation is further transformed by a resolution-adaptive feature extractor into the conditioning signal for latent refinement. Specifically, the feature extractor maps the 1D latent sequence back into a spatial feature map $\mathbf{f} \in \mathbb{R}^{320 \times \frac{H}{32} \times \frac{W}{32}}$, matching the spatial resolution of the flow model's latent space. The resulting conditioning features are concatenated with the noised latents along the channel dimension to provide semantic guidance for the generative reconstruction process (see~\cref{fig:percoV2_overview}).

Beyond latent-only reconstruction, PerCoV2 optionally supports multimodal enhancement (\cref{subsec:multimodalenhancement}) by incorporating a Vision-Language Model (VLM) to provide additional high-level semantic priors. This refinement stage leverages the semantic understanding of the VLM to guide the generative flow without requiring additional transmitted bits for textual information.

Finally, the generative flow model performs the reverse flow process to synthesize the final high-fidelity reconstruction.

\subsection{Resolution-Adaptive 1D Tokenization}\label{subsec:resadatok}

\begin{figure*}[tb]
  \centering
  \includegraphics[width=1.0\linewidth]{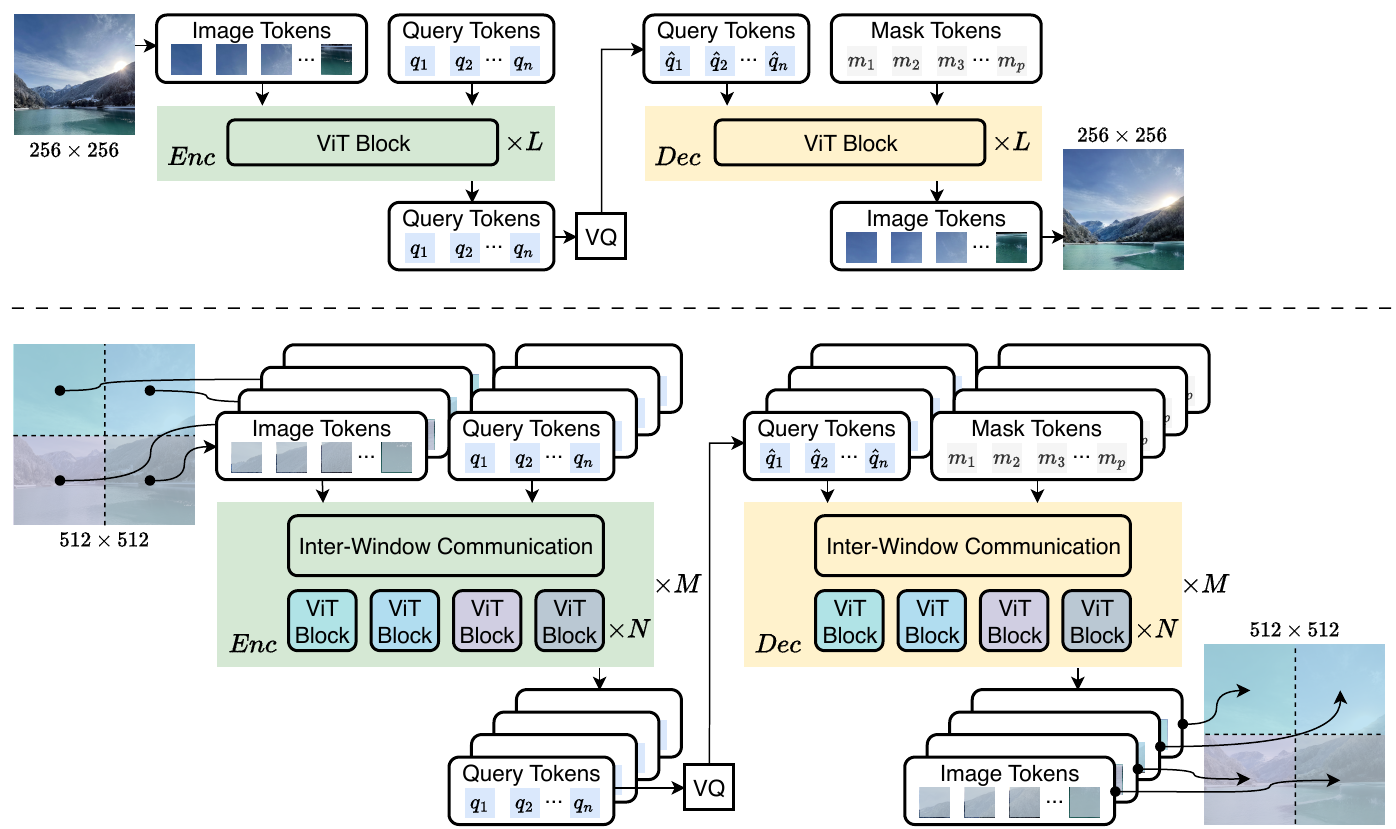}
  \caption{Fixed-resolution TiTok (top) vs.\ resolution-adaptive TiTok (bottom)
  }
  \label{fig:resadatitok_overview}
\end{figure*}

Our tokenizer extends TiTok~\citep{yu2024an} to support resolution-adaptive image compression while preserving the compact 1D latent representation and local tokenization prior of the original tokenizer. An overview of the proposed architecture is shown in~\cref{fig:resadatitok_overview}. 

To process an input image $\mathbf{x}\in\mathbb{R}^{3\times H\times W}$, we partition it into non-overlapping $256\times256$ windows, following recent high-resolution vision architectures~\citep{xue2025dlf, liu_2025}. Each window is tokenized into 256 visual embeddings with dimension $D \in \{512, 768, 1024\}$, prepended with a class token, and augmented with $Q$ learnable query tokens $\mathbf{L}\in\mathbb{R}^{Q\times D}$, resulting in a sequence of $257+Q$ tokens per window. We retain standard Vision Transformer (ViT) blocks~\citep{dosovitskiy2021an} for intra-window modeling, while our main architectural contribution lies in the inter-window communication modules, which enable information exchange across windows.

\begin{figure}[tb]
  \centering
  \includegraphics[width=1.0\linewidth]{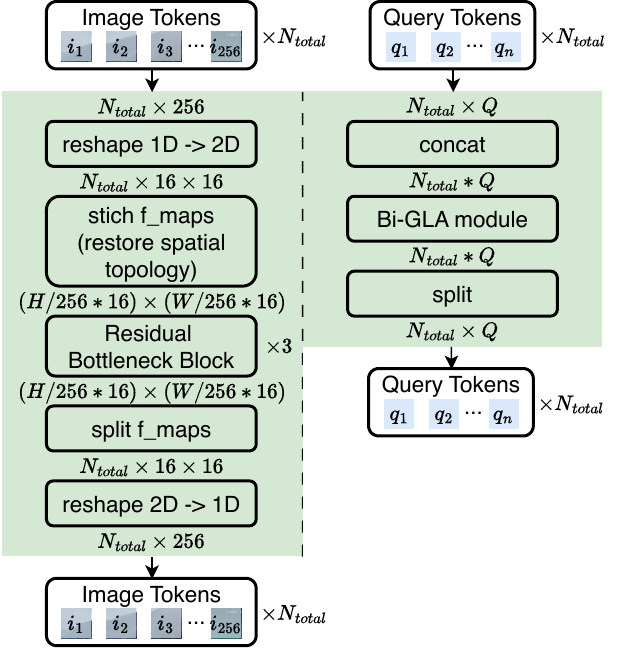}
  \caption{Inter-window communication
  }
  \label{fig:interwindow}
\end{figure}

Unlike~\citep{liu_2025}, which performs global modeling by directly concatenating all tokens into a single sequence, we explicitly separate image tokens and query tokens before inter-window communication (see~\cref{fig:interwindow}). The query tokens provide a compact representation of global image structure and are therefore used for cross-window information exchange, while avoiding the cost of globally processing all visual tokens.

Specifically, the query tokens from different windows are aggregated into a global sequence and processed using Bidirectional Gated Linear Attention (Bi-GLA)~\citep{yang2024gla}, enabling efficient information exchange across the image. To further improve efficiency and practical deployment, we implement Bi-GLA using Triton-based Flash Linear Attention kernels~\citep{yang2024fla}, replacing the Bi-RWKV modules adopted in~\citep{liu_2025}. This implementation preserves linear computational complexity while providing platform-agnostic hardware acceleration and avoiding portability limitations associated with specialized CUDA kernels.

The image tokens follow a complementary spatial refinement pathway. Image tokens from all windows are rearranged and tiled back into their original two-dimensional layout, followed by a stack of residual bottleneck blocks. Although they could be processed directly in the flattened 1D representation using linear attention modules, we find that restoring their spatial layout is beneficial for image reconstruction. In particular, it re-aligns spatially adjacent tokens that become distant in the flattened 1D sequence, facilitating the modeling of long-range dependencies (cf.~\cref{fig:erf_vis}). After refinement, the features are converted back into the original 1D token arrangement.

\begin{figure}[tb]
    \setlength{\tabcolsep}{1.0pt}  
    \renewcommand{\arraystretch}{1.0}  
    \centering
    \scriptsize
    \begin{tabular}{cccc}
        \toprule
        \multicolumn{4}{c}{PerCoV1 hyper-latent (8$\times$8 2D grid tokens)} \\
         \multicolumn{4}{c}{\citep{careil2024towards}} \\
        \midrule
        \texttt{loc=(1,1)} & \texttt{loc=(1,3)} & \texttt{loc=(1,5)} & \texttt{loc=(1,7)} \\
        \begin{subfigure}{0.11\textwidth}
            \centering
            \includegraphics[width=\linewidth]{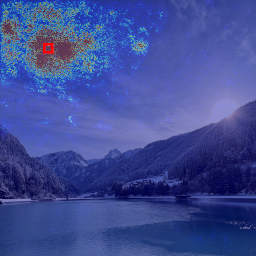}
        \end{subfigure}
        &
        \begin{subfigure}{0.11\textwidth}
            \centering
            \includegraphics[width=\linewidth]{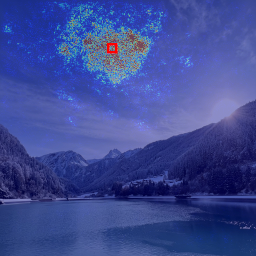}
        \end{subfigure}
        &
        \begin{subfigure}{0.11\textwidth}
            \centering
            \includegraphics[width=\linewidth]{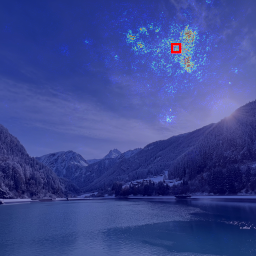}
        \end{subfigure}
        &
        \begin{subfigure}{0.11\textwidth}
            \centering
            \includegraphics[width=\linewidth]{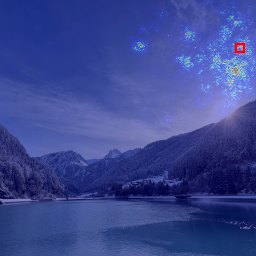}
        \end{subfigure}
        \\
        \texttt{loc=(3,1)} & \texttt{loc=(3,3)} & \texttt{loc=(3,5)} & \texttt{loc=(3,7)} \\
         \begin{subfigure}{0.11\textwidth}
            \centering
            \includegraphics[width=\linewidth]{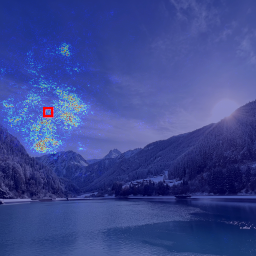}
        \end{subfigure}
        &
        \begin{subfigure}{0.11\textwidth}
            \centering
            \includegraphics[width=\linewidth]{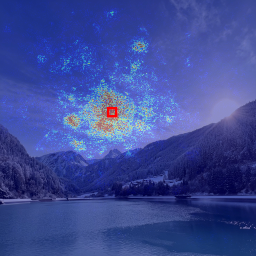}
        \end{subfigure}
        &
        \begin{subfigure}{0.11\textwidth}
            \centering
            \includegraphics[width=\linewidth]{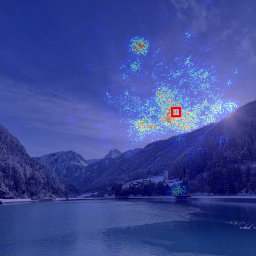}
        \end{subfigure}
        &
        \begin{subfigure}{0.11\textwidth}
            \centering
            \includegraphics[width=\linewidth]{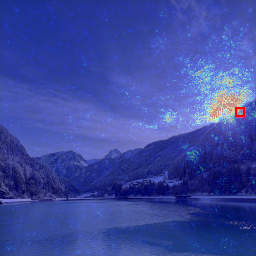}
        \end{subfigure}
        \\
        \midrule
        \multicolumn{4}{c}{PerCoV2 latent ($64$ 1D query tokens)} \\
         \multicolumn{4}{c}{Ours/ derived from TiTok~\citep{yu2024an}} \\
        \midrule
        \texttt{loc=1} & \texttt{loc=4} & \texttt{loc=10} & \texttt{loc=11} \\
        \begin{subfigure}{0.11\textwidth}
            \centering
            \includegraphics[width=\linewidth]{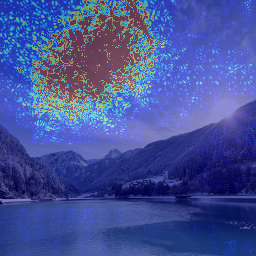}
        \end{subfigure}
        &
        \begin{subfigure}{0.11\textwidth}
            \centering
            \includegraphics[width=\linewidth]{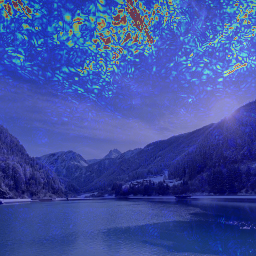}
        \end{subfigure}
        &
        \begin{subfigure}{0.11\textwidth}
            \centering
            \includegraphics[width=\linewidth]{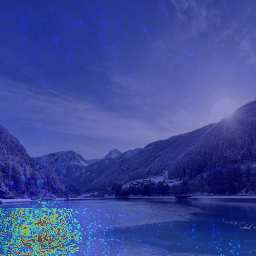}
        \end{subfigure}
        &
        \begin{subfigure}{0.11\textwidth}
            \centering
            \includegraphics[width=\linewidth]{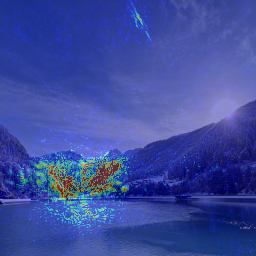}
        \end{subfigure}
        \\
        \texttt{loc=48} & \texttt{loc=51} & \texttt{loc=55} & \texttt{loc=60} \\
         \begin{subfigure}{0.11\textwidth}
            \centering
            \includegraphics[width=\linewidth]{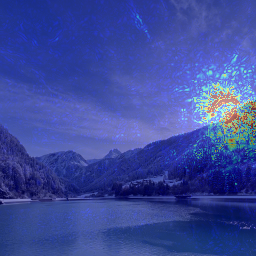}
        \end{subfigure}
        &
        \begin{subfigure}{0.11\textwidth}
            \centering
            \includegraphics[width=\linewidth]{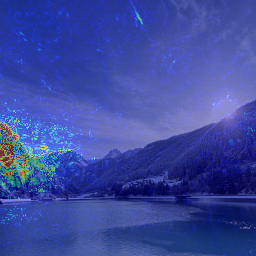}
        \end{subfigure}
        &
        \begin{subfigure}{0.11\textwidth}
            \centering
            \includegraphics[width=\linewidth]{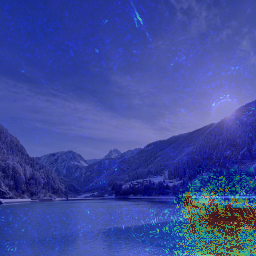}
        \end{subfigure}
        &
        \begin{subfigure}{0.11\textwidth}
            \centering
            \includegraphics[width=\linewidth]{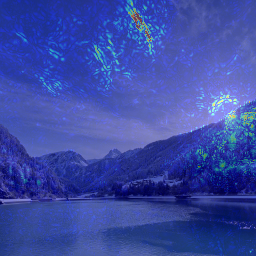}
        \end{subfigure}
        \\
        \bottomrule

    \end{tabular}

    \caption{Comparison of encoder representations between PerCoV1 (top) and PerCoV2 (bottom) through token-wise effective receptive field (ERF) visualization. The ERFs are computed from the continuous latent representations before vector quantization. \textbf{Best viewed electronically.}}

    \label{fig:spatial_vs_query}
\end{figure}

Adhering to the 1D tokenization principle, the encoder output is pruned to retain only the transformed query tokens. This yields a compact latent representation $\mathbf{z}_l$ whose length is decoupled from the pixel-space grid. Specifically, each $256 \times 256$ region is compressed into $Q \in \{32, 64, 128\}$ tokens, resulting in a total sequence length of $N_{\text{total}} = \frac{H}{256} \times \frac{W}{256} \times Q$. These tokens are subsequently quantized against a codebook of size $|\mathcal{V}| = 4096$, producing discrete tokens for subsequent entropy modeling and transmission. Unlike conventional 2D spatial grid representations, the proposed 1D formulation removes the requirement for a fixed spatial token layout, allowing the latent representation to capture semantically meaningful image content more effectively. We illustrate this in \cref{fig:spatial_vs_query} through token-wise effective receptive field (ERF) visualizations~\citep{3157382.3157645} of the encoder representations learned by PerCoV1~\citep{careil2024towards, koerber2024perco} and PerCoV2. While the 2D hyper-latent tokens of PerCoV1 predominantly exhibit localized receptive fields that follow the underlying pixel grid, the query-based latent tokens of PerCoV2 aggregate information from semantically related image regions, such as the sky, water, mountains, or the sun and its rays, yielding a more expressive representation for ultra-low bit-rate compression.

Finally, the de-tokenization stage employs a symmetric decoding architecture. To recover the spatial organization, we introduce a sequence of mask tokens $\mathbf{M}$ by replicating a learnable mask token over the $H/16\times W/16$ spatial grid~\citep{yu2024an}. These mask tokens are concatenated with the quantized latent tokens $\mathbf{z}_l$ and processed by mirrored ViT and inter-window communication blocks, which map the 1D representation back into the pixel-space or LDM feature manifold.

\subsection{1D Masked Image Model (1D MIM)}\label{subsec:1dmim}

A straightforward strategy for transmitting vector-quantized (VQ) latent tokens is uniform coding, which assumes that the codebook indices are independently and uniformly distributed. Under this assumption, the bit-rate $R$ (in bits per pixel) is given by:
\begin{equation}  
R(\mathbf{z}_l) = \frac{N_{\text{total}} \log_2 |\mathcal{V}|}{HW}.  
\end{equation}  
However, this assumption is rarely satisfied in practice, as latent representations typically contain substantial spatial and semantic redundancy. Ignoring these dependencies therefore leads to suboptimal compression efficiency~\citep{el-nouby2023image, Jia_2024_CVPR}.

To exploit the inherent structure of the latent representation, PerCoV2 introduces a Masked Image Model (MIM)~\citep{chang2022maskgit, el-nouby2023image, Mentzer_2023_ICCV} specifically designed for 1D latent sequences (1D MIM, \cref{fig:percoV2_overview}). By factorizing the entropy model into conditional distributions over $K$ token subsets, the joint probability can be written as:
\begin{equation}
    p(q) = \prod_{k=1}^{K} p(q_k \mid \mathcal{C}_k),
\end{equation}
where $q=(q_1,\dots,q_K)$ denotes the partitioned 1D latent sequence and $\mathcal{C}_k$ represents the accumulated context available at decoding step $k$. Unlike autoregressive entropy models that require strictly sequential decoding, MIM enables parallel prediction over token subsets. During training, the 1D MIM learns to recover randomly masked tokens, allowing it to capture diverse conditional distributions corresponding to different masking configurations. During inference, we adopt a Quantized Low-Discrepancy Sequence (QLDS)~\citep{Mentzer_2023_ICCV}, adapted to the 1D setting, to define a deterministic masking schedule shared between the encoder and decoder for synchronized entropy coding.

Architecturally, the 1D MIM follows the same resolution-adaptive design as the TiTok decoder (\cref{subsec:resadatok}). The input sequence is first mapped back into a spatial layout using learned grid mask replication and then processed by interleaved ViT and inter-window communication blocks. Unlike the tokenizer decoder, which uses the reconstructed features for image generation, the 1D MIM extracts the transformed query tokens and projects them into the codebook dimension $|\mathcal{V}|$ to predict the latent token probabilities. This design enables initialization from the pre-trained tokenizer decoder, substantially accelerating optimization while ensuring that the entropy model remains aligned with the reconstruction manifold.

\subsection{Multimodal Enhancement}\label{subsec:multimodalenhancement}

\begin{figure}[tb]
  \centering
  \includegraphics[width=1\linewidth]{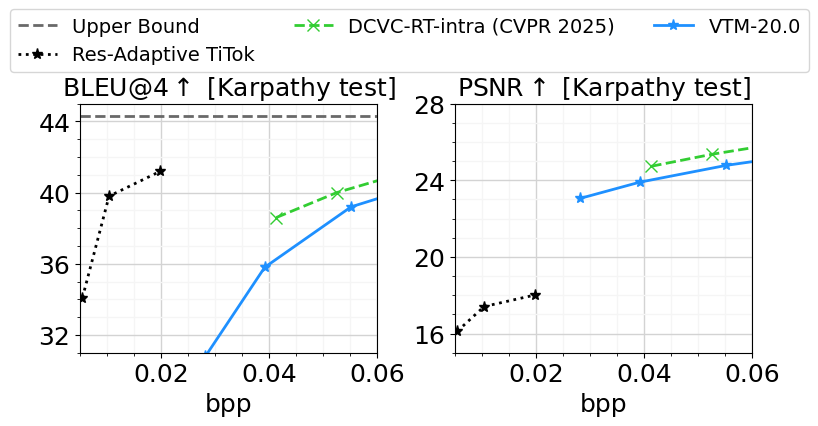}
  \caption{Performance of BLIP-2 (OPT-2.7B) on the Karpathy test set under corruptions
  }
  \label{fig:multimodal_motivation}
\end{figure}

PerCoV2 optionally enables multimodal image enhancement by integrating a Vision-Language Model (VLM), specifically Molmo~\citep{DBLP:journals/corr/abs-2409-17146}, into the decoding pipeline. This stage is not required for reconstruction and is designed as an optional post-processing enhancement, as VLM-based semantic inference introduces additional computational overhead. Given the decoded latent representation, we first employ the resolution-adaptive 1D tokenizer decoder to obtain a base reconstruction. Unlike conventional fixed-grid representations, which may achieve higher PSNR under aggressive compression at the cost of perceptual and semantic quality, our query-based tokenizer preserves meaningful image information at low bit-rates, enabling reconstructed images to remain reliable inputs for visual understanding. We demonstrate this property in~\cref{fig:multimodal_motivation} by evaluating the captioning robustness of compressed reconstructions using BLIP-2~\citep{pmlr-v202-li23q} under severe compression artifacts.

The reconstructed image is provided to Molmo, which generates a descriptive caption $\mathbf{z}_g$ using the prompt: \textit{``Briefly describe the image.''} The resulting textual representation is mapped into the semantic embedding space of the Gemma-2 LLM~\citep{gemma} and incorporated into the flow model through cross-attention layers~\citep{esser2024scaling}. Unlike approaches that explicitly transmit textual descriptions as part of the bitstream, our approach treats the generated caption as an optional post-processing enhancement, enabling richer semantic guidance without increasing the communication cost.

\subsection{Optimization}

Our optimization follows a modular three-stage framework: we first train the resolution-adaptive 1D tokenizer, followed by 1D MIM entropy modeling, and finally perform end-to-end refinement of the full PerCoV2 system.

\textbf{Resolution-adaptive 1D tokenizer.} We adopt the two-stage training paradigm from TiTok~\citep{yu2024an} to attain stable convergence within the compact 1D latent space. To accelerate optimization, we initialize our weights from pre-trained TiTok models (fixed $256\times256$) and perform progressive fine-tuning at $512\times512$ resolution to establish global dependencies. In the first stage, the model is trained via proxy-code supervision, learning to predict discrete indices from an off-the-shelf VQGAN~\citep{Esser_2021_CVPR, chang2022maskgit}. This ``warm-up'' phase bypasses unstable adversarial losses to focus optimization on the latent bottleneck. Subsequently, we freeze the encoder and quantizer to refine only the decoder toward the pixel space. This second stage employs the standard VQGAN objective~\citep{Esser_2021_CVPR}, leveraging perceptual and adversarial losses to maximize final reconstruction fidelity.

\textbf{1D MIM.} For entropy modeling, we follow~\citep{chang2022maskgit}. During training, masked tokens are replaced with a learnable mask embedding. The model is optimized using the standard cross-entropy loss on the masked tokens with a label smoothing factor of $0.1$. Additionally, we apply a fractional weight of $0.1$ to the loss on unmasked tokens to maintain global representation stability. We initialize our models with pre-trained weights from our resolution-adaptive TiTok decoder to accelerate convergence. The resulting MIM can then be employed for both compression and generation (cf.~\cref{fig:beyond_compression_tokenizer}).

\textbf{PerCoV2.} Finally, we fine-tune PerCoV2 using the conditional flow matching objective~\cref{eq:conditional_flow_matching_extended}, extended by $z=(z_l, z_g)$. Inspired by the coupling strategy in~\citep{10.1007/978-3-031-73030-6_19}, we move beyond modeling the standard transition from Gaussian noise to data; instead, we find it beneficial to replace the noise source with an initial base reconstruction distribution $p(x_0)$. In practice, $x_0$ is obtained by projecting the output of our resolution-adaptive 1D tokenizer decoder into the LDM latent space via the fixed VAE encoder. Similar to~\citep{JMLR:v23:21-0635, 10.1007/978-3-031-73030-6_19}, we apply a small amount of Gaussian noise to $x_0$ to smooth the base density. 

We note that this approach deviates considerably from the super-resolution (image restoration) application discussed in~\citep{10.1007/978-3-031-73030-6_19}; while those methods typically use a VAE-encoded low-quality image as conditioning, our method leverages rich semantic features extracted by our resolution-adaptive 1D tokenizer/ LDM feature extractor. This effectively separates the global semantic structure from high-frequency detail/ refinement (see~\cref{fig:impl_residual}). A key benefit of this separation is a simplified learning objective: by initializing the flow within the high-level semantic manifold of the target distribution, we narrow the distributional gap the model must bridge, allowing it to focus on local detail synthesis rather than recovering global coherence from scratch. The resulting objective is:
\begin{align}\label{eq:sana15}
&\mathcal{L}_{\text{CFM+}}(\Theta) = \mathbb{E}_{t, q(x_1), p(x_0)} \notag \\
&\quad \left\| v_t(\phi_t(x_0), z) - \left( x_1 - (1 - \sigma_{\text{min}}) x_0 \right) \right\|^2,
\end{align}

with $\Theta = (\theta_1, \theta_2)$ denoting the parameters of the flow model and LDM feature extractor, respectively. We initialize our feature extractor with our pre-trained resolution-adaptive 1D tokenizer decoder and keep all other components frozen. 

We note that \cref{eq:sana15} is formulated in the latent space of the auto-encoder rather than the pixel space. 
Finally, we use a uniform loss-weighting scheme and do not explore the more sophisticated non-uniform variants adopted in~\cite[Sec.~3.1]{esser2024scaling}. Different from PerCoV1, we do not apply classifier-free guidance~\citep{ho2021classifierfree, careil2024towards} at test time. However, we still drop the text conditioning $z_g$ in $50\%$ of the training iterations to evaluate the performance of our multimodal enhancement strategy discussed in~\cref{subsec:multimodalenhancement}.

\subsection{Extension to High-Resolution Images}
A key distinction of PerCoV2 is its reliance on a purely transformer-based diffusion architecture, diverging from the convolutional backbones used in almost all competing methods that rely on Stable Diffusion v1/v2 or GLIDE~\citep{pan_2022, lei2023text+sketch, careil2024towards, li2024misc, li2024towards, xia2025diffpc, Ke2025resulic}. While fully convolutional architectures are translation equivariant—allowing them to be trained on patches and applied to full-resolution images at test time—diffusion models based on self-attention lack this property.

A straightforward approach to maintaining train-test generalization is sequential or parallelized patch sampling~\citep{Hoogeboom2023a}. However, this strategy typically results in suboptimal hardware utilization and may introduce boundary artifacts where independently generated patches fail to achieve spatial consistency.

In PerCoV2, we instead \textit{natively} train on variable aspect ratio images at resolutions up to 2K. Using the SA-1B dataset~\citep{Kirillov_2023_ICCV}---averaging $1500 \times 2250$ pixels---we employ a variant of aspect ratio bucketing~\citep{novelai2022bucketing}. Specifically, we resize images with aspect-preserving interpolation (if required) and crop them randomly into 49 discrete buckets, defined in 256-pixel increments ranging from $512 \times 512, 512 \times 768, \dots, 2048 \times 2048$ pixels. To ensure structural coherence, we adopt the 2D positional embedding strategy introduced in SD3~\cite[Sec.~5.3.2]{esser2024scaling}, with subsequent center-cropping to the target bucket resolution. This native high-resolution training is feasible due to our initial choice of the SANA~\citep{xie2025sana} framework; by utilizing a linear transformer-based architecture, we effectively bypass the quadratic computational complexity of standard self-attention, making 2K training practical even on accessible hardware (2$\times$H100 GPUs).

\subsection{Towards Efficient One-Step Diffusion Decoding}

Finally, we introduce PerCoV2-distilled, a practical single-step variant derived from PerCoV2 to address the inherently slow decoding times associated with multi-step flow matching. The core idea is to directly predict $x_1$ from the noised base reconstruction $x_0$ at $t=0$, using the same PerCoV2 decoding stack. During optimization, we fix the pre-trained transformer weights and only finetune the model through LoRA~\citep{hu2022lora} adaptation layers along with the feature extractor, preserving the learned generative prior.

We adopt a hybrid pixel- and latent-domain optimization, following recent compression-oriented distillation approaches~\citep{xue2025one, jia2025cod, jia2026codlite}:
\begin{equation}
    \mathcal{L}
    =
    \mathcal{L}_{\mathrm{pixel}}
    +
    \lambda_{\mathrm{DMD}}\mathcal{L}_{\mathrm{DMD}},
\end{equation}
where $\mathcal{L}_{\mathrm{pixel}}$ is the standard VQGAN loss~\citep{Esser_2021_CVPR} used in TiTok. The DMD~\citep{Yin_2024_CVPR,3737916.3739421} objective corresponds to minimizing the KL divergence between the time-dependent distributions of the one-step decoder and the pre-trained multi-step model.

Given the one-step reconstruction $\hat{x}_1$, we construct the corresponding interpolation path by replacing the ground-truth $x_1$ in Eq.~\ref{eq:psi_t}:
\begin{equation}
    \phi_t'(x_0)
    =
    \big(1-(1-\sigma_{\mathrm{min}})t\big)x_0
    +t\hat{x}_1.
\end{equation}
The resulting gradient is given by the velocity difference propagated to the generator parameters:
\begin{align}
&\nabla_{\theta}\mathcal{L}_{\mathrm{DMD}}
=
\mathbb{E}_{t,\,q'(x_1),\,p(x_0)}
\notag\\
&\quad
\left[
\left(
v_t^{\mathrm{fake}}\!\left(\phi_t'(x_0),z\right)
-
v_t^{\mathrm{real}}\!\left(\phi_t'(x_0),z\right)
\right)
\frac{\partial \hat{x}_1}{\partial\theta}
\right].
\end{align}
where $v_t^{\mathrm{real}}$ is the pre-trained teacher velocity model, while $v_t^{\mathrm{fake}}$ is an additional velocity model trained in parallel to track the one-step decoder distribution. This velocity difference provides the KL-based distribution-matching signal, transferring the generative behavior of the multi-step model to the one-step decoder.

\section{Experimental Results}

\begin{figure*}[bt]
  \centering
  \includegraphics[width=1.0\linewidth]{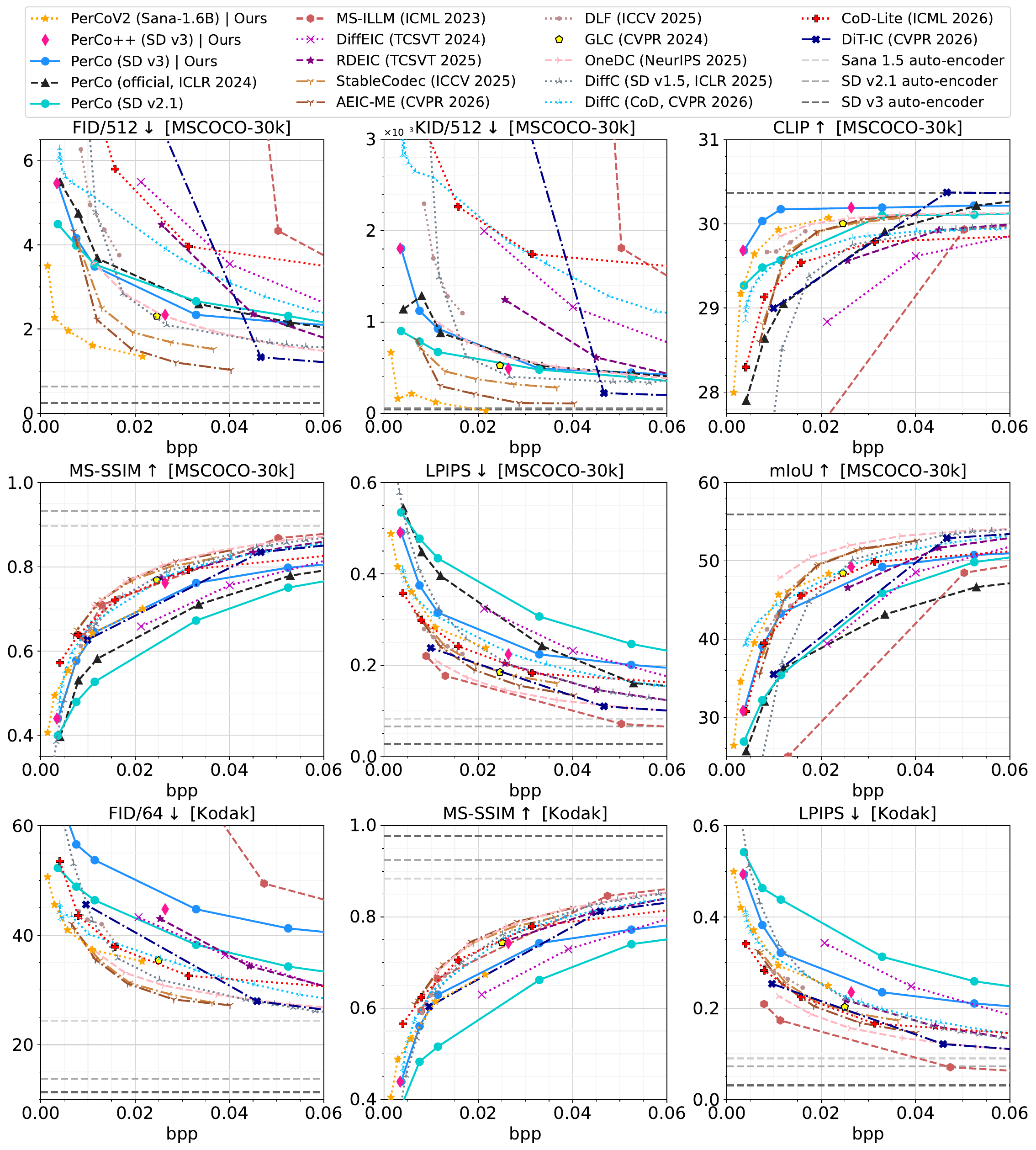}
  \caption{Quantitative comparison of PerCoV2 on MSCOCO-30k and Kodak at $512\times512$}
  \label{fig:quant_comparison}
\end{figure*}

\begin{figure*}[bt]
  \centering
  \includegraphics[width=1.0\linewidth]{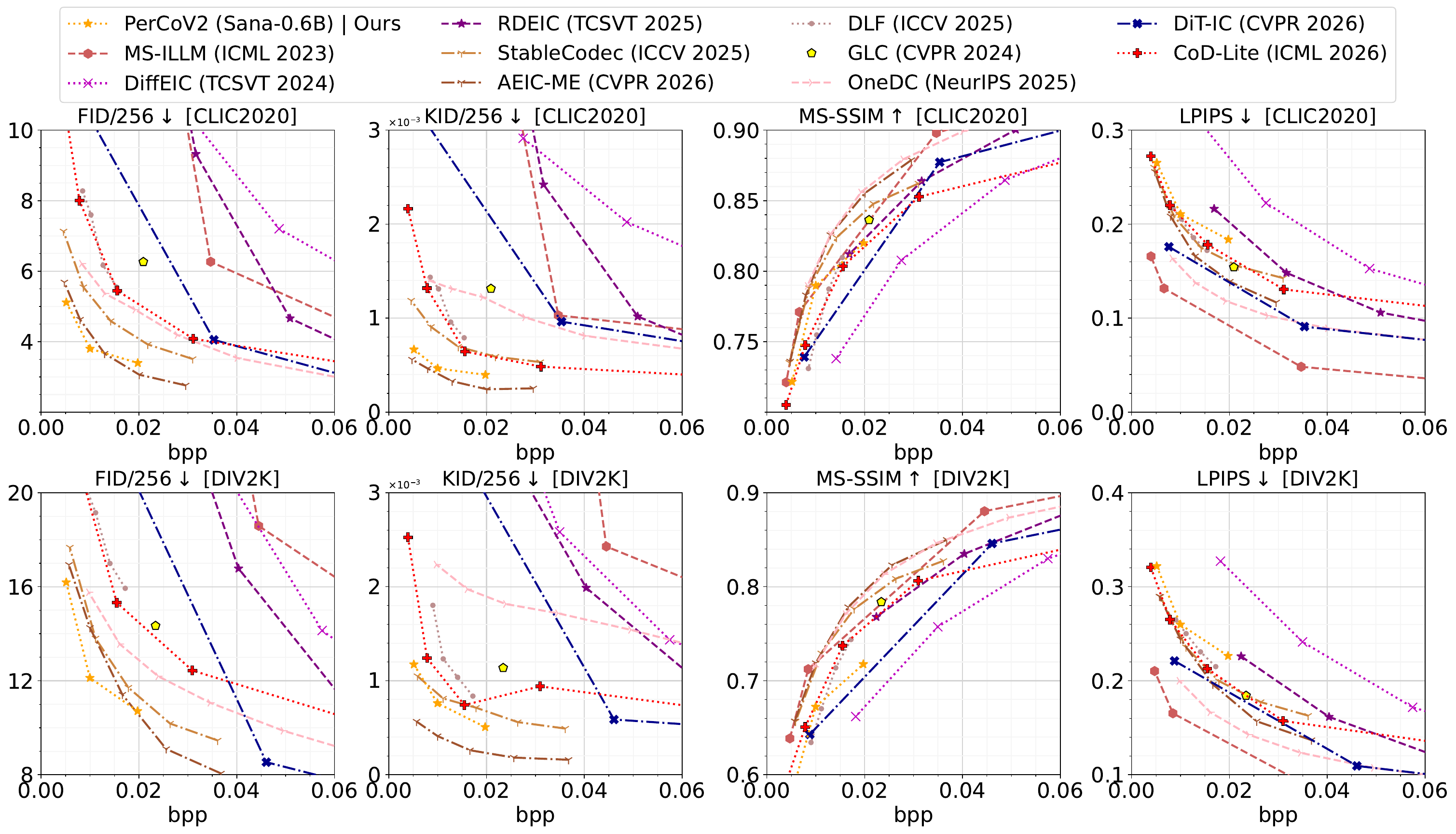}
  \caption{Zero-shot comparison of PerCoV2 on high-resolution benchmarks}
  \label{fig:quant_comparison_highres}
\end{figure*}

\begin{figure*}[bt]
    \setlength{\tabcolsep}{1.0pt}  
    \renewcommand{\arraystretch}{1.0}  
    \centering
    \scriptsize
    \begin{tabular}{cccccccc}
        \toprule
        \multicolumn{2}{c}{Original} & \multicolumn{2}{c}{MS-ILLM} & \multicolumn{2}{c}{DiT-IC} & \multicolumn{2}{c}{DiffC (CoD)} \\
        \multicolumn{2}{c}{} & \multicolumn{2}{c}{(ICML'23)} & \multicolumn{2}{c}{(CVPR'26)} & \multicolumn{2}{c}{(CVPR'26)} \\
        \midrule
        \begin{subfigure}{0.119\textwidth}
            \centering
            \includegraphics[width=\linewidth,height=\linewidth]{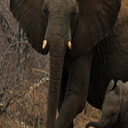}
        \end{subfigure}
        &
        \begin{subfigure}{0.119\textwidth}
            \centering
            \includegraphics[width=\linewidth,height=\linewidth]{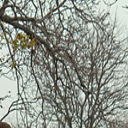}
        \end{subfigure}
        &
        \begin{subfigure}{0.119\textwidth}
            \centering
            \includegraphics[width=\linewidth,height=\linewidth]{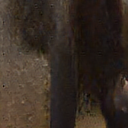}
        \end{subfigure}
        &
        \begin{subfigure}{0.119\textwidth}
            \centering
            \includegraphics[width=\linewidth,height=\linewidth]{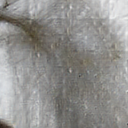}
        \end{subfigure}
        &
        \begin{subfigure}{0.119\textwidth}
            \centering
            \includegraphics[width=\linewidth,height=\linewidth]{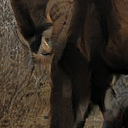}
        \end{subfigure}
        &
        \begin{subfigure}{0.119\textwidth}
            \centering
            \includegraphics[width=\linewidth,height=\linewidth]{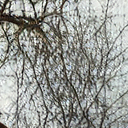}
        \end{subfigure}
        &
        \begin{subfigure}{0.119\textwidth}
            \centering
            \includegraphics[width=\linewidth,height=\linewidth]{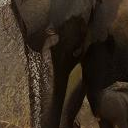}
        \end{subfigure}
        &
        \begin{subfigure}{0.119\textwidth}
            \centering
            \includegraphics[width=\linewidth,height=\linewidth]{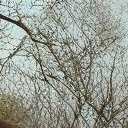}
        \end{subfigure}
        \\
        \multicolumn{2}{c}{\begin{subfigure}{0.2425\textwidth}
            \centering
            \includegraphics[width=\linewidth,height=\linewidth]{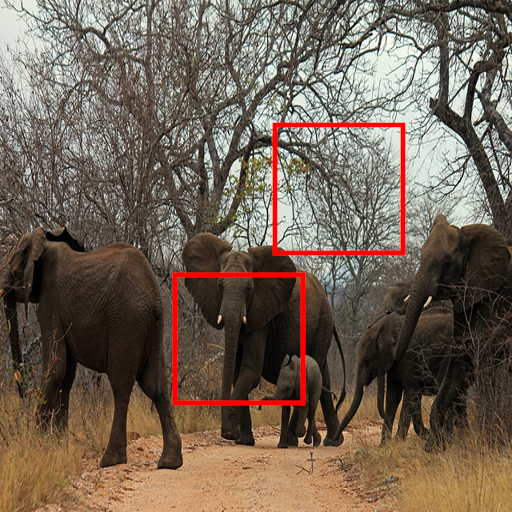}
        \end{subfigure}}
        &
        \multicolumn{2}{c}{\begin{subfigure}{0.2425\textwidth}
            \centering
            \includegraphics[width=\linewidth,height=\linewidth]{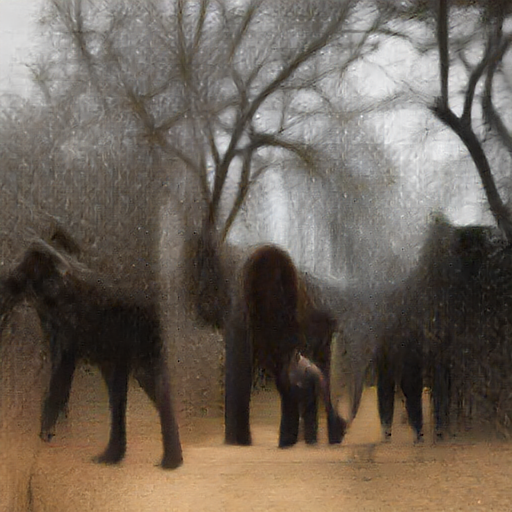}
        \end{subfigure}}
        &
        \multicolumn{2}{c}{\begin{subfigure}{0.2425\textwidth}
            \centering
            \includegraphics[width=\linewidth,height=\linewidth]{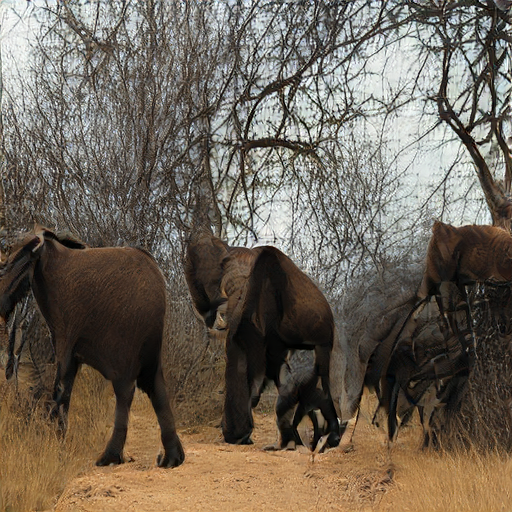}
        \end{subfigure}}
        &
        \multicolumn{2}{c}{\begin{subfigure}{0.2425\textwidth}
            \centering
            \includegraphics[width=\linewidth,height=\linewidth]{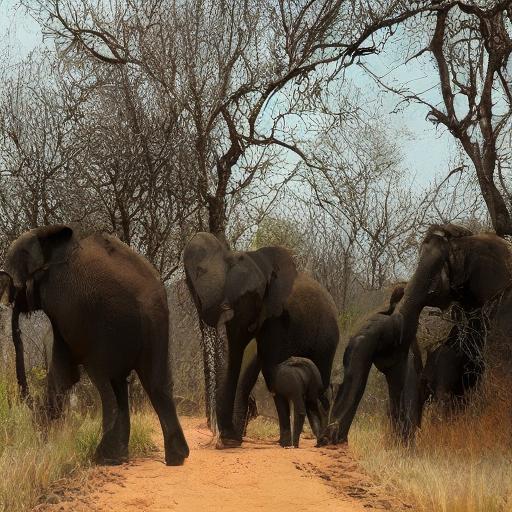}
        \end{subfigure}}
        \\ 
        \multicolumn{2}{c}{\texttt{000000000312}} & \multicolumn{2}{c}{$0.01245$~bpp ($2.12\times$)} & \multicolumn{2}{c}{$0.00928$~bpp ($1.58\times$)} & \multicolumn{2}{c}{$0.0069$~bpp ($1.18\times$)} \\
        \midrule
        \multicolumn{2}{c}{PerCoV1} & \multicolumn{2}{c}{DLF} & \multicolumn{2}{c}{AEIC-ME} & \multicolumn{2}{c}{PerCoV2} \\
        \multicolumn{2}{c}{(ICLR'24)} & \multicolumn{2}{c}{(ICCV'25)} & \multicolumn{2}{c}{(CVPR'26)} & \multicolumn{2}{c}{(Ours)} \\
        \midrule
        \begin{subfigure}{0.119\textwidth}
            \centering
            \includegraphics[width=\linewidth,height=\linewidth]{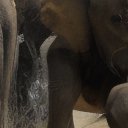}
        \end{subfigure}
        &
        \begin{subfigure}{0.119\textwidth}
            \centering
            \includegraphics[width=\linewidth,height=\linewidth]{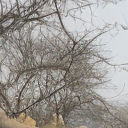}
        \end{subfigure}
        &
        \begin{subfigure}{0.119\textwidth}
            \centering
            \includegraphics[width=\linewidth,height=\linewidth]{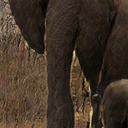}
        \end{subfigure}
        &
        \begin{subfigure}{0.119\textwidth}
            \centering
            \includegraphics[width=\linewidth,height=\linewidth]{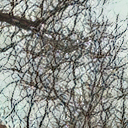}
        \end{subfigure}
        &
        \begin{subfigure}{0.119\textwidth}
            \centering
            \includegraphics[width=\linewidth,height=\linewidth]{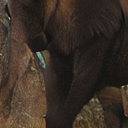}
        \end{subfigure}
        &
        \begin{subfigure}{0.119\textwidth}
            \centering
            \includegraphics[width=\linewidth,height=\linewidth]{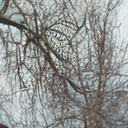}
        \end{subfigure}
        &
        \begin{subfigure}{0.119\textwidth}
            \centering
            \includegraphics[width=\linewidth,height=\linewidth]{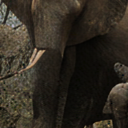}
        \end{subfigure}
        &
        \begin{subfigure}{0.119\textwidth}
            \centering
            \includegraphics[width=\linewidth,height=\linewidth]{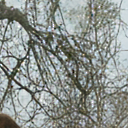}
        \end{subfigure}\\
        \multicolumn{2}{c}{\begin{subfigure}{0.2425\textwidth}
            \centering
            \includegraphics[width=\linewidth,height=\linewidth]{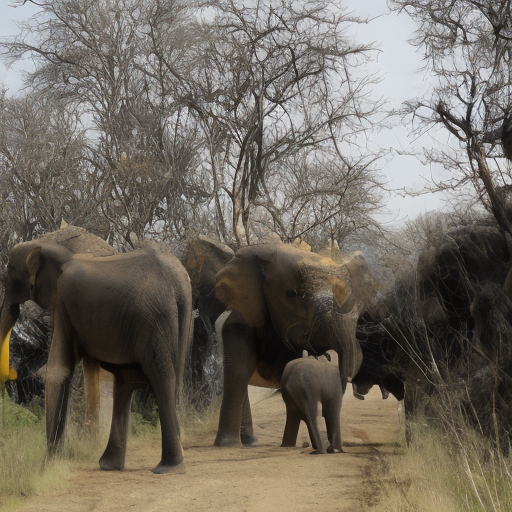}
        \end{subfigure}}
        &
        \multicolumn{2}{c}{\begin{subfigure}{0.2425\textwidth}
            \centering
            \includegraphics[width=\linewidth,height=\linewidth]{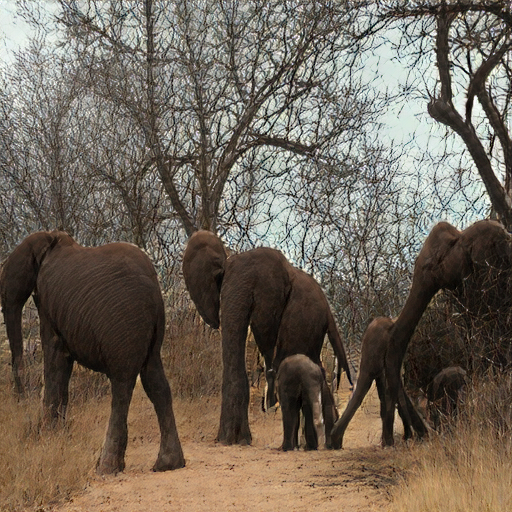}
        \end{subfigure}}
        &
        \multicolumn{2}{c}{\begin{subfigure}{0.2425\textwidth}
            \centering
            \includegraphics[width=\linewidth,height=\linewidth]{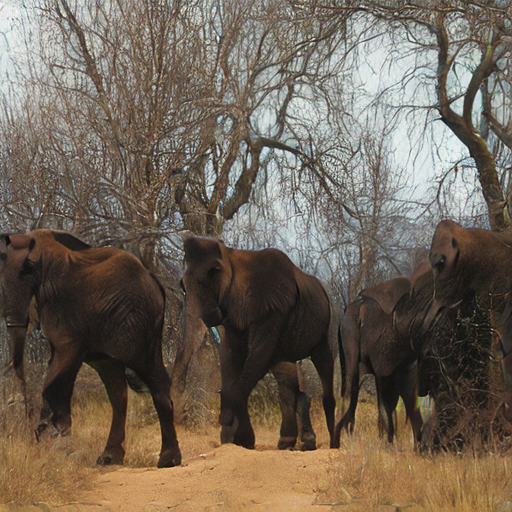}
        \end{subfigure}}
        &
        \multicolumn{2}{c}{\begin{subfigure}{0.2425\textwidth}
            \centering
            \includegraphics[width=\linewidth,height=\linewidth]{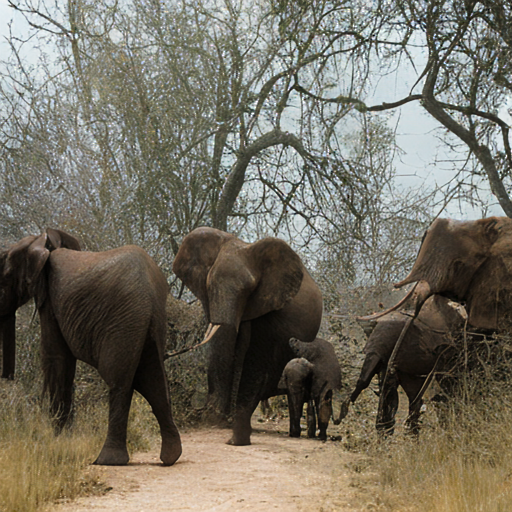}
        \end{subfigure}}\\
        \multicolumn{2}{c}{$0.0098$~bpp ($1.67\times$)} & \multicolumn{2}{c}{$0.00772$~bpp ($1.32\times$)} & \multicolumn{2}{c}{$0.00583$~bpp ($0.99\times$)} & \multicolumn{2}{c}{$0.00586$~bpp} \\
        \bottomrule
        
    \end{tabular}

    \caption{Visual comparison of PerCoV2 and competing methods on MSCOCO-30k; bit-rate increases relative to PerCoV2 are shown as $(\times)$. \textbf{Best viewed zoomed-in.}}
    \label{fig:vis_impressions_2}
\end{figure*}

\begin{figure*}[tb]
    \setlength{\tabcolsep}{1.0pt}  
    \renewcommand{\arraystretch}{1.0}  
    \centering
    \scriptsize
    \begin{tabular}{cccccccc}
        \toprule
        \multicolumn{2}{c}{Original} & \multicolumn{2}{c}{MS-ILLM} & \multicolumn{2}{c}{DiT-IC} & \multicolumn{2}{c}{DiffC (CoD)}  \\
        \multicolumn{2}{c}{} & \multicolumn{2}{c}{(ICML'23)} & \multicolumn{2}{c}{(CVPR'26)} & \multicolumn{2}{c}{(CVPR'26)}  \\
        \midrule
        \begin{subfigure}{0.118\textwidth}
            \centering
            \includegraphics[width=\linewidth,height=\linewidth]{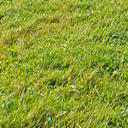}
        \end{subfigure}
        &
        \begin{subfigure}{0.118\textwidth}
            \centering
            \includegraphics[width=\linewidth,height=\linewidth]{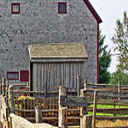}
        \end{subfigure}
        &
        \begin{subfigure}{0.118\textwidth}
            \centering
            \includegraphics[width=\linewidth,height=\linewidth]{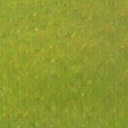}
        \end{subfigure}
        &
        \begin{subfigure}{0.118\textwidth}
            \centering
            \includegraphics[width=\linewidth,height=\linewidth]{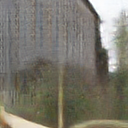}
        \end{subfigure}
        &
        \begin{subfigure}{0.118\textwidth}
            \centering
            \includegraphics[width=\linewidth,height=\linewidth]{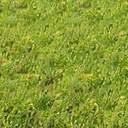}
        \end{subfigure}
        &
        \begin{subfigure}{0.118\textwidth}
            \centering
            \includegraphics[width=\linewidth,height=\linewidth]{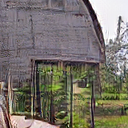}
        \end{subfigure}  
        &
        \begin{subfigure}{0.118\textwidth}
            \centering
            \includegraphics[width=\linewidth,height=\linewidth]{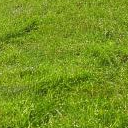}
        \end{subfigure}
        &
        \begin{subfigure}{0.118\textwidth}
            \centering
            \includegraphics[width=\linewidth,height=\linewidth]{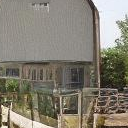}
        \end{subfigure}      
        \\
        \multicolumn{2}{c}{\begin{subfigure}{0.2425\textwidth}
            \centering
            \includegraphics[width=\linewidth,height=\linewidth]{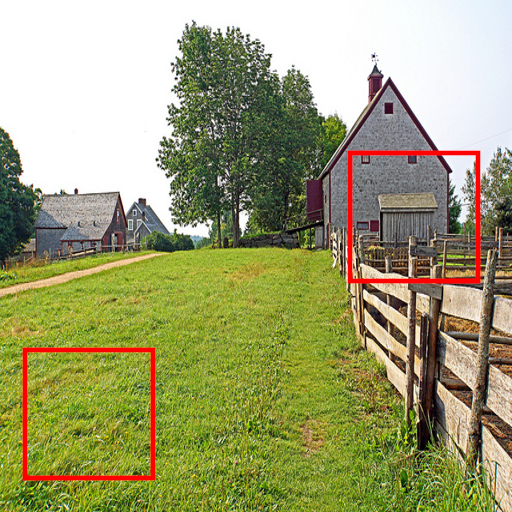}
        \end{subfigure}}
        &
        \multicolumn{2}{c}{\begin{subfigure}{0.2425\textwidth}
            \centering
            \includegraphics[width=\linewidth,height=\linewidth]{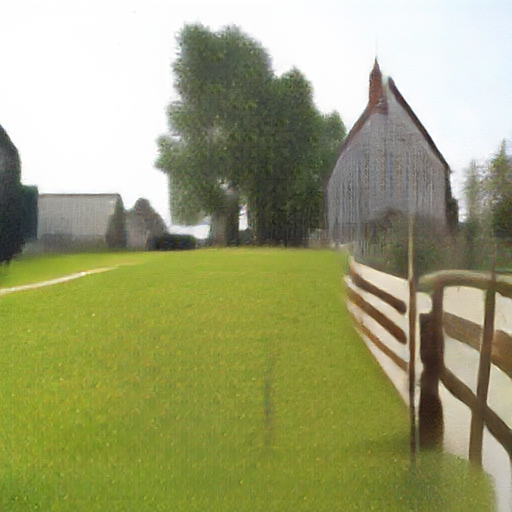}
        \end{subfigure}}
        &
        \multicolumn{2}{c}{\begin{subfigure}{0.2425\textwidth}
            \centering
            \includegraphics[width=\linewidth,height=\linewidth]{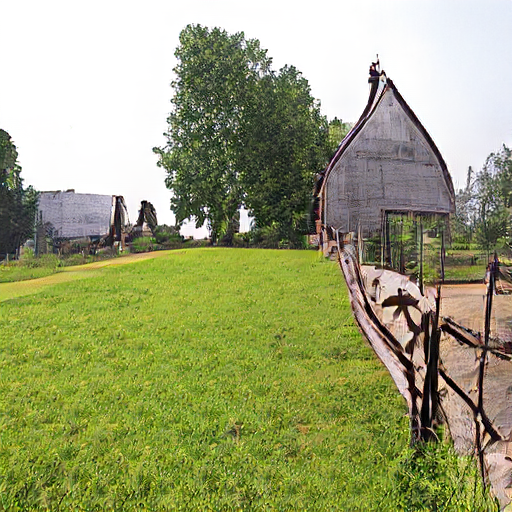}
        \end{subfigure}} 
        &
        \multicolumn{2}{c}{\begin{subfigure}{0.2425\textwidth}
            \centering
            \includegraphics[width=\linewidth,height=\linewidth]{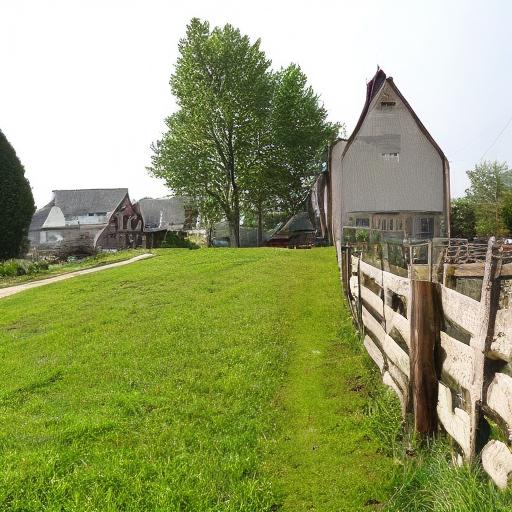}
        \end{subfigure}}       
        \\
        \multicolumn{2}{c}{\texttt{000000004517}} & \multicolumn{2}{c}{$0.01245$~bpp ($2.12\times$)} & \multicolumn{2}{c}{$0.00842$~bpp ($1.44\times$)} & \multicolumn{2}{c}{$0.0065$~bpp ($1.11\times$)} \\
        \midrule
        \multicolumn{2}{c}{PerCoV1} & \multicolumn{2}{c}{DLF} & \multicolumn{2}{c}{AEIC-ME} & \multicolumn{2}{c}{PerCoV2} \\
        \multicolumn{2}{c}{(ICLR'24)} & \multicolumn{2}{c}{(ICCV'25)} & \multicolumn{2}{c}{(CVPR'26)} & \multicolumn{2}{c}{(Ours)} \\
        \midrule
        \begin{subfigure}{0.118\textwidth}
            \centering
            \includegraphics[width=\linewidth,height=\linewidth]{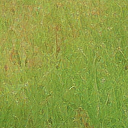}
        \end{subfigure}
        &
        \begin{subfigure}{0.118\textwidth}
            \centering
            \includegraphics[width=\linewidth,height=\linewidth]{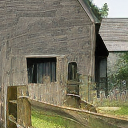}
        \end{subfigure}
        &
        \begin{subfigure}{0.118\textwidth}
            \centering
            \includegraphics[width=\linewidth,height=\linewidth]{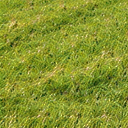}
        \end{subfigure}
        &
        \begin{subfigure}{0.118\textwidth}
            \centering
            \includegraphics[width=\linewidth,height=\linewidth]{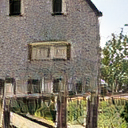}
        \end{subfigure}
        &
        \begin{subfigure}{0.118\textwidth}
            \centering
            \includegraphics[width=\linewidth,height=\linewidth]{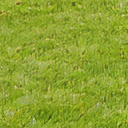}
        \end{subfigure}
        &
        \begin{subfigure}{0.118\textwidth}
            \centering
            \includegraphics[width=\linewidth,height=\linewidth]{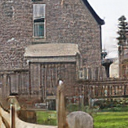}
        \end{subfigure}
        &
        \begin{subfigure}{0.118\textwidth}
            \centering
            \includegraphics[width=\linewidth,height=\linewidth]{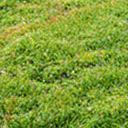}
        \end{subfigure}
        &
        \begin{subfigure}{0.118\textwidth}
            \centering
            \includegraphics[width=\linewidth,height=\linewidth]{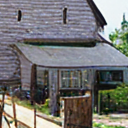}
        \end{subfigure}\\
        \multicolumn{2}{c}{\begin{subfigure}{0.2425\textwidth}
            \centering
            \includegraphics[width=\linewidth,height=\linewidth]{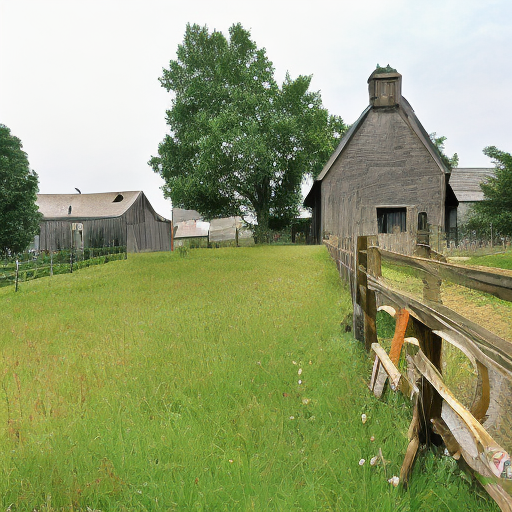}
        \end{subfigure}}
        &
        \multicolumn{2}{c}{\begin{subfigure}{0.2425\textwidth}
            \centering
            \includegraphics[width=\linewidth,height=\linewidth]{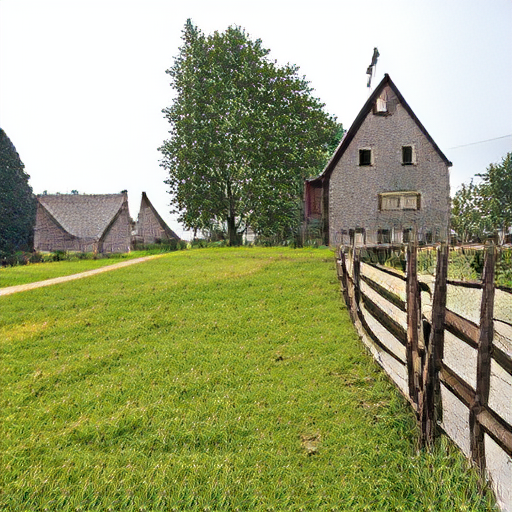}
        \end{subfigure}}
        &
        \multicolumn{2}{c}{\begin{subfigure}{0.2425\textwidth}
            \centering
            \includegraphics[width=\linewidth,height=\linewidth]{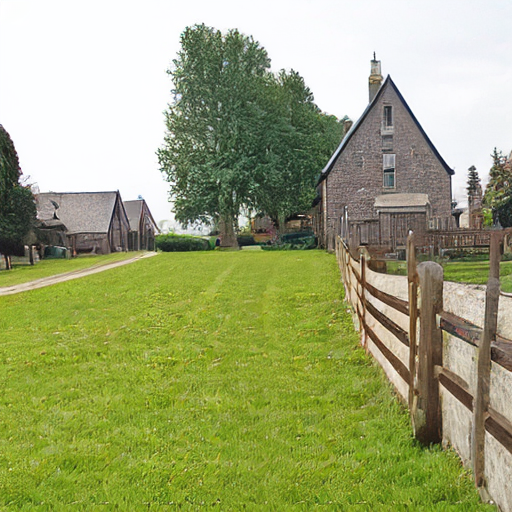}
        \end{subfigure}}
        &
        \multicolumn{2}{c}{\begin{subfigure}{0.2425\textwidth}
            \centering
            \includegraphics[width=\linewidth,height=\linewidth]{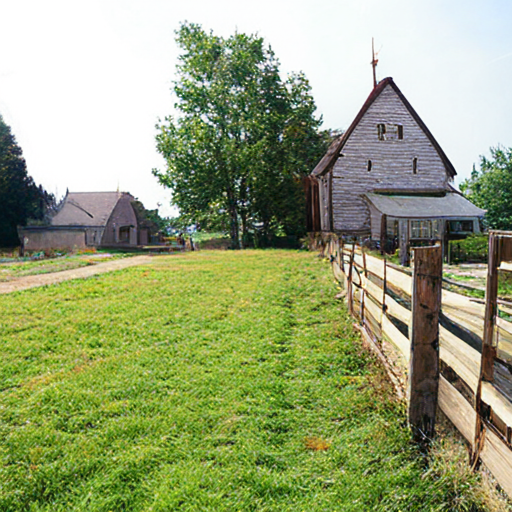}
        \end{subfigure}}\\
        \multicolumn{2}{c}{$0.0098$~bpp ($1.67\times$)} & \multicolumn{2}{c}{$0.0079$~bpp ($1.35\times$)} & \multicolumn{2}{c}{$0.00623$~bpp ($1.06\times$)} & \multicolumn{2}{c}{$0.00586$~bpp} \\
        \bottomrule
        
    \end{tabular}

    \caption{Extended comparison of PerCoV2 and competing methods on MSCOCO-30k; bit-rate increases relative to PerCoV2 are shown as $(\times)$. \textbf{Best viewed zoomed-in.}}
    \label{fig:vis_impressions_2_ext1}
\end{figure*}

\begin{figure*}[tb]
    \setlength{\tabcolsep}{1.0pt}  
    \renewcommand{\arraystretch}{1.0}  
    \centering
    \scriptsize
    \begin{tabular}{cccccccc}
        \toprule
        \multicolumn{2}{c}{Original} & \multicolumn{2}{c}{MS-ILLM} & \multicolumn{2}{c}{DiT-IC} & \multicolumn{2}{c}{DiffC (CoD)} \\
        \multicolumn{2}{c}{} & \multicolumn{2}{c}{(ICML'23)} & \multicolumn{2}{c}{(CVPR'26)} & \multicolumn{2}{c}{(CVPR'26)}  \\
        \midrule
        \begin{subfigure}{0.118\textwidth}
            \centering
            \includegraphics[width=\linewidth,height=\linewidth]{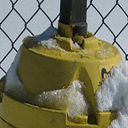}
        \end{subfigure}
        &
        \begin{subfigure}{0.118\textwidth}
            \centering
            \includegraphics[width=\linewidth,height=\linewidth]{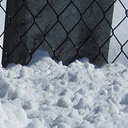}
        \end{subfigure}
        &
        \begin{subfigure}{0.118\textwidth}
            \centering
            \includegraphics[width=\linewidth,height=\linewidth]{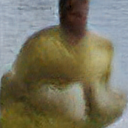}
        \end{subfigure}
        &
        \begin{subfigure}{0.118\textwidth}
            \centering
            \includegraphics[width=\linewidth,height=\linewidth]{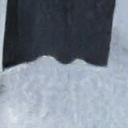}
        \end{subfigure}
        &
        \begin{subfigure}{0.118\textwidth}
            \centering
            \includegraphics[width=\linewidth,height=\linewidth]{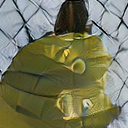}
        \end{subfigure}
        &
        \begin{subfigure}{0.118\textwidth}
            \centering
            \includegraphics[width=\linewidth,height=\linewidth]{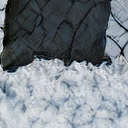}
        \end{subfigure} 
        &
        \begin{subfigure}{0.118\textwidth}
            \centering
            \includegraphics[width=\linewidth,height=\linewidth]{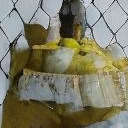}
        \end{subfigure}
        &
        \begin{subfigure}{0.118\textwidth}
            \centering
            \includegraphics[width=\linewidth,height=\linewidth]{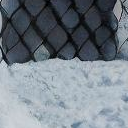}
        \end{subfigure}      
        \\
        \multicolumn{2}{c}{\begin{subfigure}{0.2425\textwidth}
            \centering
            \includegraphics[width=\linewidth,height=\linewidth]{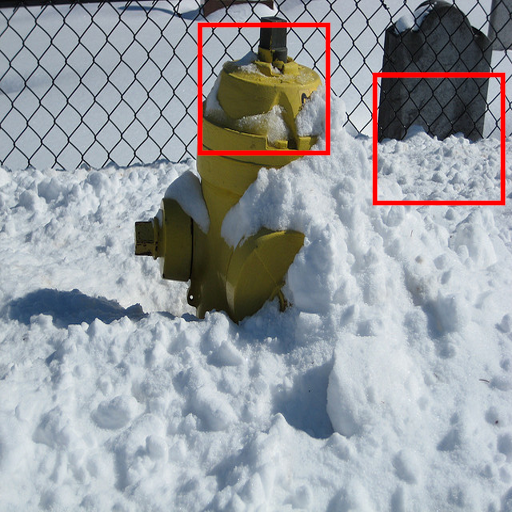}
        \end{subfigure}}
        &
        \multicolumn{2}{c}{\begin{subfigure}{0.2425\textwidth}
            \centering
            \includegraphics[width=\linewidth,height=\linewidth]{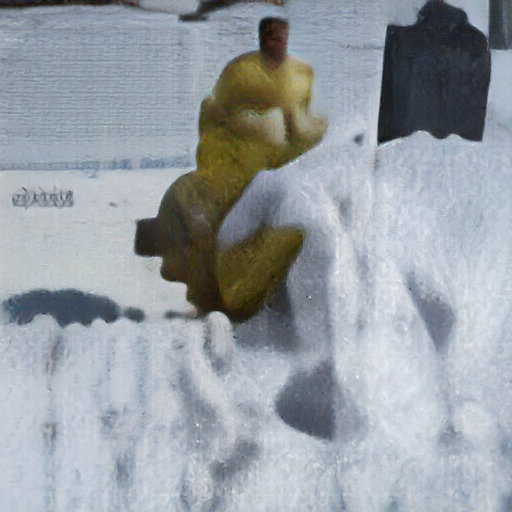}
        \end{subfigure}}
        &
        \multicolumn{2}{c}{\begin{subfigure}{0.2425\textwidth}
            \centering
            \includegraphics[width=\linewidth,height=\linewidth]{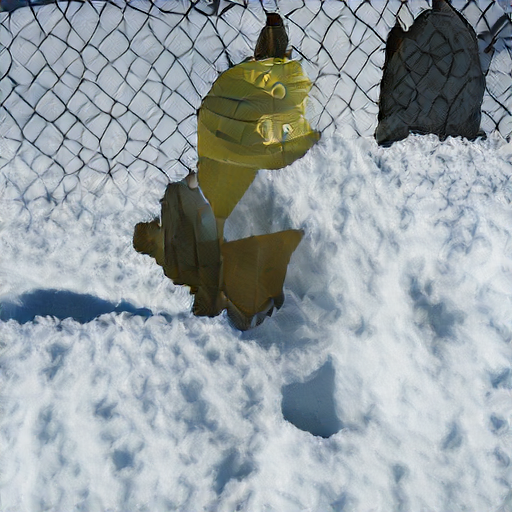}
        \end{subfigure}}
        &
        \multicolumn{2}{c}{\begin{subfigure}{0.2425\textwidth}
            \centering
            \includegraphics[width=\linewidth,height=\linewidth]{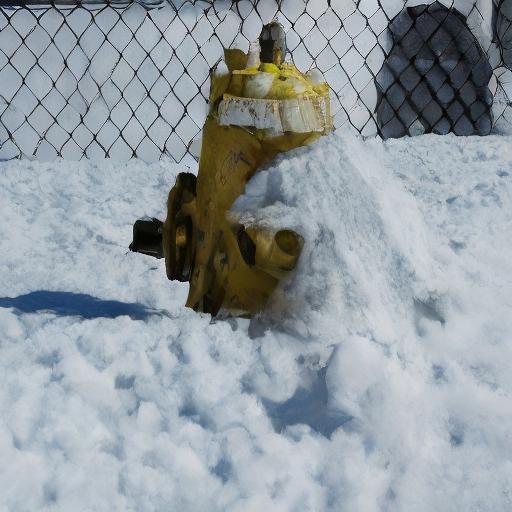}
        \end{subfigure}}     
        \\ 
        \multicolumn{2}{c}{\texttt{000000004394}} & \multicolumn{2}{c}{$0.01196$~bpp ($2.04\times$)} & \multicolumn{2}{c}{$0.00928$~bpp ($1.58\times$)} & \multicolumn{2}{c}{$0.0068$~bpp ($1.16\times$)} \\
        \midrule
        \multicolumn{2}{c}{PerCoV1} & \multicolumn{2}{c}{DLF} & \multicolumn{2}{c}{AEIC-ME} & \multicolumn{2}{c}{PerCoV2} \\
        \multicolumn{2}{c}{(ICLR'24)} & \multicolumn{2}{c}{(ICCV'25)} & \multicolumn{2}{c}{(CVPR'26)} & \multicolumn{2}{c}{(Ours)} \\
        \midrule
        \begin{subfigure}{0.118\textwidth}
            \centering
            \includegraphics[width=\linewidth,height=\linewidth]{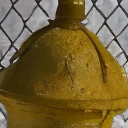}
        \end{subfigure}
        &
        \begin{subfigure}{0.118\textwidth}
            \centering
            \includegraphics[width=\linewidth,height=\linewidth]{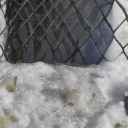}
        \end{subfigure}
        &
        \begin{subfigure}{0.118\textwidth}
            \centering
            \includegraphics[width=\linewidth,height=\linewidth]{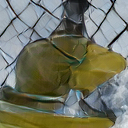}
        \end{subfigure}
        &
        \begin{subfigure}{0.118\textwidth}
            \centering
            \includegraphics[width=\linewidth,height=\linewidth]{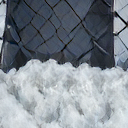}
        \end{subfigure}
        &
        \begin{subfigure}{0.118\textwidth}
            \centering
            \includegraphics[width=\linewidth,height=\linewidth]{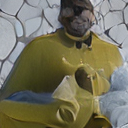}
        \end{subfigure}
        &
        \begin{subfigure}{0.118\textwidth}
            \centering
            \includegraphics[width=\linewidth,height=\linewidth]{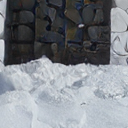}
        \end{subfigure}
        &
        \begin{subfigure}{0.118\textwidth}
            \centering
            \includegraphics[width=\linewidth,height=\linewidth]{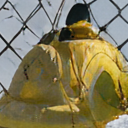}
        \end{subfigure}
        &
        \begin{subfigure}{0.118\textwidth}
            \centering
            \includegraphics[width=\linewidth,height=\linewidth]{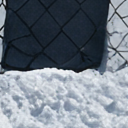}
        \end{subfigure}\\
        \multicolumn{2}{c}{\begin{subfigure}{0.2425\textwidth}
            \centering
            \includegraphics[width=\linewidth,height=\linewidth]{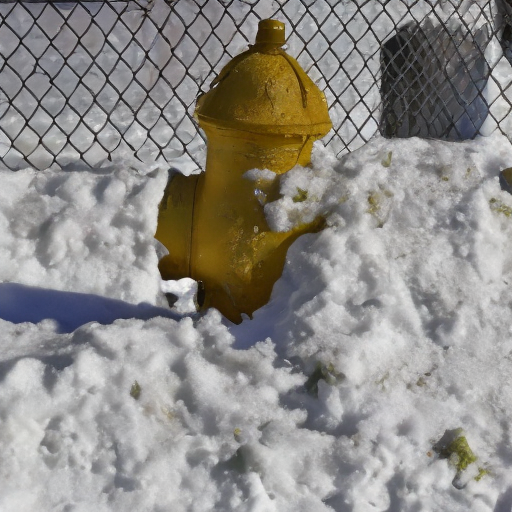}
        \end{subfigure}}
        &
        \multicolumn{2}{c}{\begin{subfigure}{0.2425\textwidth}
            \centering
            \includegraphics[width=\linewidth,height=\linewidth]{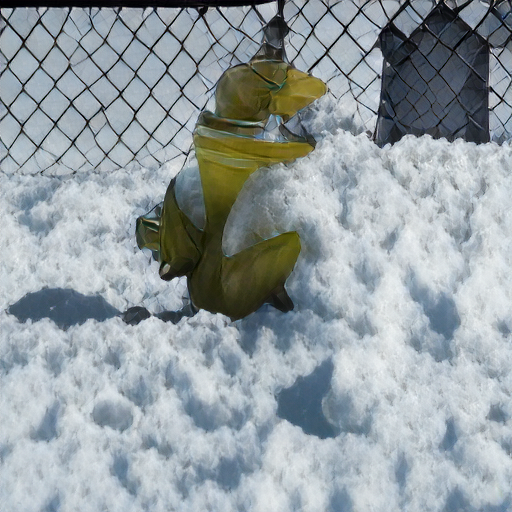}
        \end{subfigure}}
        &
        \multicolumn{2}{c}{\begin{subfigure}{0.2425\textwidth}
            \centering
            \includegraphics[width=\linewidth,height=\linewidth]{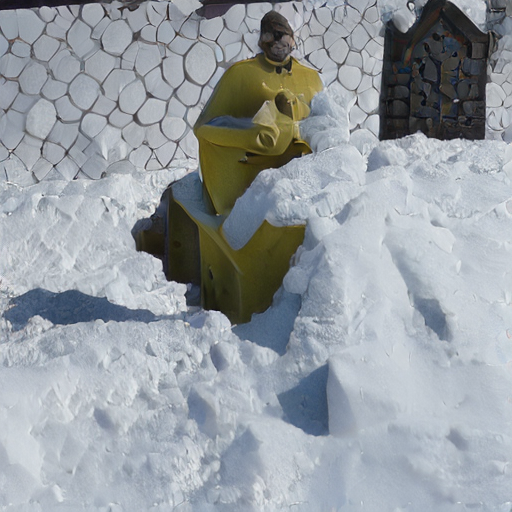}
        \end{subfigure}}
        &
        \multicolumn{2}{c}{\begin{subfigure}{0.2425\textwidth}
            \centering
            \includegraphics[width=\linewidth,height=\linewidth]{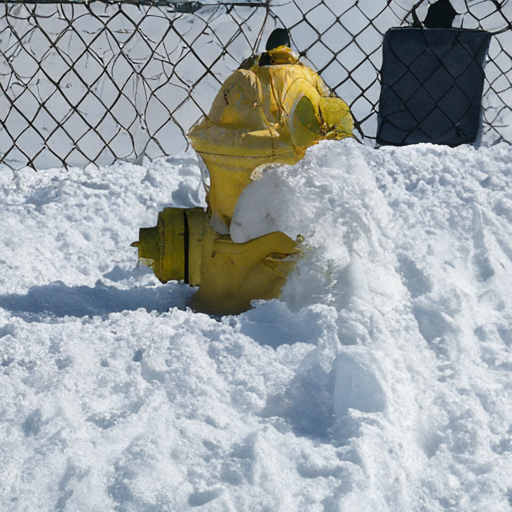}
        \end{subfigure}}\\
        \multicolumn{2}{c}{$0.0098$~bpp ($1.67\times$)} & \multicolumn{2}{c}{$0.00879$~bpp ($1.5\times$)} & \multicolumn{2}{c}{$0.00607$~bpp ($1.04\times$)} & \multicolumn{2}{c}{$0.00586$~bpp} \\
        \bottomrule
        
    \end{tabular}

    \caption{Extended comparison of PerCoV2 and competing methods on MSCOCO-30k; bit-rate increases relative to PerCoV2 are shown as $(\times)$. \textbf{Best viewed zoomed-in.}}
    \label{fig:vis_impressions_2_ext2}
\end{figure*}

\begin{figure*}[tb]
    \setlength{\tabcolsep}{1.0pt}  
    \renewcommand{\arraystretch}{1.0}  
    \centering
    \scriptsize
    \begin{tabular}{cccc}
        \toprule
        Original & JPEG & DCVC-RT-intra & CoD-Lite \\
        & (PIL) & (CVPR'25) & (ICML'26)\\
        \midrule
        \begin{subfigure}{0.245\textwidth}
            \centering
            \includegraphics[width=\linewidth]{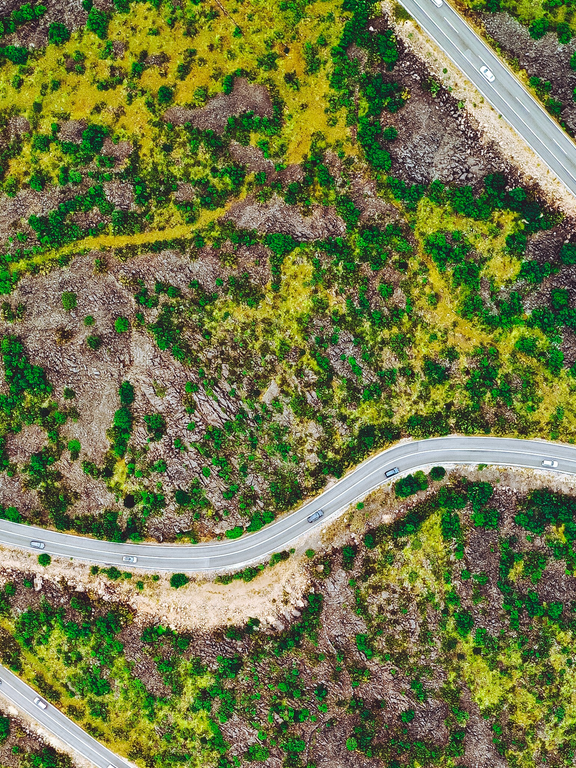}
        \end{subfigure}
        &
        \begin{subfigure}{0.245\textwidth}
            \centering
            \includegraphics[width=\linewidth]{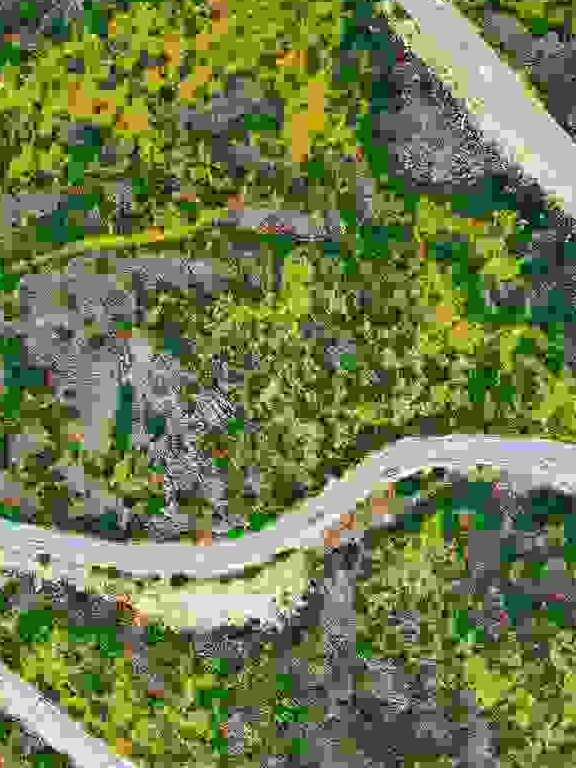}
        \end{subfigure}        
        &
        \begin{subfigure}{0.245\textwidth}
            \centering
            \includegraphics[width=\linewidth]{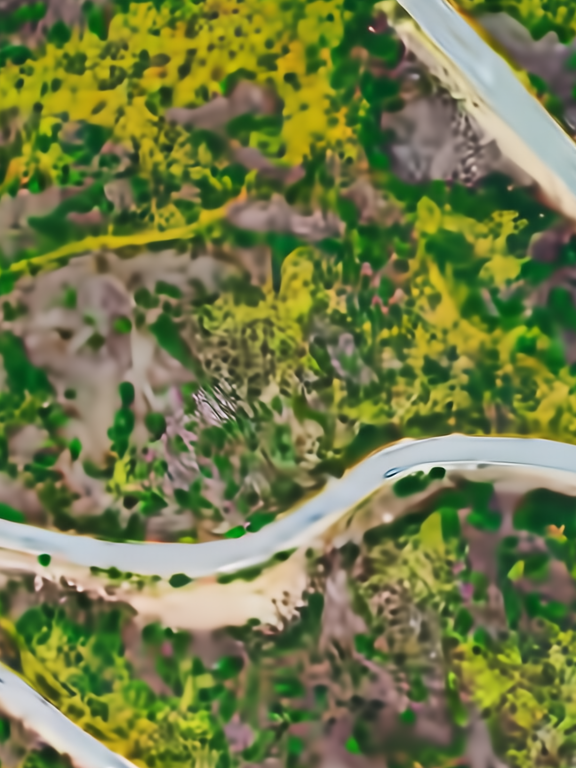}
        \end{subfigure}
        &
        \begin{subfigure}{0.245\textwidth}
            \centering
            \includegraphics[width=\linewidth]{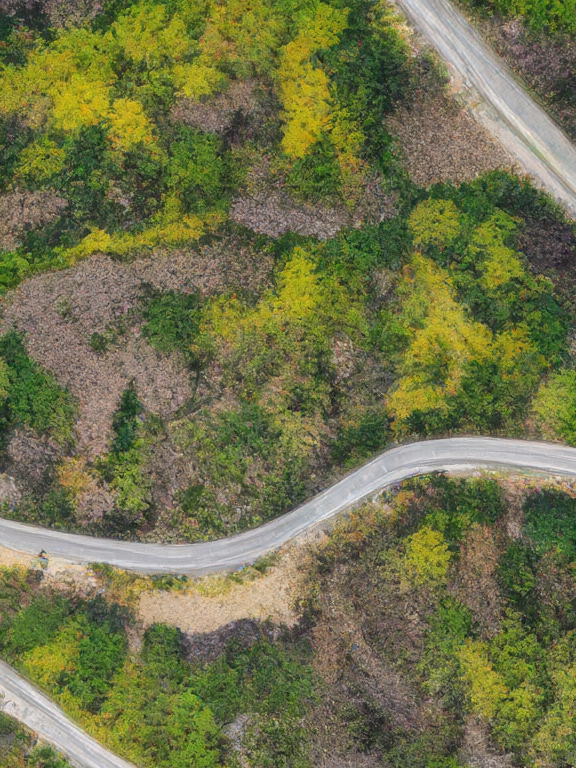}
        \end{subfigure}
        \\
        \texttt{a1810} & $0.11826$~bpp ($\times 22.66$) & $0.08160$~bpp ($\times 15.63$) & $0.00784$~bpp ($\times 1.50$) \\
        \midrule
        DLF & StableCodec & AEIC-ME & PerCoV2\\
        (ICCV'25) & (ICCV'25) & (CVPR'26) & (Ours)\\
        \midrule
        \begin{subfigure}{0.245\textwidth}
            \centering
            \includegraphics[width=\linewidth]{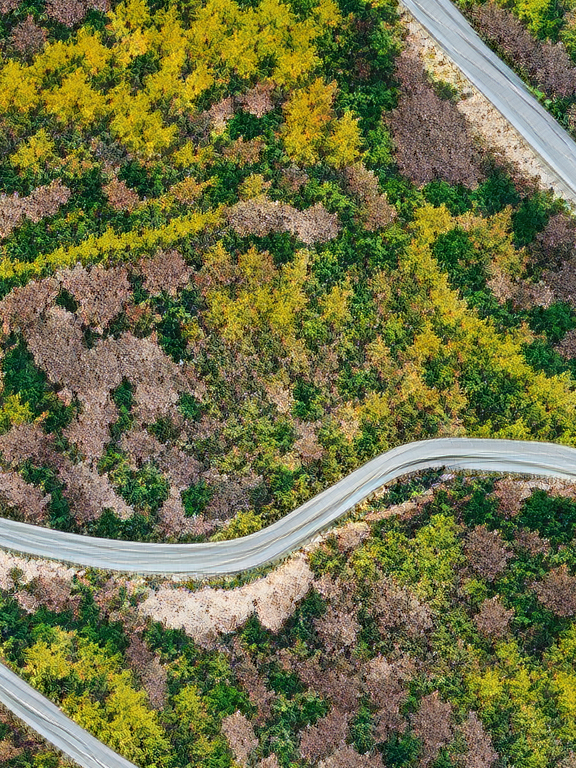}
        \end{subfigure}
        &
        \begin{subfigure}{0.245\textwidth}
            \centering
            \includegraphics[width=\linewidth]{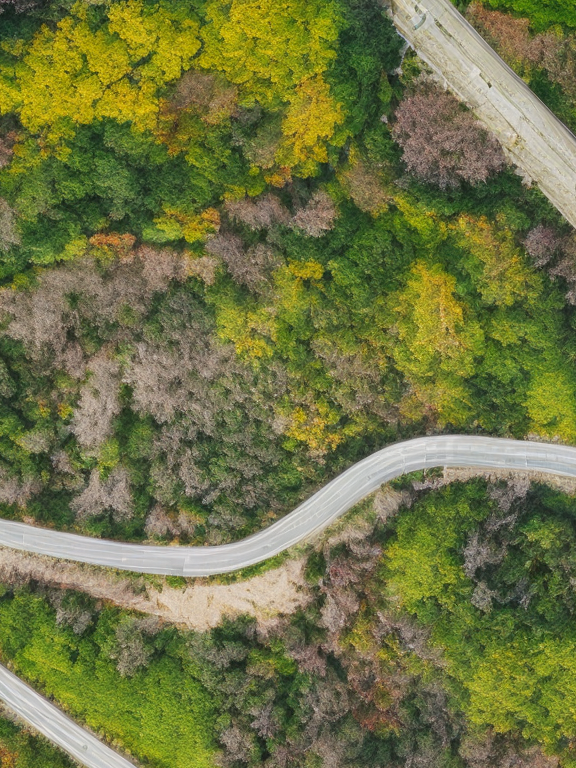}
        \end{subfigure}        
        &
        \begin{subfigure}{0.245\textwidth}
            \centering
            \includegraphics[width=\linewidth]{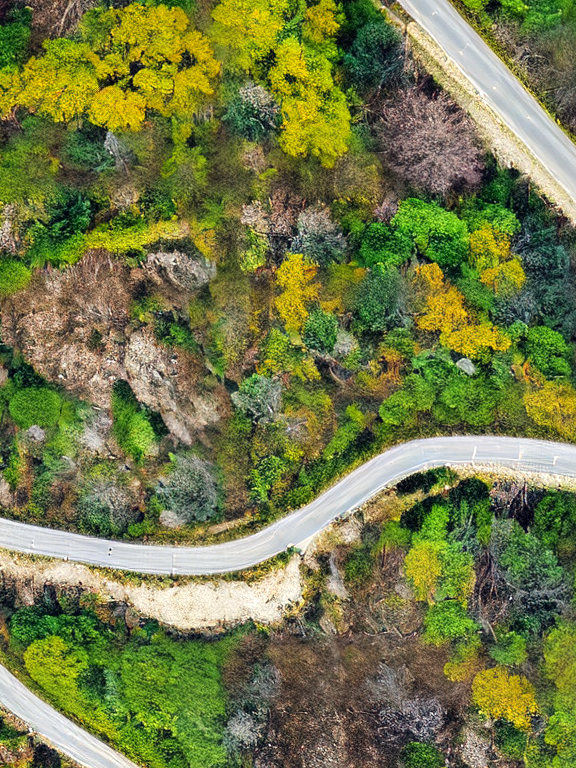}
        \end{subfigure}
        &
        \begin{subfigure}{0.245\textwidth}
            \centering
            \includegraphics[width=\linewidth]{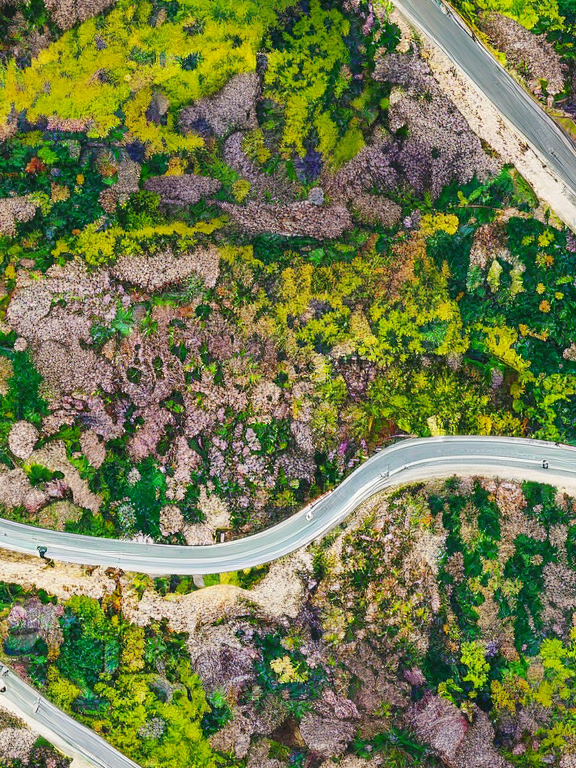}
        \end{subfigure}
        \\
        $0.00886$~bpp ($\times 1.70$) & $0.00545$~bpp ($\times 1.04$) & $0.00529$~bpp ($\times 1.01$) & $0.00522$~bpp \\
        \bottomrule

    \end{tabular}

    \caption{Visual comparison on CLIC 2020. We show $768 \times 576$ crops taken from high-resolution images of size $2048 \times 1152$; bit-rate increases relative to PerCoV2 are shown as $(\times)$. \textbf{Best viewed zoomed-in.}}
    \label{fig:high_res1}
\end{figure*}

\begin{figure*}[tb]
    \setlength{\tabcolsep}{1.0pt}  
    \renewcommand{\arraystretch}{1.0}  
    \centering
    \scriptsize
    \begin{tabular}{cccc}

        \toprule
        Original & JPEG & DCVC-RT-intra & CoD-Lite \\
        & (PIL) & (CVPR'25) & (ICML'26)\\
        \midrule
    
        \begin{subfigure}{0.245\textwidth}
            \centering
            \includegraphics[width=\linewidth]{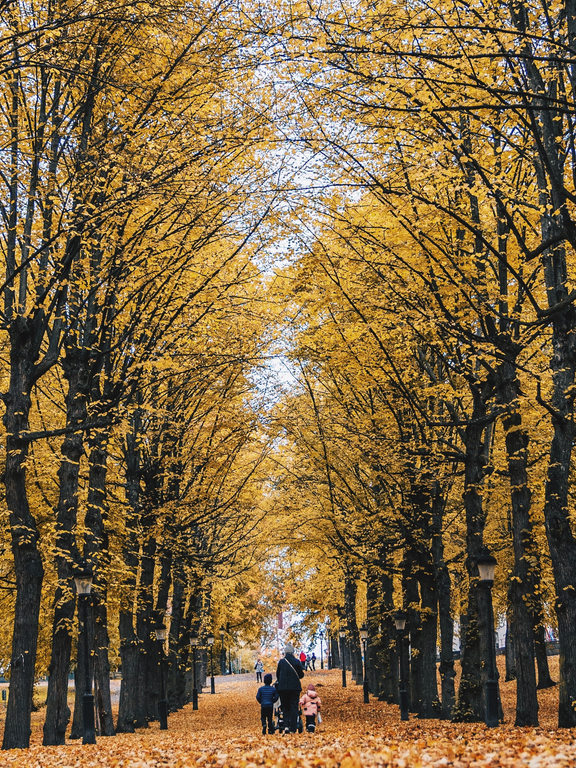}
        \end{subfigure}
        &
        \begin{subfigure}{0.245\textwidth}
            \centering
            \includegraphics[width=\linewidth]{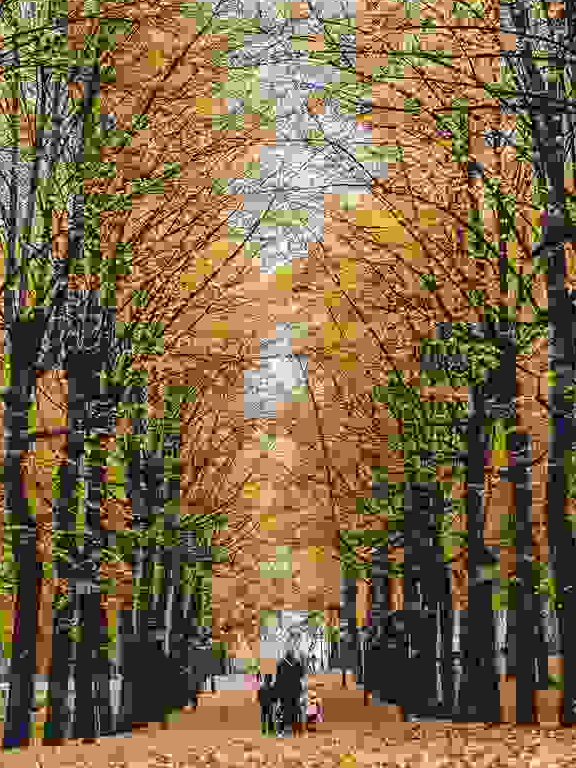}
        \end{subfigure}        
        &
        \begin{subfigure}{0.245\textwidth}
            \centering
            \includegraphics[width=\linewidth]{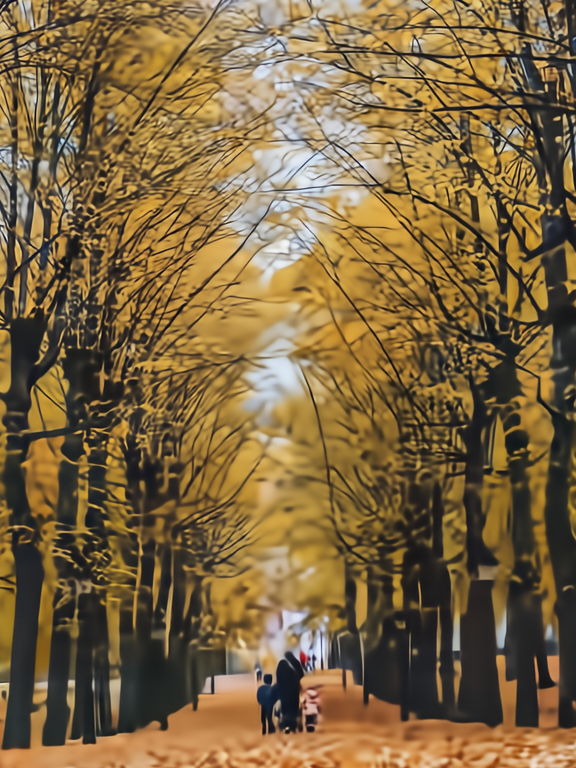}
        \end{subfigure}
        &
        \begin{subfigure}{0.245\textwidth}
            \centering
            \includegraphics[width=\linewidth]{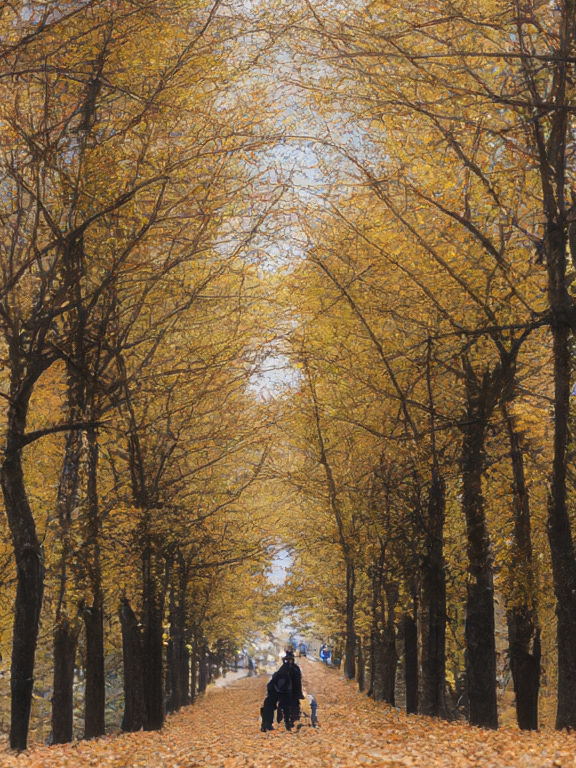}
        \end{subfigure}
        \\
        \texttt{8d8e9} & $0.17050$~bpp ($\times 32.85$) & $0.13225$~bpp ($\times 25.48$) & $0.00784$~bpp ($\times 1.51$) \\

        \midrule
        DLF & StableCodec & AEIC-ME & PerCoV2 \\
        (ICCV'25) & (ICCV'25) & (CVPR'26) & (Ours)\\
        \midrule
        
        \begin{subfigure}{0.245\textwidth}
            \centering
            \includegraphics[width=\linewidth]{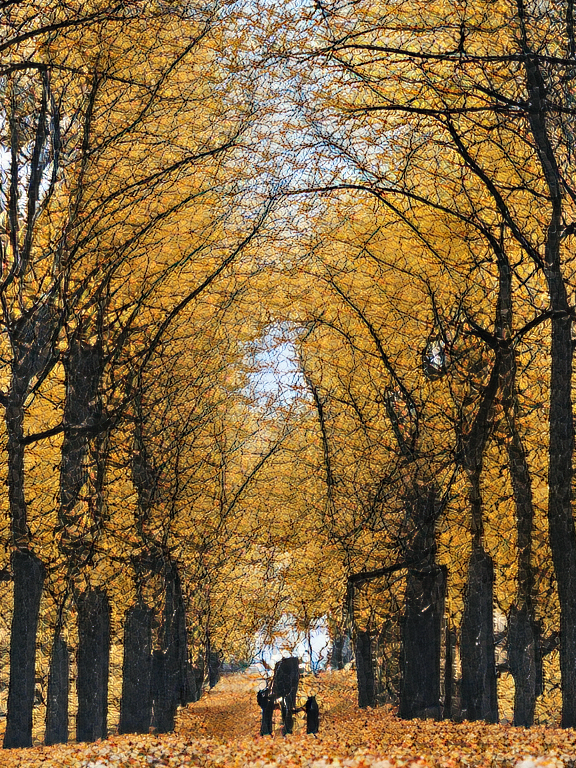}
        \end{subfigure}
        &
        \begin{subfigure}{0.245\textwidth}
            \centering
            \includegraphics[width=\linewidth]{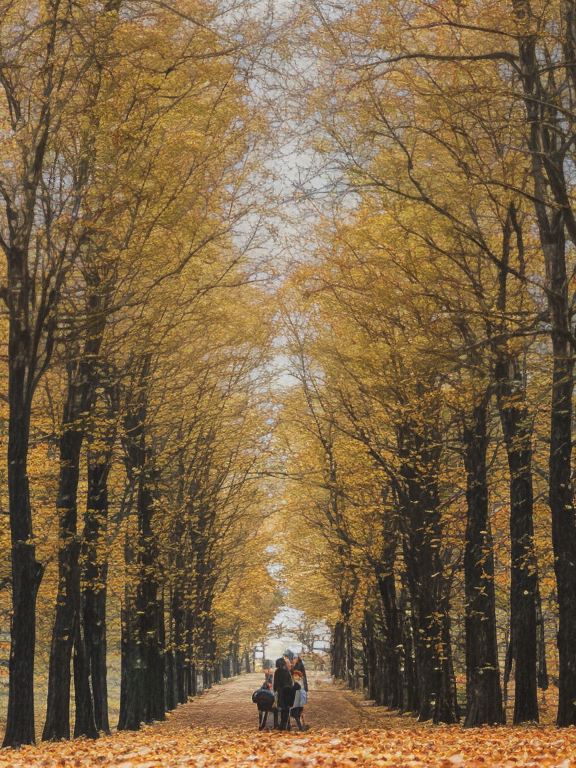}
        \end{subfigure}        
        &
        \begin{subfigure}{0.245\textwidth}
            \centering
            \includegraphics[width=\linewidth]{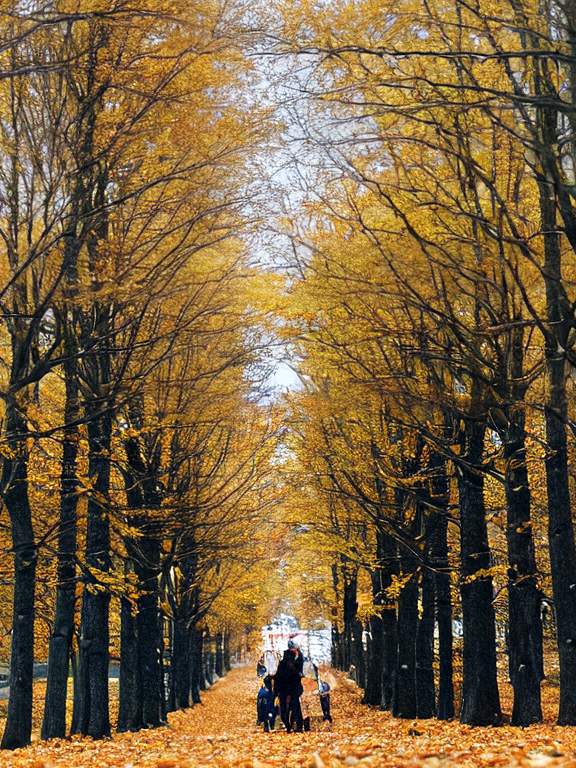}
        \end{subfigure}
        &
        \begin{subfigure}{0.245\textwidth}
            \centering
            \includegraphics[width=\linewidth]{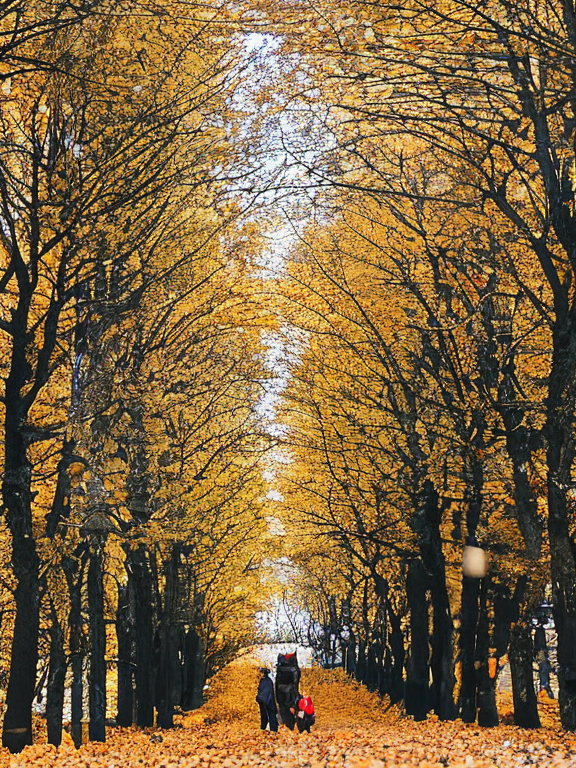}
        \end{subfigure}
        \\
        $0.00902$~bpp ($\times 1.74$) & $0.006$~bpp ($\times 1.16$) & $0.00591$~bpp ($\times 1.14$) & $0.00519$~bpp \\
        \bottomrule

    \end{tabular}

    \caption{Visual comparison on CLIC 2020. We show $768 \times 576$ crops taken from high-resolution images of size $2048 \times 1152$; bit-rate increases relative to PerCoV2 are shown as $(\times)$. \textbf{Best viewed zoomed-in.}}
    \label{fig:high_res2}
\end{figure*}


\textbf{Implementation Details.} We base PerCoV2 on the open-source implementation of PerCo~\citep{koerber2024perco} and the \texttt{diffusers} library~\citep{von-platen-etal-2022-diffusers}. For training, we utilize a distinct subset of the COCO 2017 training set~\citep{caesar2018cvpr} and the SA-1B dataset~\citep{Kirillov_2023_ICCV} for fixed- and high-resolution (2K) evaluations, comprising 93,287 and 11M images, respectively. For fixed-resolution evaluation, we additionally consider a global TiTok-variant, in which the whole image is mapped to a single 1D latent token sequence (as opposed to the patched variant) to achieve even lower bit-rates, yielding a total of 5 evaluation points. Training is performed on two H100 GPUs using mixed-precision computation.

\textbf{Resolution-adaptive 1D tokenizer.} We derive our finetuning configurations from the official TiTok \citep{yu2024an} repository, specifically the \texttt{titok\_l32}, \texttt{titok\_b64}, and \texttt{titok\_s128} setups found at \url{https://github.com/bytedance/1d-tokenizer/tree/main/configs/training/TiTok}. All hyper-parameters remain fixed unless otherwise noted\footnote{In TiTok, the configurations \texttt{titok\_l32}, \texttt{titok\_b64}, and \texttt{titok\_s128} correspond to different complexity scales ($307$M, $86$M, and $22$M parameters for encoder and decoder, respectively), with more computational resources allocated to the lower bit-rates. For simplicity, we did not explore larger configurations, although we expect further compression gains, particularly at higher bit-rates, by scaling up the tokenizer.}. For both stage 1 and stage 2, we employ $100$k optimization steps with a global batch size of $16$ for fixed-resolution training, and $250$k steps with a batch size of $8$ for high-resolution training, per stage.

\textbf{1D~MIM.} We use $500$k optimization steps with a global batch size of $64$ for fixed-resolution training, and $1$M steps with a batch size of $16$ for high-resolution training. Following previous work, we employ the AdamW~\citep{loshchilov2018decoupled} optimizer with a weight decay of $0.03$ and a maximum gradient norm of $1.0$. The learning rate is initialized at $2 \times 10^{-4}$ with $10$k warmup steps, following a cosine schedule that decays to $1 \times 10^{-5}$. Training is performed using \texttt{bf16} mixed precision and TF32 for numerical efficiency.

\textbf{PerCoV2.} We train PerCoV2 using the SANA 1.6B diffusion model~\citep{xie2025sana} as the base architecture. The model is trained for $50$k steps with a constant learning rate of $1 \times 10^{-4}$, preceded by $10$k warmup steps. We utilize the AdamW~\citep{loshchilov2018decoupled} optimizer with a weight decay of $0.01$ and a maximum gradient norm of $1.0$. Training is conducted at a resolution of $512 \times 512$ with a global batch size of $40$, employing \texttt{bf16} mixed precision, TF32, and gradient checkpointing to optimize memory efficiency. Standard data augmentations, including center cropping and random horizontal flipping, are applied.

For high-resolution evaluation, we train PerCoV2 using the SANA 0.6B diffusion model~\citep{xie2025sana} as the base architecture without multimodal enhancement, as we did not observe further improvements from its inclusion. Training is performed natively on high-resolution images for $300$k-$1$M steps using aspect-ratio bucketing and a reduced global batch size of $8$, while keeping all other hyper-parameters unchanged.

\textbf{PerCoV2-distilled.} We derive PerCoV2-distilled from the PerCoV2 (SANA 0.6B) teacher model without multimodal enhancement, applying LoRA with rank $r=32$ and scaling factor $\alpha=64$ to obtain a faster and more lightweight variant. As a proof of concept, we limit our study to a single operating point with fixed-resolution training, leaving a broader investigation across resolutions and bit-rates for future work.

For optimization, we first train the model using only the MSE and LPIPS losses for $200$k steps. We then gradually introduce the GAN loss for an additional $100$k steps, followed by the DMD loss for the final $100$k steps. During DMD training, we additionally train a fake velocity model in parallel, following the setup of CoD/CoD-Lite~\citep{jia2025cod, jia2026codlite}. Specifically, following the DMD procedure, we update the one-step model every $10$ steps, while using the remaining $9$ steps to train the fake velocity model. We set $\lambda_{\mathrm{DMD}}=0.1$. All other hyper-parameters are kept consistent with the PerCoV2 configuration.

\textbf{Evaluation Setup.} We follow the official evaluation protocol of PerCo~\citep{careil2024towards}, assessing performance on the Kodak~\citep{kodak} and MSCOCO-30k~\citep{caesar2018cvpr} datasets at a resolution of $512\times512$. These datasets contain 24 and 30k images, respectively. We report FID~\citep{NIPS2017_8a1d6947} and KID~\citep{bińkowski2018demystifying} to quantify statistical fidelity, PSNR, MS-SSIM~\citep{msssim2003}, and LPIPS~\citep{Zhang_2018_CVPR} to quantify pairwise distortion, including learned perceptual distortion, CLIP-score~\citep{hessel-etal-2021-clipscore} to measure global alignment between reconstructed images and ground-truth captions, and mean intersection over union (mIoU) to assess semantic preservation~\citep{schoenfeld2021you}. For the latter, we use the ViT-Adapter segmentation network~\citep{chen2023vision}. Bit-rate saving is measured by BD-Rate~\citep{bjontegaard2001calculation}. All evaluations are performed on a single H100 GPU.

We further evaluate the performance on high-resolution benchmarks (DIV2K~\citep{Agustsson_2017_CVPR_Workshops} and CLIC 2020~\citep{CLIC2020}).

\textbf{Baselines.} We compare PerCoV2 against a broad range of learned and conventional compression methods. We include PerCoV1++ (SD v3.0), our initial effort toward an open flow-based reimplementation of the PerCo architecture (\cref{sec:percov1}), considering two variants: PerCoV1 with index coding and PerCoV1++ augmented with an entropy model, as well as MS-ILLM~\citep{pmlr-v202-muckley23a}, DiffC~\citep{vonderfecht2025lossy}, PerCo~\citep{careil2024towards, koerber2024perco}, DiffEIC~\citep{li2024towards}, RDEIC~\citep{li2025rdeic}, GLC~\citep{Jia_2024_CVPR}, DLF~\citep{xue2025dlf}, StableCodec~\citep{Zhang_2025_ICCV}, AEIC-ME~\citep{Zhang_2026_CVPR}, OneDC~\citep{xue2025one}, CoD~\citep{jia2025cod}, CoD-Lite~\citep{jia2026codlite}, and DiT-IC~\citep{Shi_2026_CVPR}. For PerCo, we evaluate both the official implementation, if possible, and the community reimplementation~\citep{koerber2024perco}. For DiffC, we use the Stable Diffusion v1.5 backbone instead of v2.1, as it better matches the target image distribution ($512\times512$px). We additionally evaluate DiffC with CoD~\citep{jia2025cod}; by default, we use CoD latent, as it provides improved statistical fidelity over its pixel-space counterpart (CoD pixel). In contrast, CoD-Lite is derived as a real-time variant based on CoD pixel.

We further provide visual comparisons with JPEG, a widely adopted non-learned image codec, as well as DCVC-RT-intra~\citep{jia2025towards}, one of the strongest recent learned video compression methods. While these approaches achieve excellent distortion-based quality, particularly in terms of PSNR, they primarily target the distortion component of the RDP trade-off and are not directly optimized for preserving statistical fidelity. Additionally, they typically do not operate in the ultra-low bit-rate regime targeted by PerCoV2. Therefore, we include these methods for completeness and qualitative reference only.

Finally, we report the auto-encoder reconstruction bounds of the underlying generative models, namely \textit{SD v2.1 auto-encoder}, \textit{SD v3 auto-encoder}, and \textit{SANA auto-encoder}, corresponding to PerCo (SD)~\citep{koerber2024perco}, PerCoV1++ (SD v3.0), and PerCoV2, respectively.

\subsection{Main Results}\label{subsec:main_results}

\begin{table}[tb]
\caption{BD-Rate (\%) $\downarrow$ comparison on MSCOCO-30k, grouped into VAE-based, one-step diffusion, and multi-step diffusion methods. PerCoV2 (Ours) is used as the anchor.}
\label{tab:bd}
\centering
\renewcommand{\arraystretch}{1.0}
\setlength{\tabcolsep}{0.8pt}
\begin{tabular}{l|cccccc}
\toprule
Method & FID & KID & LPIPS & MS-SSIM & mIoU & CLIP \\
\midrule
MS-ILLM & \pos{$2708.8$} & \pos{$2173.8$} & -- & \nega{$-29.5$} & \pos{$318.0$} & \pos{$181.9$} \\
DLF & -- & -- & \nega{$-31.5$} & \nega{$-20.3$} & \pos{$34.0$} & \pos{$51.9$} \\
\midrule
CoD-Lite & \pos{$5018.2$} & -- & \nega{$-19.8$} & \nega{$-32.1$} & \pos{$52.8$} & \pos{$216.0$} \\
OneDC & \pos{$712.3$} & \pos{$874.6$} & -- & -- & \nega{$-37.6$} & \pos{$36.8$} \\
StableCodec & \pos{$454.4$} & \pos{$154.3$} & \nega{$-25.2$} & \nega{$-43.5$} & \pos{$27.9$} & \pos{$188.7$} \\
AEIC & \pos{$303.5$} & \pos{$79.2$} & \nega{$-28.8$} & \nega{$-42.5$} & \pos{$44.2$} & \pos{$197.3$} \\
\midrule
DiffEIC & \pos{$2183.5$} & \pos{$1618.3$} & \pos{$140.8$} & \pos{$47.0$} & \pos{$260.6$} & \pos{$722.8$} \\
DiffC (CoD) & \pos{$2126.3$} & -- & \pos{$13.3$} & \nega{$-6.2$} & \nega{$-5.5$} & \pos{$104.3$} \\
DiffC (SD) & \pos{$1302.0$} & \pos{$832.2$} & \pos{$71.3$} & \pos{$22.5$} & \pos{$141.9$} & \pos{$63$} \\
PerCoV1 & \pos{$1285.5$} & \pos{$900.6$} & \pos{$187.3$} & \pos{$89.8$} & \pos{$233.4$} & \pos{$271.9$} \\
\rowcolor{verylightgray} PerCoV2 & $0.0$ & $0.0$ & $0.0$ & $0.0$ & $0.0$ & $0.0$ \\

\bottomrule
\end{tabular}
\end{table}

\textbf{Fixed-Resolution (PerCoV1) Setting.} On MSCOCO-30k, PerCoV2 achieves the strongest statistical fidelity, with the lowest FID and KID across the evaluated bit range (cf.~\cref{fig:quant_comparison,tab:bd}). The advantage is particularly pronounced at ultra-low bit-rates, where competing methods exhibit a marked degradation in statistical fidelity. Consistent with earlier findings~\citep{Agustsson_2019_ICCV,careil2024towards}, pixel-wise measures such as PSNR provide limited insight at these compression levels: methods with strong distortion scores can nonetheless introduce structured artifacts, including grid patterns and repetitive textures, that are weakly penalized by conventional metrics but clearly visible in qualitative comparisons (cf.~\cref{fig:vis_impressions,fig:vis_impressions_2,fig:vis_impressions_2_ext1,fig:vis_impressions_2_ext2}). For instance, DiffC~\citep{vonderfecht2025lossy} produces semantically inconsistent structures, while MS-ILLM~\citep{pmlr-v202-muckley23a} tends toward overly smooth and desaturated reconstructions. In contrast, PerCoV2 preserves coherent textures and semantic structure, consistent with its superior statistical fidelity.

Notably, PerCoV2 improves statistical fidelity over AEIC, requiring on average $75.2\%$ and $44.2\%$ fewer bits for comparable FID and KID, respectively. Compared to PerCoV1~\citep{careil2024towards,koerber2024perco}, PerCoV2 further improves distortion-oriented performance, reducing the average bit-rate required for comparable LPIPS and MS-SSIM by $65.2\%$ and $47.3\%$, respectively. Nevertheless, PerCoV2 remains behind VAE-based and one-step diffusion methods, such as MS-ILLM and OneDC, on these metrics. This can be attributed in part to the inherent distortion--perception trade-off~\citep{ICML-2019-BlauM}. It is also worth noting that these methods explicitly optimize for (perceptual) distortion, including LPIPS, whereas PerCoV2/SANA is trained solely using a latent flow-matching objective. Importantly, improved statistical fidelity and distortion alignment also translate to stronger segmentation performance and better semantic correspondence with ground-truth captions  (cf.~\cref{fig:vis_segmentation,tab:bd}), demonstrating practical relevance beyond visual appearance.

On Kodak, we observe similar trends, with PerCoV2 being particularly effective at ultra-low bit-rates. We report patch-based FID following HiFiC~\citep[A.7]{mentzer2020high} with $64\times64$ crops, as the 24 Kodak images alone provide too few samples for a stable FID estimate. The patch-based FID can nevertheless exhibit different trends and stronger correlation with distortion errors than the global FID computed on MSCOCO-30k. At higher bit-rates, PerCoV2 is somewhat less effective in terms of FID/64, arguably due in part to differences in the reconstruction-quality upper bound of the underlying auto-encoder, with the SANA-1.5 auto-encoder lagging behind the SD-based variants.

\textbf{High-Resolution Setting.} We limit this section to the smaller Sana-0.6B backbone as a side study. PerCoV2 further demonstrates strong zero-shot generalization to widely adopted high-resolution benchmarks, including DIV2K and CLIC 2020, outperforming AEIC-ME in terms of FID while achieving slightly weaker KID scores (cf.~\cref{fig:quant_comparison_highres,fig:high_res1,fig:high_res2}). Notably, PerCoV2 is trained solely on SA-1B with aspect-ratio bucketing and evaluated without further domain adaptation. In contrast, StableCodec~\citep{Zhang_2025_ICCV} and AEIC~\citep{Zhang_2026_CVPR} train directly on the CLIC 2020 and DIV2K training partitions, while CoD/CoD-Lite~\citep{jia2025cod, jia2026codlite} use a broader mix of datasets, including ImageNet~\citep{ILSVRC15}, OpenImages~\citep{OpenImages}, and SA-1B~\citep{Kirillov_2023_ICCV}. We leave a systematic study of training-resolution effects and domain-specific adaptation to future work.

\textbf{PerCoV2-distilled.} We provide a quantitative comparison of PerCoV2-distilled at $0.0029$~bpp in~\cref{fig:teaser}, alongside our strongest PerCoV2 model; the teacher model is omitted from the figure for clarity. A qualitative comparison is provided in~\cref{fig:vis_segmentation}. PerCoV2-distilled largely preserves the perceptual compression performance of its teacher model, with only a $0.54$-point increase in FID, while substantially improving distortion-oriented metrics, achieving a $1.3$~dB gain in PSNR and a $0.06$ improvement in MS-SSIM. These results demonstrate the general feasibility of distilling PerCoV2, retaining strong statistical fidelity and substantially outperforming both PerCoV1 and CoD (latent) at this bit-rate.

We note that our current setup employs a relatively simple PatchGAN discriminator and is intended only as a preliminary investigation of distillation. More advanced adversarial formulations, such as those incorporating VQ-VAE-style~\citep{pmlr-v202-muckley23a} or semantic discriminators~\citep{10.1007/978-3-031-72761-0_12}, could potentially further improve the resulting rate--distortion--perception trade-off.

\begin{table*}[tb]
  \caption{Ablation of PerCoV2 components across complexity tiers/ bit-rates}
  \label{tab:unified_modular_ablation}
  \centering
  \renewcommand{\arraystretch}{1.0}
  \setlength{\tabcolsep}{10pt} 
  \begin{tabular}{lcc | cc | cc}
    \toprule
    Config / Component & {Bpp $\downarrow$} & {Sav. (\%) $\uparrow$} & FID $\downarrow$ & $\Delta\%$ & KID $\downarrow$ & $\Delta\%$ \\
    \midrule
    \rowcolor{verylightgray}
    Base (Image Tokenizer) & $0.00586$ & -- & $2.771$ & -- & $0.00081$ & -- \\
    \quad + Flow Model (SANA) & $0.00586$ & $0.00$ & $2.498$ & \pos{$9.85$} & $0.00059$ & \pos{$27.16$} \\
    \quad \quad + Multim. Enh. & $0.00586$ & $0.00$ & $2.198$ & \pos{$20.6$} & $0.00035$ & \pos{$56.8$} \\ 
    \quad \quad \quad + Entropy Model & $0.00565$ & \pos{$3.58$} & $2.198$ & $0.00$ & $0.00035$ & $0.00$ \\
    \midrule
    \rowcolor{verylightgray}
    
    Base (Image Tokenizer) & $0.01172$ & -- & $1.824$ & -- & $0.00037$ & -- \\
    \quad + Flow Model (SANA) & $0.01172$ & $0.00$ & $1.697$ & \pos{$6.96$} & $0.00015$ & \pos{$59.46$} \\
    \quad \quad + Multim. Enh. & $0.01172$ & $0.00$ & $1.611$ & \pos{$11.68$} & $0.00012$ & \pos{$67.57$} \\ 
    \quad \quad \quad + Entropy Model & $0.01094$ & \pos{$6.66$} & $1.611$ & $0.00$ & $0.00012$ & $0.00$ \\    

    \midrule
    \rowcolor{verylightgray}
        \rowcolor{verylightgray}
    Base (Image Tokenizer) & $0.02344$ & -- & $1.400$ & -- & $0.00014$ & -- \\
    \quad + Flow Model (SANA) & $0.02344$ & $0.00$ & $1.422$ & \nega{$-1.57$} & $0.00010$ & \pos{$28.57$} \\
    \quad \quad + Multim. Enh. & $0.02344$ & $0.00$ & $1.365$ & \pos{$2.5$} & $0.00007$ & \pos{$50.0$} \\ 
    \quad \quad \quad + Entropy Model & $0.02158$ & \pos{$7.94$} & $1.365$ & $0.00$ & $0.00007$ & $0.00$ \\ 
    \bottomrule
  \end{tabular}
\end{table*}

\begin{figure}[bt]
    \setlength{\tabcolsep}{1.0pt}  
    \renewcommand{\arraystretch}{1.0}  
    \centering
    \scriptsize
    \begin{tabular}{c|cccc}
        \toprule
        Original & PerCoV1 & PerCoV2 & PerCoV2 \\
        & (ICLR'24) & (Ours) & (Ours, dist.) \\
        \midrule
  
        \begin{subfigure}{0.114\textwidth}
            \centering
            \includegraphics[width=\linewidth,height=\linewidth]{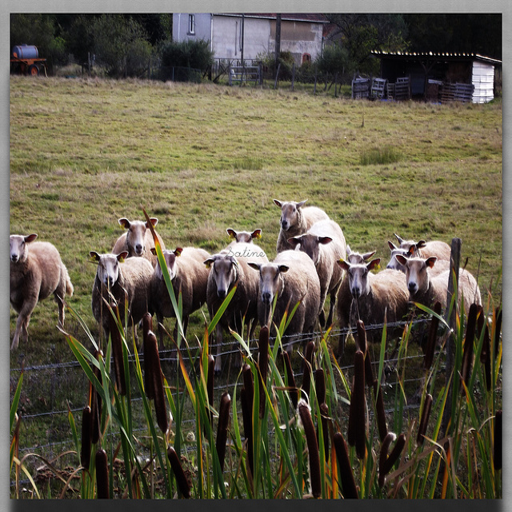}
        \end{subfigure}
        &
        \begin{subfigure}{0.114\textwidth}
            \centering
            \includegraphics[width=\linewidth,height=\linewidth]{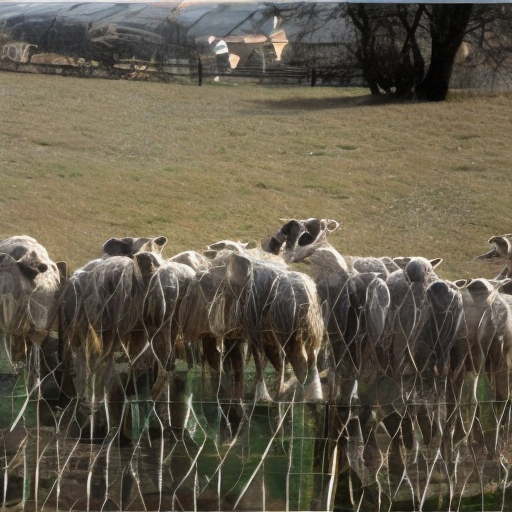}
        \end{subfigure}
        &
        \begin{subfigure}{0.114\textwidth}
            \centering
            \includegraphics[width=\linewidth,height=\linewidth]{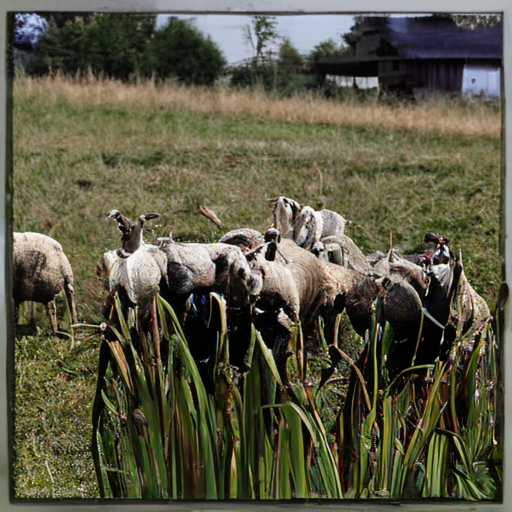}
        \end{subfigure}
        &
        \begin{subfigure}{0.114\textwidth}
            \centering
            \includegraphics[width=\linewidth,height=\linewidth]{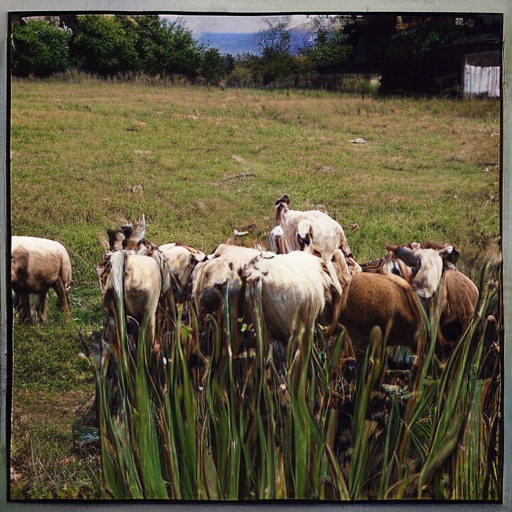}
        \end{subfigure}
        \\ 
        GT label & Prediction & Prediction & Prediction \\
        \begin{subfigure}{0.114\textwidth}
            \centering
            \includegraphics[width=\linewidth,height=\linewidth]{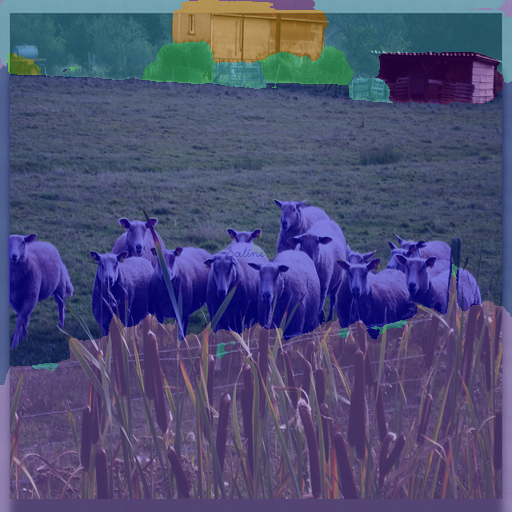}
        \end{subfigure}
        &
        \begin{subfigure}{0.114\textwidth}
            \centering
            \includegraphics[width=\linewidth,height=\linewidth]{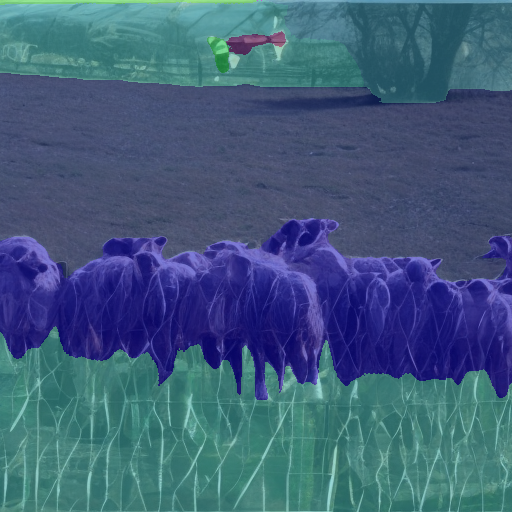}
        \end{subfigure}
        &
        \begin{subfigure}{0.114\textwidth}
            \centering
            \includegraphics[width=\linewidth,height=\linewidth]{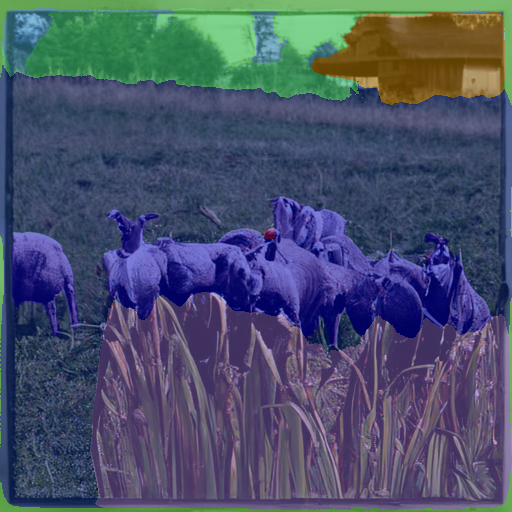}
        \end{subfigure}
        &
        \begin{subfigure}{0.114\textwidth}
            \centering
            \includegraphics[width=\linewidth,height=\linewidth]{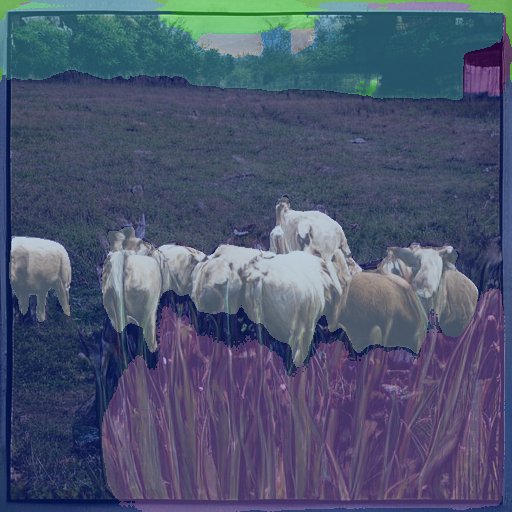}
        \end{subfigure}
        
        \\ 
        \texttt{442539} & $0.00363$~bpp & $0.00293$~bpp & $0.00293$~bpp \\
        & ($1.24\times$) & & 
        \\
        \bottomrule
        
    \end{tabular}

    \caption{Semantic preservation on MSCOCO-30k. Bit-rate increases are indicated by $(\times)$. \textbf{Best viewed electronically.}}
    \label{fig:vis_segmentation}
\end{figure}

\begin{figure}[tb]
    \setlength{\tabcolsep}{1pt} 
    \renewcommand{\arraystretch}{1.0}
    \centering
    \scriptsize
    \begin{tabular}{cccc}  

        \multicolumn{4}{c}{\begin{subfigure}{0.47\textwidth}
            \centering
            \includegraphics[width=\linewidth]{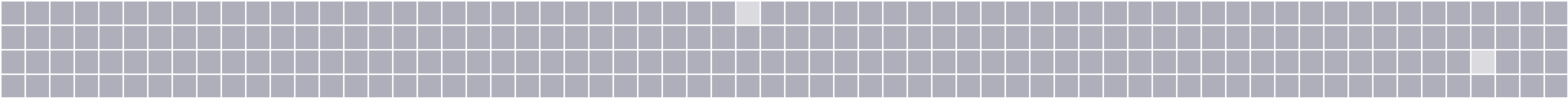}
        \end{subfigure}}\\
        \multicolumn{4}{c}{\begin{subfigure}{0.47\textwidth}
            \centering
            \includegraphics[width=\linewidth]{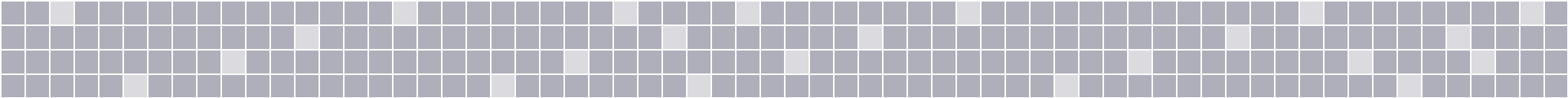}
        \end{subfigure}}\\
        \multicolumn{4}{c}{\begin{subfigure}{0.47\textwidth}
            \centering
            \includegraphics[width=\linewidth]{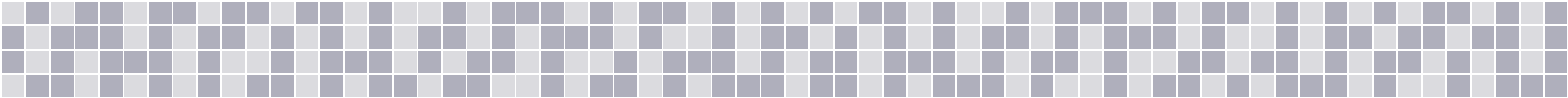}
        \end{subfigure}}\\
        \multicolumn{4}{c}{\begin{subfigure}{0.47\textwidth}
            \centering
            \includegraphics[width=\linewidth]{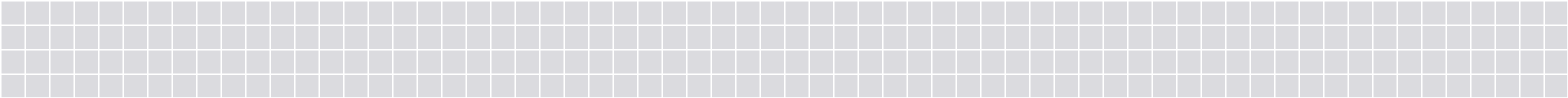}
        \end{subfigure}}\\
        
        \begin{subfigure}{0.114\textwidth}
            \centering
            \includegraphics[width=\linewidth]{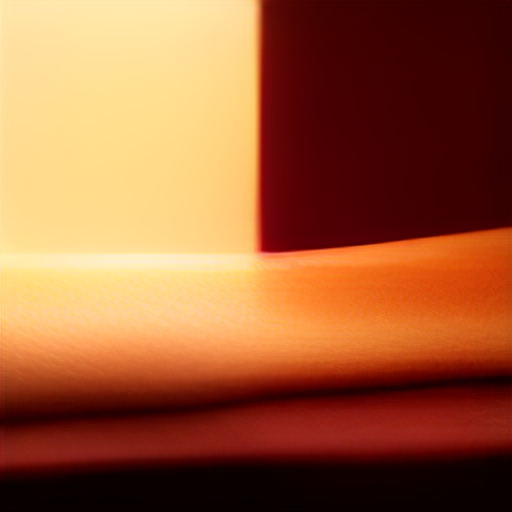}
        \end{subfigure} &
        \begin{subfigure}{0.114\textwidth}
            \centering
            \includegraphics[width=\linewidth]{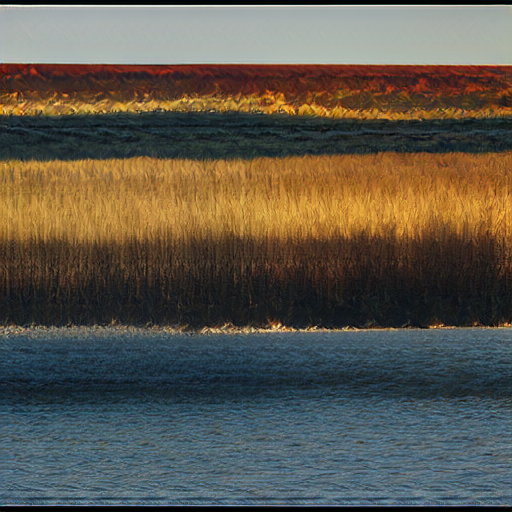}
        \end{subfigure} &
        \begin{subfigure}{0.114\textwidth}
            \centering
            \includegraphics[width=\linewidth]{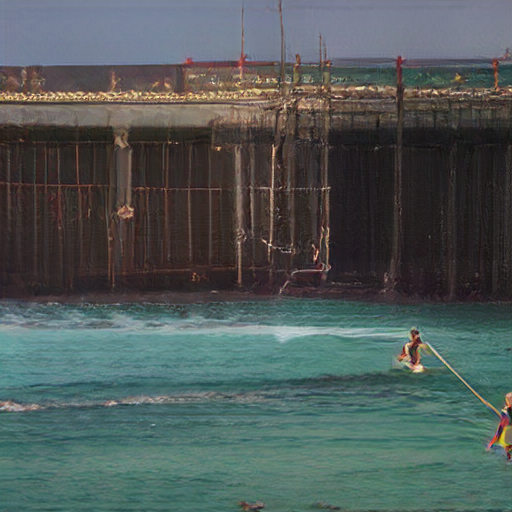}
        \end{subfigure} &
        \begin{subfigure}{0.114\textwidth}
            \centering
            \includegraphics[width=\linewidth]{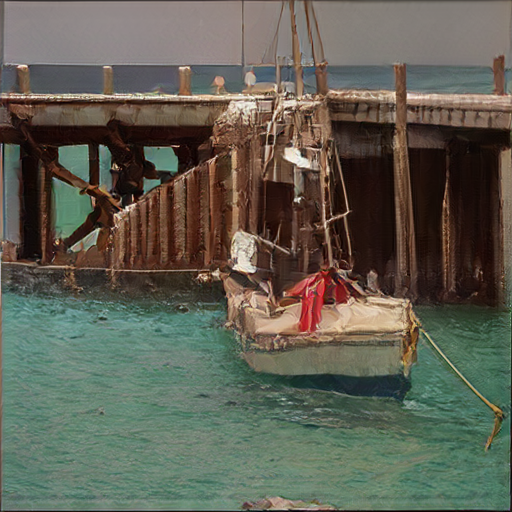}
        \end{subfigure} \\
        $2/256$ & $23/256$ & $105/256$ & $256/256$ \\ 
        $0.00012$~bpp & $0.00111$~bpp & $0.00477$~bpp & $0.01092$~bpp \\    

    \end{tabular}
    \caption{Visualization of the generative capabilities of our hybrid 1D MIM. Revealed tokens are shown, while the remaining tokens are generated.}
    \label{fig:beyond_compression_tokenizer}
\end{figure}

\subsection{Ablations}\label{subsec:main_ablations}

\textbf{Modular Component Analysis.} \Cref{tab:unified_modular_ablation} details the impact of each architectural component across three configurations: large (l32), medium (b64), and small (s128). We observe that our components are most effective at ultra-low bit-rates, where the generative prior plays a more critical role. For instance, in the l32 configuration ($0.00586$~bpp), the integration of the SANA flow model and multimodal enhancement yields a combined FID improvement of over $20$\%. As the bit-rate increases (s128), these relative gains moderate, as more direct reconstruction information is available in the bit-stream. Furthermore, our entropy model provides consistent bit-rate savings, reaching $7.94$\% in the small configuration without any degradation in visual fidelity.

\begin{table*}[tb]
  \caption{Global semantic guidance ablation. Gains ($\Delta\%$) are w.r.t.\ the No Text baseline.}
  \label{tab:global_configs}
  \centering
  \renewcommand{\arraystretch}{1.0}
  \setlength{\tabcolsep}{7pt} 
  \begin{tabular}{l | cc | cc | cc}
    \toprule
    Strategy & FID $\downarrow$ & $\Delta\%$ & KID $\downarrow$ & $\Delta\%$ & LPIPS $\downarrow$ & $\Delta\%$ \\
    \midrule
    \rowcolor{verylightgray} SANA (No Text) & $3.636$ & -- & $0.00008$ & -- & $0.486$ & -- \\
    \quad + Fixed Text~\citep{Zhang_2025_ICCV} & 3.703 & \nega{$-5.89$} & $0.00078$ & \pos{$2.50$} & $0.489$ & \nega{$-0.62$} \\
    \quad + Multim. Enh. (Ours) & {\boldmath $3.496$} & \pos{\boldmath $3.85$} & {\boldmath $0.00067$} &  \pos{\boldmath $16.25$} & {\boldmath $0.488$} &  \nega{\boldmath $-0.41$} \\
    \quad + GT Text (Up. Bound) & $2.544$ & \pos{$30.03$} & $0.00023$ & \pos{$71.25$} & $0.480$ & \pos{$1.23$} \\
    \midrule
    \rowcolor{verylightgray} SANA (No Text) & $2.473$ & -- & $0.00028$ & -- & $0.417$ & -- \\
    \quad + Fixed Text~\citep{Zhang_2025_ICCV} & $2.474$ & \nega{$-0.04$} & $0.00029$ & \nega{$-3.57$} & $0.420$ & \nega{$-0.72$} \\
    \quad + Multim. Enh. (Ours) & {\boldmath $2.268$} & \pos{\boldmath $8.29$} & {\boldmath $0.00016$} & \pos{\boldmath $42.86$} & {\boldmath $0.415$} & \pos{\boldmath $0.48$} \\
    \quad + GT Text (Up. Bound) & $2.06$ & \pos{$16.70$} & $0.00016$ & \pos{$42.86$} & $0.411$ & \pos{$1.44$} \\
    \midrule
    \rowcolor{verylightgray} SANA (No Text) & $2.072$ & -- & $0.00028$ & -- & $0.364$ & -- \\
    \quad + Fixed Text~\citep{Zhang_2025_ICCV} & $2.122$ & \nega{$-2.41$} & $0.00032$ & \nega{$-14.29$} & $0.364$ & $0.00$ \\
    \quad + Multim. Enh. (Ours) & {\boldmath $1.959$} & \pos{\boldmath $5.45$} & {\boldmath $0.00022$} & \pos{\boldmath $21.43$} & {\boldmath $0.361$} & \pos{\boldmath $0.82$} \\
    \quad + GT Text (Up. Bound) & $1.819$ & \pos{$12.21$} & $0.00012$ & \pos{$57.14$} & $0.359$ & \pos{$1.37$} \\
    \bottomrule
  \end{tabular}
\end{table*}

\textbf{Impact of Global Semantic Guidance.} \Cref{tab:global_configs} evaluates different text-guidance strategies across varying global configurations ($0.00146$~bpp, $0.00293$~bpp, $0.00586$~bpp). Across all bit-rates, utilizing a fixed text prompt provides negligible benefits and can even slightly degrade performance. In contrast, multimodal enhancement consistently improves FID and KID, with the most pronounced gains occurring at lower bit-rates (e.g., a $42.86$\% KID improvement at $0.00293$~bpp). While a gap remains relative to the Ground Truth (GT) text upper bound, these results demonstrate that adaptive semantic guidance is essential for pushing the perception frontier at extreme compression levels.

\begin{table}[tb]
\caption{Loss ablation. Relative improvements ($\Delta\%$) are reported w.r.t.\ VQ + \text{CFM+}.}
\label{tab:loss_ablation}
\centering
\renewcommand{\arraystretch}{1.0}
\setlength{\tabcolsep}{5pt}
\begin{tabular}{l|cc}
\toprule
Method & FID $\downarrow$ & $\Delta$ (\%) \\
\midrule

\rowcolor{verylightgray} VQ + \text{CFM+} (noise, N2N) & 50.165 & -- \\

VQ + Proxy + \text{CFM+} (noise, N2N) & 5.339 & \pos{$89.36$} \\

\text{CFM+} (noise, frozen VQ) & 4.771 & \pos{$90.49$} \\

\text{CFM+} (semantic, frozen VQ) & \textbf{4.186} & \pos{\boldmath $91.65$} \\

\bottomrule
\end{tabular}
\end{table}

\textbf{Loss Formulation Ablations.} We ablate the loss formulation in~\cref{tab:loss_ablation}. Joint end-to-end optimization using only \text{CFM+} and VQ losses is highly unstable. Introducing proxy-code supervision, learning to predict discrete indices from an off-the-shelf VQGAN, substantially stabilizes training and improves both perceptual and distortion metrics. Freezing the tokenizer further simplifies the objective and yields additional gains. Finally, replacing noise initialization with noised semantic initialization and learning the mapping to the ground-truth image leads to our final formulation, achieving the best overall performance.

\begin{figure}[bt]
    \setlength{\tabcolsep}{1.0pt}  
    \renewcommand{\arraystretch}{1.0}  
    \centering
    \scriptsize
    \begin{tabular}{cc|cc}

        \begin{subfigure}{0.11\textwidth}
            \centering
            \includegraphics[width=\linewidth]{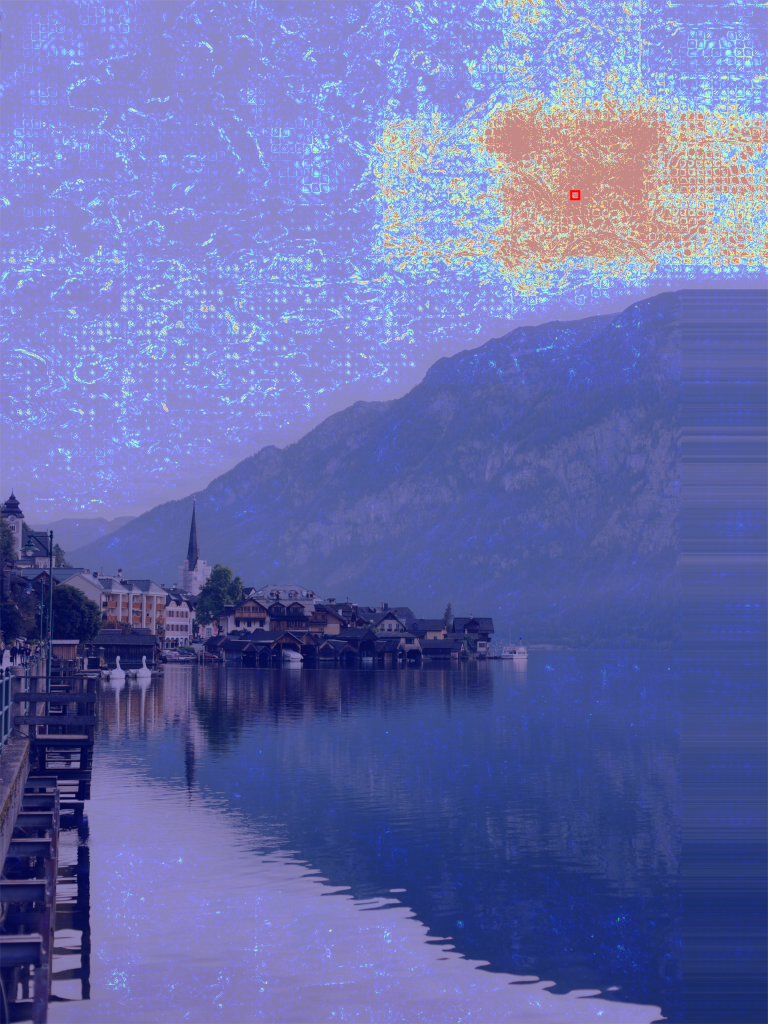}
        \end{subfigure}
        &
        \begin{subfigure}{0.11\textwidth}
            \centering
            \includegraphics[width=\linewidth]{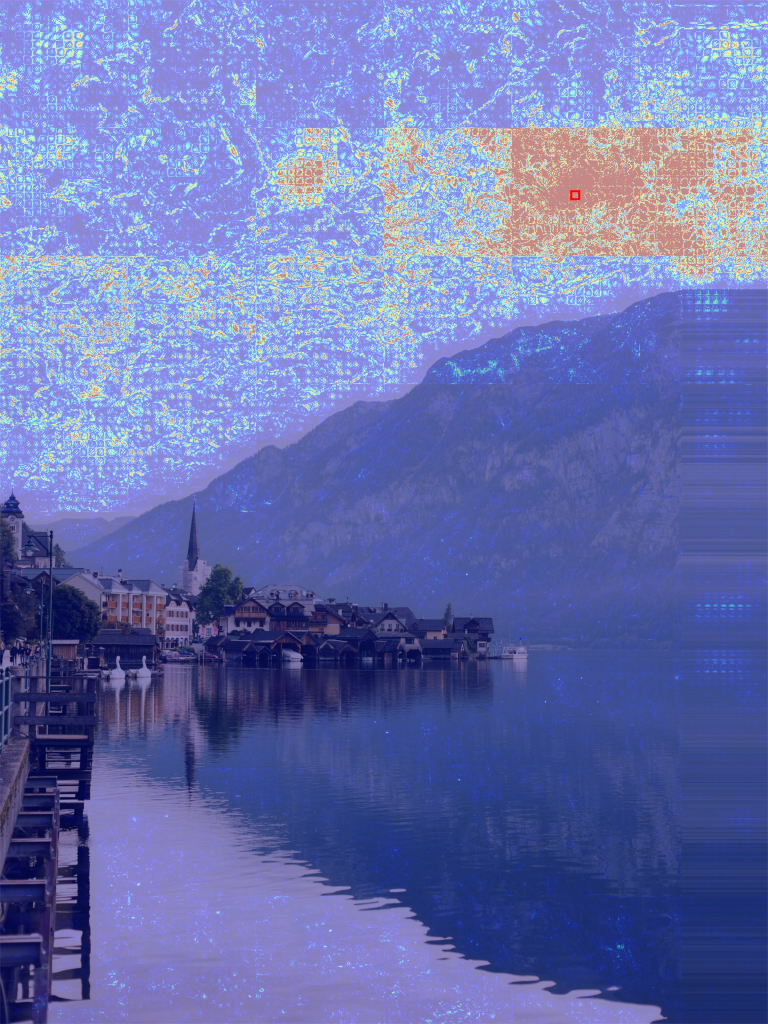}
        \end{subfigure}
        &
        \begin{subfigure}{0.11\textwidth}
            \centering
            \includegraphics[width=\linewidth]{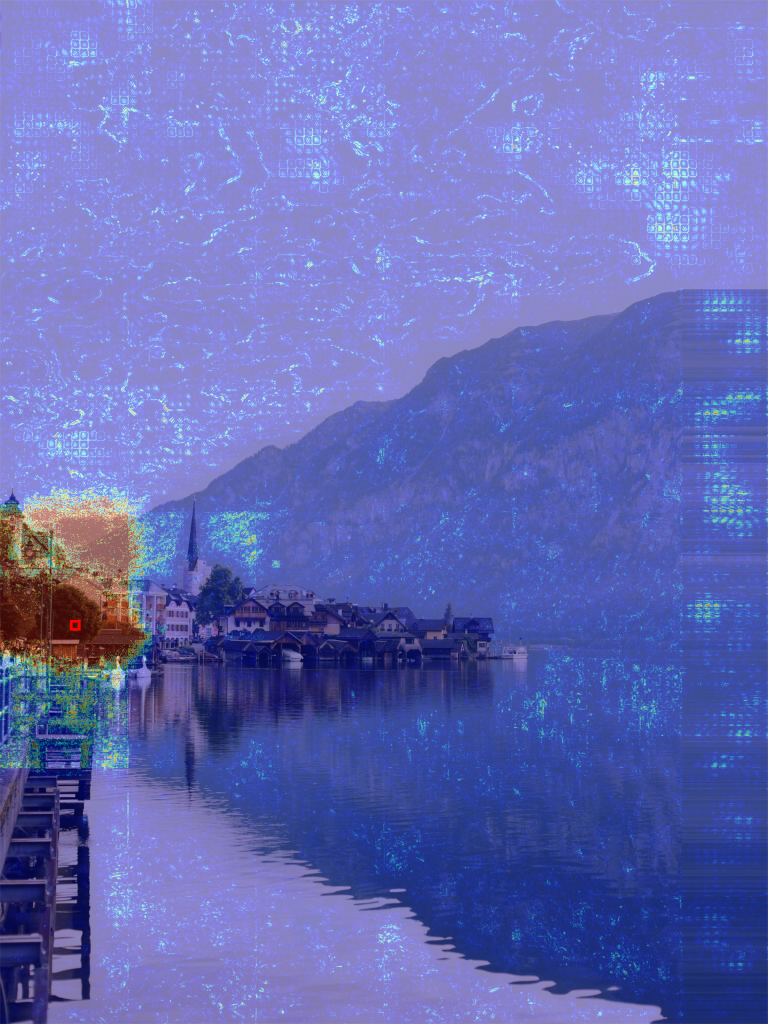}
        \end{subfigure}
        &
        \begin{subfigure}{0.11\textwidth}
            \centering
            \includegraphics[width=\linewidth]{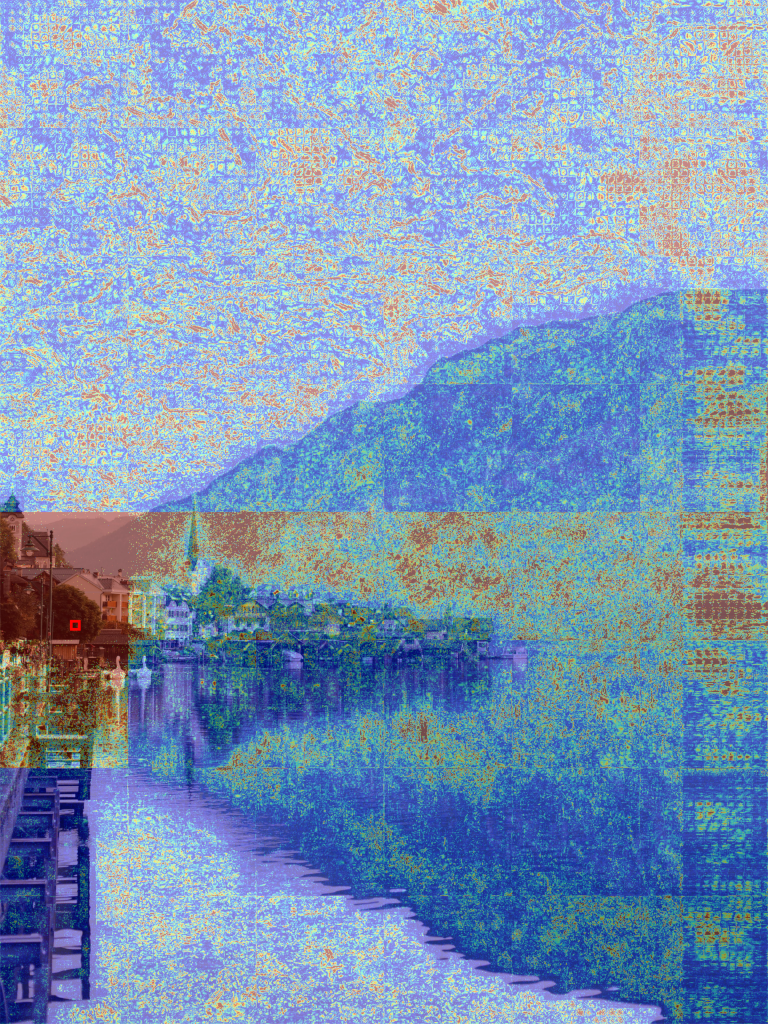}
        \end{subfigure}
        \\
        Ours & Bi-GLA & Ours & Bi-GLA \\
    \end{tabular}

    \caption{ERF visualization of inter-window communication at native $2048\times1536$ resolution. \textbf{Best viewed zoomed-in.}}
    \label{fig:erf_vis}
\end{figure}

\textbf{Inter-Window Communication Design.} We compare our design from~\cref{subsec:resadatok} with a simpler Bi-GLA variant (similar to~\citep{liu_2025}) using global concatenation without restoring spatial topology. As the source code of~\citep{liu_2025} is unavailable, we implement this variant ourselves. On CLIC 2020, our design achieves $21.95$ PSNR and $8.063$ FID, compared to $21.81$ and $8.803$ for Bi-GLA, corresponding to relative gains of $0.64\%$ and $8.41\%$, respectively. ERF visualizations in~\cref{fig:erf_vis} further show more spatially coherent receptive fields, with a structured checkerboard pattern and less noise, whereas Bi-GLA exhibits a predominantly row-wise pattern. This indicates that restoring spatial topology facilitates more effective inter-window interactions.

\textbf{Model Scaling Study.} We further investigate the effect of model scaling on the perceptual compression performance by evaluating PerCoV2 with Sana-0.6B and Sana-1.6B backbones on MSCOCO-30k, focusing on the $0.00146$--$0.00586$ bit range. We find that scaling the backbone from 0.6B to 1.6B consistently improves performance across all evaluated metrics. Specifically, PerCoV2 with the Sana-1.6B backbone requires on average $14.8\%$, $7.5\%$, and $6.4\%$ fewer bits for comparable FID, LPIPS, and MS-SSIM, respectively.







\subsection{Inference Latency}

We summarize the inference latency of PerCoV2 at $512\times512$ resolution on the Kodak dataset. We report the mean and standard deviation excluding the first three samples to mitigate CUDA warm-up effects. For PerCoV2, the flow model is run in \texttt{bf16}, and we assume pre-computed or empty captions. All measurements are performed on an NVIDIA H100 GPU. For each configuration, we select the operating point closest to $0.004$~bpp.

\begin{figure}[tb]
    \setlength{\tabcolsep}{1.0pt}  
    \renewcommand{\arraystretch}{1.0}  
    \centering
    \scriptsize
    \begin{tabular}{c|ccc}

        \toprule
        Original & PerCoV2 & PerCoV2 & PerCoV2 \\ 
         & $0.00146$~bpp & $0.00293$~bpp & $0.00586$~bpp \\ 
        \midrule
    
        \begin{subfigure}{0.114\textwidth}
            \centering
            \includegraphics[width=\linewidth]{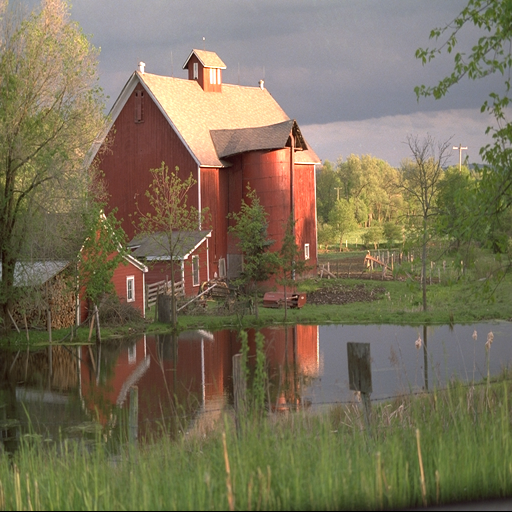}
        \end{subfigure}
        &
        \begin{subfigure}{0.114\textwidth}
            \centering
            \includegraphics[width=\linewidth]{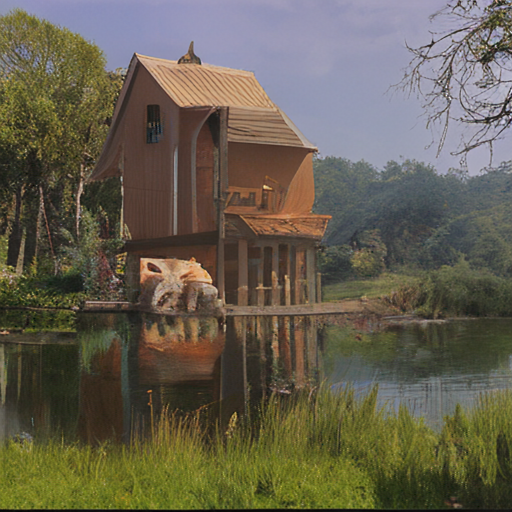}
        \end{subfigure}
        &
        \begin{subfigure}{0.114\textwidth}
            \centering
            \includegraphics[width=\linewidth]{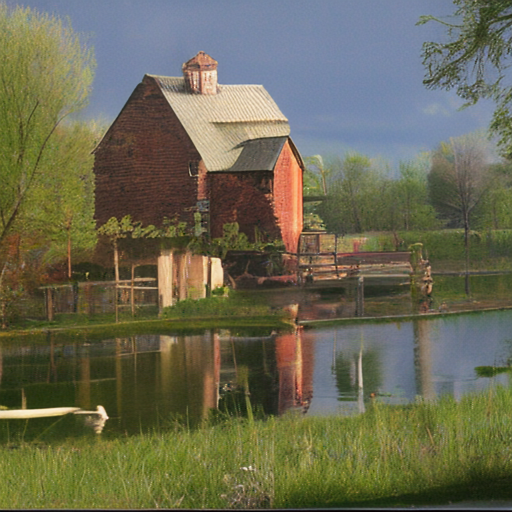}
        \end{subfigure}
        &
        \begin{subfigure}{0.114\textwidth}
            \centering
            \includegraphics[width=\linewidth]{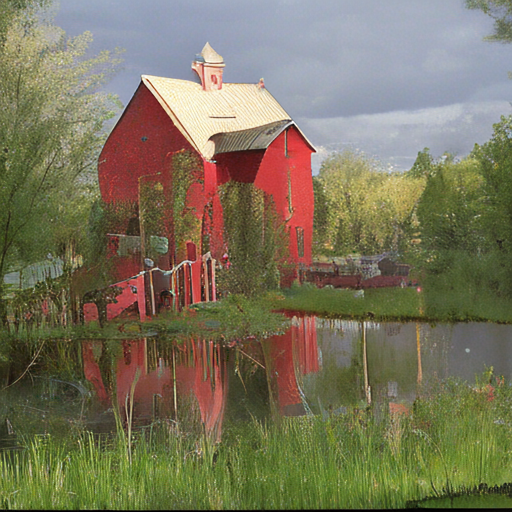}
        \end{subfigure}\\

        \begin{subfigure}{0.114\textwidth}
            \centering
            \includegraphics[width=\linewidth]{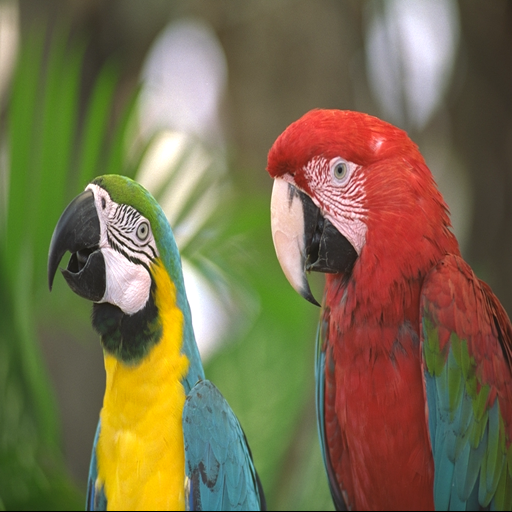}
        \end{subfigure}
        &
        \begin{subfigure}{0.114\textwidth}
            \centering
            \includegraphics[width=\linewidth]{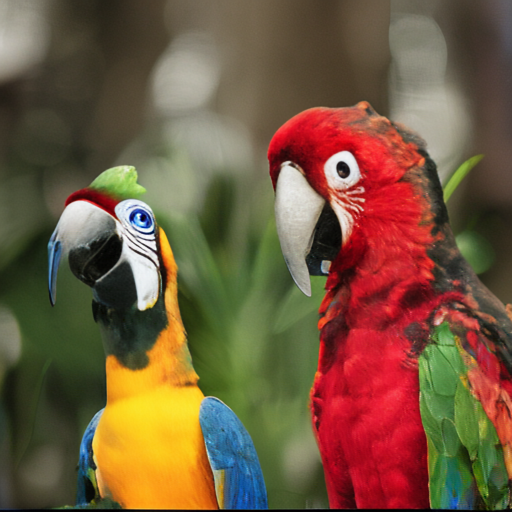}
        \end{subfigure}
        &
        \begin{subfigure}{0.114\textwidth}
            \centering
            \includegraphics[width=\linewidth]{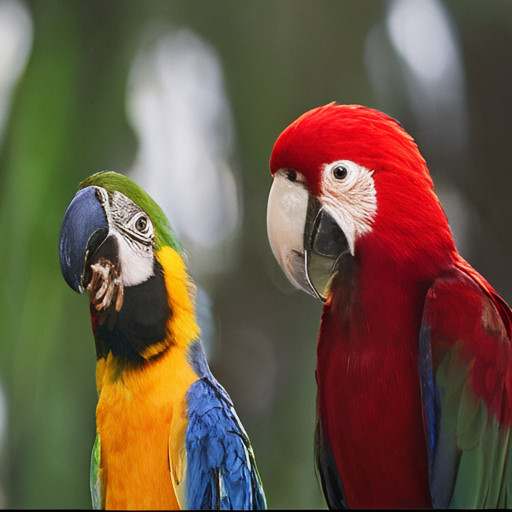}
        \end{subfigure}
        &
        \begin{subfigure}{0.114\textwidth}
            \centering
            \includegraphics[width=\linewidth]{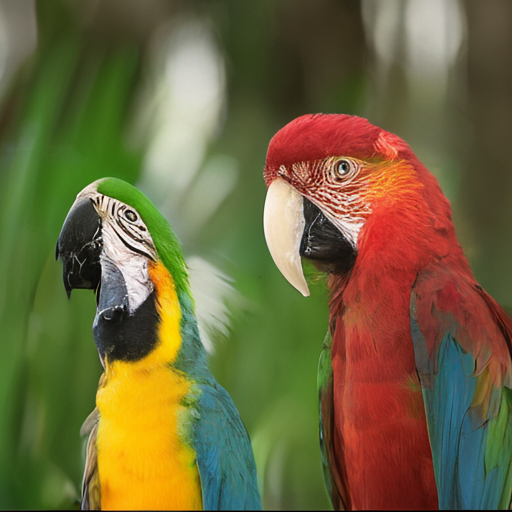}
        \end{subfigure}\\
        
        \bottomrule

    \end{tabular}

    \caption{Visual impressions at increasing bit rates. \textbf{Best viewed electronically.}}

    \label{fig:increasing_rate}
\end{figure}

\Cref{tab:computational_complexity} shows that PerCoV2 maintains low encoding latency, comparable to VAE-based methods such as MS-ILLM and DLF, while decoding is primarily dominated by flow-model inference. Our global configuration uses uniform coding with an encoding latency of $0.009$s, while patch-based TiTok variants with 1D-MIM entropy coding incur encoding latencies of up to $0.098$s due to additional window processing. In contrast, DiffC requires substantially higher latency due to repeated diffusion-model evaluations throughout the encoding process. On the decoding side, PerCoV2 (Sana-1.6B) takes $0.805$s, while its distilled counterpart reduces decoding latency by approximately $5.37\times$ and $5.16\times$ relative to PerCoV1 and PerCoV2, respectively. CoD-Lite, AEIC-ME, and OneDC achieve even lower inference latency through advanced architectural optimizations, including model pruning, feature distillation, and lightweight convolutional diffusion blocks; further efficiency improvements are orthogonal to our contributions and left for future work.


\subsection{Failure Modes} While PerCoV2 generally produces sharp and vibrant reconstructions across bit-rates (\cref{fig:increasing_rate}), we occasionally observe color drift, which is consistent with the behavior of VQ-based bottlenecks that prioritize semantic structure over exact chromatic fidelity. At ultra-low bit-rates, we also observe minor geometric deviations from the reference image, such as a slightly enlarged elephant head (see \cref{fig:vis_impressions_2}). These deviations reflect the generative nature of PerCoV2, which can favor perceptually plausible structures over strict pixel-level correspondence when the available bit-rate is highly constrained.

\begin{table}[tb]
\centering
\renewcommand{\arraystretch}{1.0} 
\caption{Encoding and decoding latency}
\label{tab:computational_complexity}
\begin{tabular}{l|cc}
\toprule
Model & Encoding (s) & Decoding (s) \\
\midrule
MS-ILLM & $0.062\pm0.003$ & $0.061\pm0.002$ \\
DLF & $0.063\pm0.002$ & $0.09\pm0.001$ \\
\midrule
CoD-Lite & $0.008\pm0.001$ & $0.025\pm0.0$ \\
OneDC & $0.065\pm0.001$ & $0.149\pm0.006$ \\
AEIC-ME & $0.028\pm0.0$ & $0.045\pm0.0$ \\
\midrule
DiffEIC & $0.117\pm0.019$ & $3.145\pm0.012$ \\
DiffC (SD v1.5) & $10.194\pm0.445$ & $1.158\pm0.027$ \\
DiffC (CoD) & $16.7\pm0.431$ & $1.236\pm0.024$ \\
CoD (latent) & $0.008\pm0.0$ & $0.848\pm0.004$ \\
PerCoV1 (SD v2.1) & $0.022\pm0.003$ & $0.837\pm0.002$ \\
\midrule
PerCoV2 (1.6B) & $0.009\pm0.0$ & $0.805\pm0.004$ \\
PerCoV2 (1.6B, dist.) & $0.009\pm0.0$ & $0.156\pm0.001$ \\
PerCoV2 (0.6B) & $0.009\pm0.0$ & $0.855\pm0.009$ \\
PerCoV2 (0.6B, dist.) & $0.009\pm0.0$ & $0.17\pm0.001$ \\

\bottomrule
\end{tabular}
\end{table}

\section{Conclusion}

In this work, we presented PerCoV2, an ultra-low bit-rate perceptual image compression system that unifies semantic tokenization, flow-based generation, and learned entropy modeling within a single framework. Building on the fully open flow-based SANA architecture, PerCoV2 introduces a novel resolution-adaptive 1D query-based tokenizer that produces compact image tokens serving as both a reconstruction prior and a conditioning signal for flow-based generation. By explicitly decoupling semantic representation from perceptual generation, PerCoV2 simplifies the flow-based learning objective, leading to more stable optimization and improved perceptual compression performance. Extensive experiments demonstrate state-of-the-art statistical fidelity, measured by FID and KID, across ultra-low and extreme bit-rates ($0.0015$--$0.025$ bpp) on MSCOCO-30k, while achieving strong zero-shot generalization to high-resolution benchmarks, including CLIC 2020 and DIV2K, when trained solely on SA-1B. These improvements also translate to better semantic fidelity, reflected in improved mIoU and CLIP scores, demonstrating practical relevance for downstream applications. Finally, the distilled single-step variant substantially accelerates decoding while preserving perceptual compression performance, taking a step toward practical perceptual compression with realistic reconstructions at ultra-low bit-rates.

\backmatter

\bmhead{Acknowledgements}

This work was supported by the German Federal Ministry of Education and Research through the Forschung an Fachhochschulen program (FKZ 13FH019KI2). We thank Lambda Labs for GPU cloud credits through its Research Grant Program, Marlène Careil for her valuable insights and evaluation data, and Jeremy Vonderfecht for helpful clarifications regarding DiffC.

\bibliography{sn-bibliography}

\end{document}